%% file: ms.tex
\documentclass[10pt,twocolumn,letterpaper]{article}

\usepackage[pagenumbers]{cvpr} %

\usepackage[dvipsnames,table]{xcolor}
\usepackage{graphicx}
\usepackage{amsmath}
\usepackage{amssymb}
\usepackage{booktabs}
\usepackage{tocloft}

\usepackage{colortbl}
\usepackage{amssymb}%
\usepackage{pifont}%
\usepackage{tocloft}
\definecolor{cvprblue}{rgb}{0.21,0.49,0.74}
\usepackage[pagebackref,breaklinks,colorlinks,citecolor=cvprblue]{hyperref}

\usepackage[capitalize]{cleveref}
\crefname{section}{Sec.}{Secs.}
\Crefname{section}{Sec.}{Sections}
\Crefname{table}{Table}{Tables}
\crefname{table}{Tab.}{Tabs.}
\Crefname{figure}{Fig.}{Figs.}
\Crefname{equation}{Eq.}{Eqs.}

\definecolor{mygreen}{HTML}{3C5A01}
\definecolor{myred}{HTML}{C61A1A}
\definecolor{myblue}{HTML}{01B0F1}

\input{math_commands}

\def\paperID{10507} %
\def\confName{CVPR}
\def\confYear{2024}

\title{\vspace{-2.8em} Align Your Gaussians:\\Text-to-4D with Dynamic 3D Gaussians and Composed Diffusion Models\vspace{-0.2em}}

\author{
Huan Ling\textsuperscript{1,2,3\;\,*}
\quad
Seung Wook Kim\textsuperscript{1,2,3\;\,*}
\quad
Antonio Torralba\textsuperscript{4}
\quad
Sanja Fidler\textsuperscript{1,2,3}
\quad
Karsten Kreis\textsuperscript{1}
\vspace{0.2cm}
\\
{\small\textsuperscript{1}NVIDIA \quad \textsuperscript{2}Vector Institute \quad \textsuperscript{3}University of Toronto \quad \textsuperscript{4}MIT\vspace{0pt}}
\vspace{0.15cm}
\\
\small \textit{Project page:} \url{https://research.nvidia.com/labs/toronto-ai/AlignYourGaussians/}
}

\begin{document}

\twocolumn[{
    \renewcommand\twocolumn[1][]{#1}
    \maketitle
    \begin{center}
        \input{figures_tex/teaser.tex}
    \end{center}
}]

\input{sections/0_abstract}

\input{sections/1_intro}

\input{sections/2_background}
\input{sections/3_method}

\input{sections/4_experiments}

\input{sections/5_conclusions}

{
    \small
    \bibliographystyle{ieeenat_fullname}
    \addcontentsline{toc}{section}{References}
    \bibliography{bibliography}
}

\newpage
\appendix
\onecolumn

\clearpage
\begingroup
\hypersetup{linkcolor=black}
\setlength{\cftbeforesecskip}{8pt}
\tableofcontents
\endgroup
\clearpage

\input{appendices/overview_supp_videos}
\input{appendices/realted_work_extended}

\input{appendices/extended_4d_rep_details}

\input{appendices/extended_composit_gen_details}

\input{appendices/experiment_details}
\input{appendices/more_quantitative_results}

\input{appendices/more_qualitative_results}

\end{document}

%% file: math_commands.tex
\usepackage{amsmath,amsfonts,bm,mathtools}

\def\eqref#1{equation~\ref{#1}}

\def\1{\bm{1}}

\def\rvx{{\mathbf{x}}}

\def\rvz{{\mathbf{z}}}

\def\rmX{{\mathbf{X}}}

\def\rmZ{{\mathbf{Z}}}

\def\mI{{\bm{I}}}

\DeclareMathAlphabet{\mathsfit}{\encodingdefault}{\sfdefault}{m}{sl}
\SetMathAlphabet{\mathsfit}{bold}{\encodingdefault}{\sfdefault}{bx}{n}

%% file: figures_tex/teaser.tex
\vspace{-17pt}
    \centering
    \includegraphics[width=\linewidth]{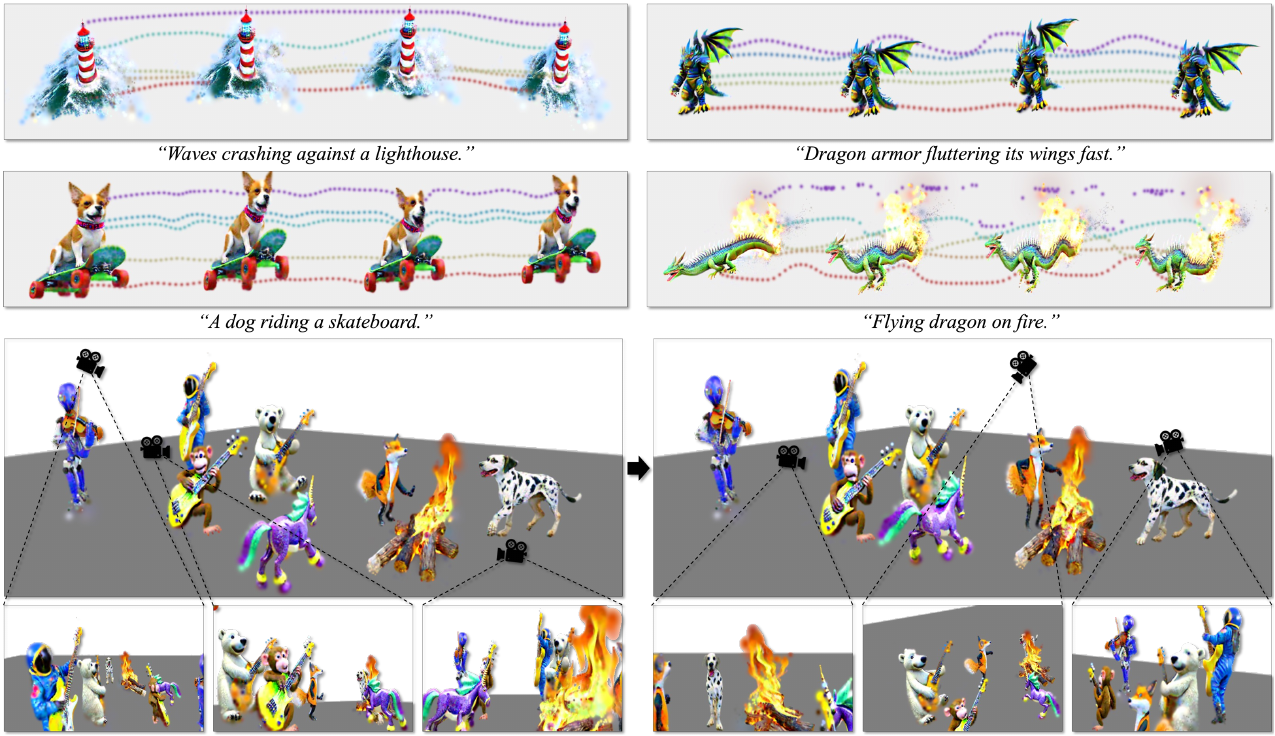}
\vspace{-0.7cm}
\captionof{figure}{
    \small \textbf{Text-to-4D synthesis with \textit{Align Your Gaussians (AYG)}.} \textit{Top:} Different dynamic 4D sequences. Dotted lines represent dynamics of deformation field. \textit{Bottom:} Multiple dynamic 4D objects are composed within a large dynamic scene; two time frames shown.
}
\label{fig:teaser}

%% file: sections/0_abstract.tex
\begin{abstract}
Text-guided diffusion models have revolutionized image and video generation and have also been successfully used for optimization-based 3D object synthesis. Here, we instead focus on the underexplored text-to-4D setting and synthesize dynamic, animated 3D objects using score distillation methods with an additional temporal dimension. Compared to previous work, we pursue a novel compositional generation-based approach, and combine text-to-image, text-to-video, and 3D-aware multiview diffusion models to provide feedback during 4D object optimization, thereby simultaneously enforcing temporal consistency, high-quality visual appearance and realistic geometry. Our method, called \textbf{Align Your Gaussians (AYG)}, leverages dynamic 3D Gaussian Splatting with deformation fields as 4D representation. 
Crucial to AYG is a novel method to regularize the distribution of the moving 3D Gaussians and thereby stabilize the optimization and induce motion. We also propose a motion amplification mechanism as well as a new autoregressive synthesis scheme to generate and combine multiple 4D sequences for longer generation.
These techniques allow us to synthesize vivid dynamic scenes, outperform previous work qualitatively and quantitatively and achieve state-of-the-art text-to-4D performance. Due to the Gaussian 4D representation, different 4D animations can be seamlessly combined, as we demonstrate. AYG opens up promising avenues for animation, simulation and digital content creation as well as synthetic data generation.
{\let\thefootnote\relax\footnote{{\textsuperscript{*}Equal contribution.}}}
\end{abstract}

%% file: sections/1_intro.tex
\vspace{-5mm}
\section{Introduction}\label{sec:intro}
\input{figures_tex/pipeline_second_stage}
Generative modeling of dynamic 3D scenes has the potential to revolutionize how we create games, movies, simulations, animations and entire virtual worlds. 
Many works have shown how a wide variety of 3D objects can be synthesized via score distillation techniques~\cite{poole2023dreamfusion,lin2023magic3d,chen2023fantasia3d,wang2023score,tsalicoglou2024textmesh,zhu2023hifa,huang2023dreamtime,chen2023it3d,wang2023prolificdreamer,metzer2023latent,singer2023mav3d},
but they typically only synthesize static 3D scenes, although we live in a moving, dynamic world.
While image diffusion models have been successfully extended to video generation~\cite{blattmann2023videoldm,singer2023makeavideo,ho2022imagen,zhou2023magicvideo,wang2023videofactory,wang2023lavie,an2023latentshift,ge2022pyoco}, there is little research on similarly extending 3D synthesis
to 4D generation with an additional temporal dimension.

We propose \textit{Align Your Gaussians (AYG)}, a novel method for 4D content creation. In contrast to previous work~\cite{singer2023mav3d}, we leverage dynamic 3D Gaussians~\cite{kerbl20233Dgaussians} as backbone 4D representation, where a deformation field~\cite{pumarola2020d,park2021nerfies} captures scene dynamics and transforms the collection of 3D Gaussians to represent object motion. AYG takes a compositional generation-based perspective and leverages the combined gradients of latent text-to-image~\cite{rombach2021highresolution}, text-to-video~\cite{blattmann2023videoldm} and 3D-aware text-to-multiview-image~\cite{shi2023mvdream} diffusion models in a score distillation-based synthesis framework. 
A 3D-aware multiview diffusion model and a regular text-to-image model are used to generate an initial high-quality 3D shape. Afterwards, we compose the gradients of a text-to-video and a text-to-image model; the gradients of the text-to-video model optimize the deformation field to capture temporal dynamics, while the text-to-image model ensures that high visual quality is maintained for all time frames (\Cref{fig:pipeline}). To this end, we trained a dedicated text-to-video model; it is conditioned on the frame rate and can create useful gradients both for short and long time intervals, which allows us to generate long and smooth 4D sequences.

We developed several techniques to ensure stable optimization and learn vivid dynamic 4D scenes in AYG: We employ a novel regularization method that 
uses a modified version of the Jensen-Shannon divergence to regularize the locations of the 3D Gaussians such that the mean and variance of the set of 3D Gaussians is preserved as they move. Furthermore, we use a motion amplification method that carefully scales the gradients from the text-to-video model and enhances motion.
To extend the length of the 4D sequences or combine different dynamic scenes with changing text guidance, we introduce an autoregressive generation scheme which interpolates the deformation fields of consecutive sequences. We also propose a new view-guidance method to generate consistent 3D scenes for initialization of the 4D stage, and we leverage the concurrent classifier score distillation method~\cite{yu2023csd}.

We find that AYG can generate diverse, vivid, detailed and 3D-consistent dynamic scenes (\Cref{fig:teaser}), achieving state-of-the-art text-to-4D performance. 
We also show long, autoregressively extended 4D scenes, including ones with varying text guidance, which has not been demonstrated before. A crucial advantage of AYG's 4D Gaussian backbone representation is that
different 4D animations can trivially be combined and composed together, which we also show.

We envision broad applications in digital content creation, where AYG takes a step beyond the literature on text-to-3D and captures our world's rich dynamics. 
Moreover, AYG can generate 4D scenes with exact tracking labels for free, a promising feature for synthetic data generation.

\textbf{Contributions.} \textit{(i)} We propose AYG, a system for text-to-4D content creation leveraging dynamic 3D Gaussians with deformation fields as 4D representation. \textit{(ii)} We show how to tackle the text-to-4D task through score distillation within a new compositional generation framework, combining 2D, 3D, and video diffusion models. 
\textit{(iii)} To scale AYG, we introduce a novel regularization method and a new motion amplification technique.
\textit{(iv)} Experimentally, we achieve state-of-the-art text-to-4D performance and generate high-quality, diverse, and dynamic 4D scenes.
\textit{(v)} For the first time, we also show how our 4D sequences can be extended in time with a new autoregressive generation scheme and even creatively composed in large scenes.

%% file: figures_tex/pipeline_second_stage.tex
\begin{figure*}[t!]
  \vspace{-0.3cm}
  \begin{minipage}[c]{0.69\textwidth}
  \vspace{-5mm}
    \includegraphics[width=1.0\textwidth]{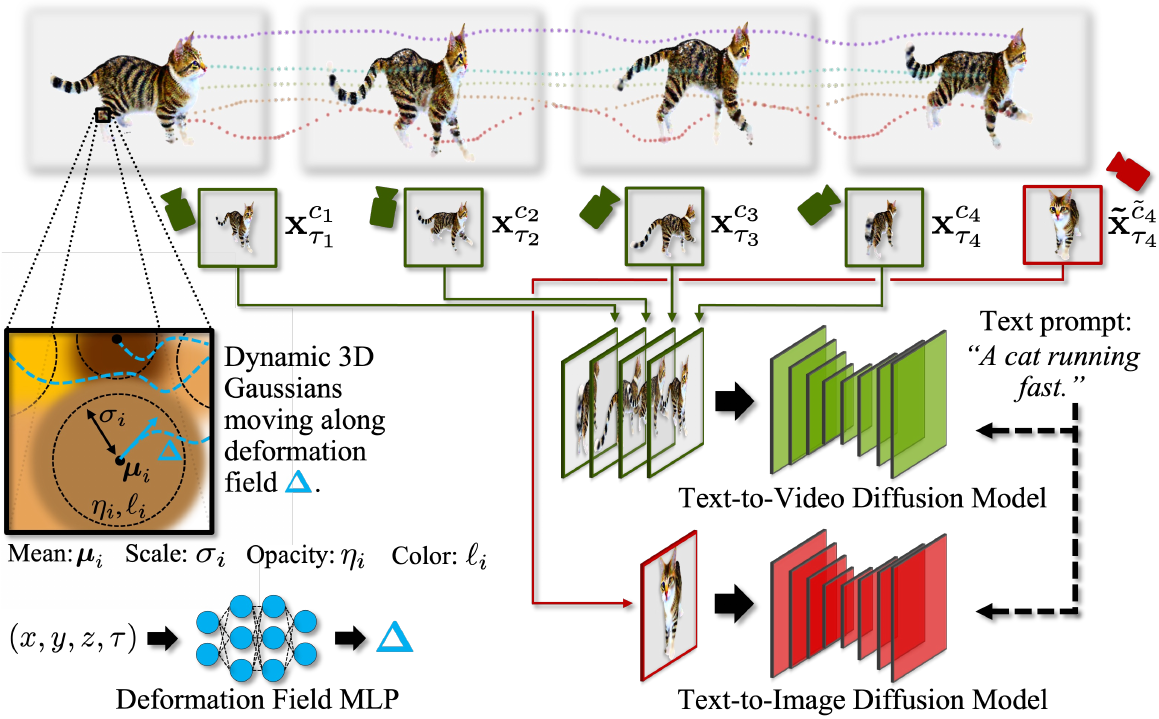}
  \end{minipage}\hfill
  \begin{minipage}[c]{0.3\textwidth}
  \vspace{-3mm}
    \caption{\small \textbf{Text-to-4D synthesis with AYG.} We generate dynamic 4D scenes via score distillation. We initialize the 4D sequence from a static 3D scene (generated first, \Cref{fig:static3dgen}), which is represented by 3D Gaussians with means $\boldsymbol{\mu}_i$, scales $\sigma_i$, opacities $\eta_i$ and colors $\ell_i$. Consecutive rendered frames $\mathbf{x}_{\tau_j}^{c_j}$ from the 4D sequence at times $\tau_j$ and camera positions $c_j$ are diffused and fed to a text-to-video diffusion model~\cite{blattmann2023videoldm} (\textcolor{mygreen}{\textbf{green arrows}}), which provides a distillation gradient that is backpropagated through the rendering process into a deformation field $\color{myblue}{\boldsymbol{\Delta}}\color{black}{(x,y,z,\tau)}$ (dotted lines) that captures scene motion. Simultaneously, random frames $\tilde{\mathbf{x}}_{\tau_j}^{\tilde{c}_j}$ are diffused and given to a text-to-image diffusion model~\cite{rombach2021highresolution} (\textcolor{myred}{\textbf{red arrows}}) whose gradients ensure that high visual quality is maintained frame-wise.}
    \label{fig:pipeline}
  \end{minipage}
  \vspace{-9mm}
\end{figure*}

%% file: sections/2_background.tex
\section{Background}\label{sec:background}
\textbf{3D Gaussian Splatting~\cite{kerbl20233Dgaussians}} represents 3D scenes by $N$ 3D Gaussians with positions $\boldsymbol{\mu}_i$, covariances $\Sigma_i$, opacities $\eta_i$ and colors $\ell_i$ (\Cref{fig:pipeline}). Rendering corresponds to projection of the 3D Gaussians onto the 2D camera's image plane, producing 2D Gaussians with projected means $\hat{\boldsymbol{\mu}}_i$ and covariances $\hat{\Sigma}_i$. The color $\mathcal{C}(\mathbf{p})$ of image pixel $\mathbf{p}$ can be calculated through point-based volume rendering~\cite{zwicker2001splatting} as
\begin{align}
    \mathcal{C}(\mathbf{p}) & = \sum_{i=1}^N \ell_i \alpha_i \prod^{i-1}_{j=1}\left(1-\alpha_j\right), \\ 
    \alpha_i &= \eta_i \,\textrm{exp}\left[-\frac{1}{2}\left(\mathbf{p}-\hat{\boldsymbol{\mu}}_i\right)^\top\hat{\Sigma}^{-1}_i\left(\mathbf{p}-\hat{\boldsymbol{\mu}}_i\right)\right],
\end{align}
where $j$ iterates over the Gaussians along the ray through the scene from pixel $\mathbf{p}$ until Gaussian $i$. To accelerate rendering, the image plane can be divided into tiles, which are processed in parallel.
Initially proposed for 3D scene reconstruction, 3D Gaussian Splatting uses gradient-based thresholding to densify areas that need more Gaussians to capture fine details, and unnecessary Gaussians with low opacity are pruned every few thousand optimization steps.

\textbf{Diffusion Models and Score Distillation Sampling.}
Diffusion-based generative models (DMs)~\cite{sohl2015deep,ho2020ddpm,song2020score,dhariwal2021diffusion,nichol2021improved} use a forward diffusion process that gradually perturbs data, such as images or entire videos, towards entirely random noise, while a neural network is learnt to denoise and reconstruct the data. 
DMs have also been widely used for score distillation-based generation of 3D objects~\cite{poole2023dreamfusion}. In that case, a 3D object, represented for instance by a neural radiance field (NeRF)~\cite{mildenhall2020nerf} or 3D Gaussians~\cite{kerbl20233Dgaussians}, like here, with parameters $\boldsymbol{\theta}$ is rendered from different camera views and the renderings $\rvx$ are diffused and given to a text-to-image DM. In the score distillation sampling (SDS) framework, the DM's denoiser is then used to construct a gradient that is backpropagated through the differentiable rendering process $g$ into the 3D scene representation and updates the scene representation to make the scene rendering look more realistic, like images modeled by the DM. Rendering and using DM feedback from many different camera perspectives then encourages the scene representation to form a geometrically consistent 3D scene.
The SDS gradient~\cite{poole2023dreamfusion} is
\begin{equation}\nonumber
\nabla_{\boldsymbol{\theta}}\mathcal{L}_{\textrm{SDS}}(\mathbf{x}=g(\boldsymbol{\theta}))=\mathbb{E}_{t,\boldsymbol{\epsilon}}\left[w(t)\left(\hat{\boldsymbol{\epsilon}}_\phi(\mathbf{z}_t,v,t)-\boldsymbol{\epsilon}\right)\frac{\partial\mathbf{x}}{\partial\boldsymbol{\theta}}\right],
\end{equation}
where $\rvx$ denotes the 2D rendering, $t$ is the time up to which the diffusion is run to perturb $\rvx$, $w(t)$ is a weighting function, and $\rvz_t$ is the perturbed rendering. Further, $\hat{\boldsymbol{\epsilon}}_\phi(\mathbf{z}_t,v,t)$ is the DM's denoiser neural network that predicts the diffusion noise $\boldsymbol{\epsilon}$. It is conditioned on $\rvz_t$, the diffusion time $t$ and a text prompt $v$ for guidance. Classifier-free guidance (CFG)~\cite{ho2021classifierfree} typically amplifies the text conditioning. 

\subsection{Related Work}
\input{figures_tex/pipeline_first_stage}
\textit{See Supp. Material for an extended discussion. Here, we only briefly mention the most relevant related literature.}

As discussed, AYG builds on text-driven image~\cite{saharia2022photorealistic,rombach2021highresolution,ramesh2022dalle2,balaji2022ediffi,feng2023ernievilg,podell2023sdxl,dai2023emu,xue2023raphael}, video~\cite{blattmann2023videoldm,singer2023makeavideo,ho2022imagen,zhou2023magicvideo,wang2023videofactory,wang2023lavie,an2023latentshift,ge2022pyoco,wu2023tune,khachatryan2023text2videozero,guo2023animatediff} and 3D-aware DMs~\cite{liu2023zero123,qian2023magic123,liu2023syncdreamer,shi2023mvdream,shi2023zero123plus,liu2023one2345,nichol2022pointe,zeng2022lion}, uses score distillation sampling~\cite{poole2023dreamfusion,lin2023magic3d,chen2023fantasia3d,wang2023score,tsalicoglou2024textmesh,zhu2023hifa,huang2023dreamtime,chen2023it3d,wang2023prolificdreamer,metzer2023latent,deng2023nerdi,xu2023neuralLift,lorraine2023att3d} and leverages 3D Gaussian Splatting~\cite{kerbl20233Dgaussians} as well as deformation fields~\cite{pumarola2020d,park2021nerfies,Cai2022NDR,tretschk2021nonrigid,park2021hypernerf} for its 4D representation. The concurrent works DreamGaussian~\cite{tang2023dreamgaussian}, GSGEN~\cite{chen2023gsgen} and GaussianDreamer~\cite{yi2023gaussiandreamer} use 3D Gaussian Splatting to synthesize static 3D scenes, but do not consider dynamics. Dynamic 3D Gaussian Splatting has been used for 4D reconstruction~\cite{luiten2023dynamic,wu20234d,zielonka2023drivable}, but not for 4D generation. The idea to compose the gradients of multiple DMs has been used before for controllable image generation~\cite{liu2022compositional,du2023reduce}, but has received little attention in 3D or 4D synthesis.\looseness=-1

Most related to AYG is \textit{Make-A-Video3D (MAV3D)}~\cite{singer2023mav3d}, to the best of our knowledge the only previous work that generates dynamic 4D scenes with score distillation. MAV3D uses NeRFs with HexPlane~\cite{Cao2022hexplane} features as 4D representation, in contrast to AYG's dynamic 3D Gaussians, and it does not disentangle its 4D representation into a static 3D representation and a deformation field modeling dynamics. MAV3D's representation prevents it from composing multiple 4D objects into large dynamic scenes, which our 3D Gaussian plus deformation field representation easily enables, as we show. Moreover, MAV3D's sequences are limited in time, while we show a novel autoregressive generation scheme to extend our 4D sequences.
AYG outperforms MAV3D qualitatively and quantitatively and synthesizes significantly higher-quality 4D scenes. 
Our novel compositional generation-based approach contributes to this, which MAV3D does not pursue. Finally, instead of regular SDS, used by MAV3D, in practice AYG employs classifier score distillation~\cite{yu2023csd} (see \Cref{sec:scaling_ayg}).

%% file: figures_tex/pipeline_first_stage.tex
\begin{figure}[t!]
  \vspace{-0.4cm}
    \begin{center}
    \includegraphics[width=0.42\textwidth]{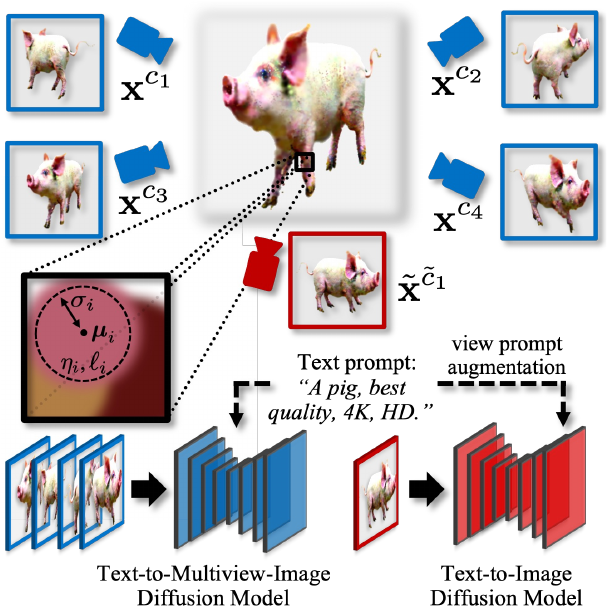}
    \end{center}
    \vspace{-0.5cm}
    \caption{\small In \textbf{AYG's initial 3D stage} we synthesize a static 3D scene leveraging a text-guided multiview diffusion model~\cite{shi2023mvdream} and a regular text-to-image model~\cite{rombach2021highresolution}. The text-to-image model receives viewing angle-dependent text prompts and leverages view guidance (\Cref{sec:scaling_ayg}). See \Cref{fig:pipeline} for 4D stage and descriptions.} 
    \label{fig:static3dgen}
  \vspace{-6mm}
\end{figure}

%% file: sections/3_method.tex
\vspace{-1mm}
\section{Align Your Gaussians}\label{sec:method}
\vspace{-1mm}
\input{figures_tex/jsd_reg_figure}
In \Cref{sec:ayg_4drep}, we present AYG's 4D representation, and in \Cref{sec:compo_gen}, we introduce its compositional generation framework with multiple DMs. In \Cref{sec:ayd_score_distill}, we lay out AYG's score distillation framework in practice, and in \Cref{sec:scaling_ayg}, we discuss several novel methods and extensions to scale AYG.

\subsection{AYG's 4D Representation} \label{sec:ayg_4drep}
AYG's 4D representation combines 3D Gaussian Splatting~\cite{kerbl20233Dgaussians} with deformation fields~\cite{pumarola2020d,park2021nerfies} to capture the 3D scene and its temporal dynamics in a disentangled manner. Specifically, each 4D scene consists of a set of $N$ 3D Gaussians as in \Cref{sec:background}. 
Following \citet{kerbl20233Dgaussians}, we also use two degrees of spherical harmonics to model view-dependent effects, this is, directional color, and thereby improve the 3D Gaussians' expressivity.
Moreover, we restrict the 3D Gaussians' covariance matrices to be isotropic with scales $\sigma_i$. We made this choice as our 3D Gaussians move as a function of time and learning expressive dynamics is easier for spherical Gaussians. We denote the collection of learnable parameters of our 3D Gaussians as $\boldsymbol{\theta}$.
The scene dynamics are modeled by a deformation field $\boldsymbol{\Delta}_{\boldsymbol{\Phi}}(x,y,z,\tau)=(\Delta x,\Delta y,\Delta z)$, defined through a multi-layer perceptron (MLP) with parameters $\boldsymbol{\Phi}$. Specifically, for any 3D location $(x,y,z)$ and time $\tau$, the deformation field predicts a displacement $(\Delta x,\Delta y,\Delta z)$. The 3D Gaussians smoothly follow these displacements to represent a moving and deforming 4D scene (\Cref{fig:pipeline}).
Note that in practice we preserve the initial 3D Gaussians for the first frame, \ie $\boldsymbol{\Delta}_{\boldsymbol{\Phi}}(x,y,z,0) = (0,0,0)$, by setting 
$\boldsymbol{\Delta}_{\boldsymbol{\Phi}}(x,y,z,\tau) = (\xi(\tau)\Delta x,\xi(\tau)\Delta y,\xi(\tau)\Delta z)$
where $\xi(\tau)=\tau^{0.35}$ such that $\xi(0) = 0$ and $\xi(1) = 1$.
Following \citet{luiten2023dynamic}, we regularize the deformation field so that nearby Gaussians deform similarly (``rigidity regularization'', see Supp. Mat.).

Apart from the intuitive decomposition into a backbone 3D representation and a deformation field to model dynamics, a crucial advantage of AYG's dynamic 3D Gaussian-based representation is that different dynamic scenes, each with its own set of Gaussians and deformation field, can be easily combined, thereby enabling promising 3D dynamic content creation applications (see \Cref{fig:teaser}). This is due to the explicit nature of this representation, in contrast to typical NeRF-based representations. Moreover, learning 4D scenes with score distillation requires many scene renderings. This also makes 3D Gaussians ideal due to their rendering efficiency~\cite{kerbl20233Dgaussians}. Note that early on we also explored MAV3D's HexPlane- and NeRF-based 4D representation~\cite{singer2023mav3d}, but we were not able to achieve satisfactory results. 

\input{figures_tex/autoreg_concept_figure}
\input{figures_tex/main_results_figure}
\subsection{Text-to-4D as Compositional Generation} \label{sec:compo_gen}
We would like AYG's synthesized dynamic 4D scenes to be of high visual quality, be 3D-consistent and geometrically correct, and also feature expressive and realistic temporal dynamics. This suggests to compose different text-driven DMs during the distillation-based generation to capture these different aspects. \textit{(i)} We use the text-to-image model Stable Diffusion (SD)~\cite{rombach2021highresolution}, which has been trained on a broad set of imagery and provides a strong general image prior. 
\textit{(ii)} We also utilize the 3D-aware text-conditioned multi-view DM MVDream~\cite{shi2023mvdream}, which generates multiview images of 3D objects, was fine-tuned from SD on the object-centric 3D dataset Objaverse~\cite{deitke2023objaverse,deitke2023objaverseXL} and provides a strong 3D prior. 
It defines a distribution over four multiview-consistent images corresponding to object renderings from four different camera perspectives $c_1,...,c_4$.
Moreover, we train a text-to-video DM, following VideoLDM~\cite{blattmann2023videoldm}, but with a larger text-video dataset (HDVG-130M~\cite{wang2023videofactory} and Webvid-10M~\cite{bain21frozen}) and additional conditioning on the videos' frame rate (see Supp. Material for details). This video DM provides temporal feedback when rendering 2D frame sequences from our dynamic 4D scenes. 
All used DMs are \textit{latent} DMs~\cite{rombach2021highresolution,vahdat2021score}, which means that in practice we first encode renderings of our 4D scenes into the models' latent spaces, calculate score distillation gradients there, and backpropagate them through the models' encoders. All DMs leverage the SD 2.1 backbone and share the same encoder. To keep the notation simple, we do not explicitly incorporate the encoding into our mathematical description below and the visualizations (\Cref{fig:pipeline,fig:static3dgen}).

We disentangle optimization into first synthesizing a static 3D Gaussian-based object $\boldsymbol{\theta}$, and then learning the deformation field $\boldsymbol{\Phi}$ to add scene dynamics.  

\textbf{Stage 1: 3D Synthesis (\Cref{fig:static3dgen}).} We first use MVDream's multiview image prior to generate a static 3D scene via score distillation (Supp. Mat. for details). Since MVDream on its own would generate objects in random orientations, we enforce a canonical pose by combining MVDream's gradients with those of regular SD, while augmenting the text-conditioning for SD with directional texts ``front view'', ``side view'', ``back view'' and ``overhead view''~\cite{poole2023dreamfusion}. Formally, we can derive a score distillation gradient (see \Cref{sec:ayd_score_distill}) by minimizing the reverse Kulback-Leibler divergence (KLD) from the rendering distribution to the product of the composed MVDream and SD model distributions
\begin{equation}\nonumber
    \textrm{KL}\biggl(q_{\boldsymbol{\theta}}\left(\{\rvz^{c_i}\}_4,\{\tilde{\rvz}^{\tilde{c}_j}\}_K\right)\bigg|\bigg|p^\alpha_\textrm{3D}\left(\{\rvz^{c_i}\}_4\right) \prod_{j=1}^K p^\beta_\textrm{im}\left(\tilde{\rvz}^{\tilde{c}_j}\right)\biggr),
\end{equation}
similar to \citet{poole2023dreamfusion} (App. A.4). Here, $p_\textrm{3D}(\{\rvz^{c_i}\}_4)$ represents the MVDream-defined multiview image distribution over four diffused renderings from camera views $c_i$, denoted as the set $\{\rvz^{c_i}\}_4$ (we omit the diffusion time $t$ subscript for brevity). Moreover, $p_\textrm{im}(\tilde{\rvz}^{\tilde{c}_j})$ is the SD-based general image prior and $\{\tilde{\rvz}^{\tilde{c}_j}\}_K$ is another set of $K$ diffused scene renderings. In principle, the renderings for SD and MVDream can be from different camera angles $c_i$ and $\tilde{c}_j$, but in practice we choose $K{=}4$ and use the same renderings. Furthermore, $\alpha$ and $\beta$ are adjustable temperatures of the distributions $p_\textrm{3D}$ and $p_\textrm{im}$, and $q_{\boldsymbol{\theta}}$ denotes the distribution over diffused renderings defined by the underlying 3D scene representation ${\boldsymbol{\theta}}$, which is optimized through the differentiable rendering.
We also use the Gaussian densification method discussed in \Cref{sec:background} (see Supp. Material).

\input{figures_tex/autoreg_loopy_results}
\textbf{Stage 2: Adding Dynamics for 4D Synthesis (\Cref{fig:pipeline}).} While in stage 1, we only optimize the 3D Gaussians, in stage 2, the main 4D stage, we optimize (only) the deformation field $\boldsymbol{\Phi}$ to capture motion and extend the static 3D scene to a dynamic 4D scene with temporal dimension $\tau$. To this end, we compose the text-to-image and text-to-video DMs and formally minimize a reverse KLD of the form
\begin{equation} \nonumber
    \textrm{KL}\biggl(q_{\boldsymbol{\Phi}}\left(\{\rvz_{\tau_i}^{c_i}\}_F,\{\tilde{\rvz}_{\tilde{\tau}_j}^{\tilde{c}_j}\}_M\right)\bigg|\bigg| p^\gamma_\textrm{vid}\left(\{\rvz_{\tau_i}^{c_i}\}_F\right)
    \prod_{j=1}^M p^\kappa_\textrm{im}\left(\tilde{\rvz}_{\tilde{\tau}_j}^{\tilde{c}_j}\right)\biggr),
\end{equation}
where $p_\textrm{vid}(\{\rvz_{\tau_i}^{c_i}\}_F)$ is the video DM-defined distribution over $F$ 4D scene renderings $\{\rvz_{\tau_i}^{c_i}\}_F$ taken at times $\tau_i$ and camera angles $c_i$ ($F{=}16$ for our model). Similar to before, $M$ additional renderings are given to the SD-based general image prior, and $\gamma$ and $\kappa$ are temperatures. The renderings $\{\tilde{\rvz}_{\tilde{\tau}_j}^{\tilde{c}_j}\}_M$ fed to regular SD can be taken at different times $\tilde{\tau}_j$ and cameras $\tilde{c}_j$ than the video model frames, but in practice $M{=}4$ and we use three random renderings as well as the 8th middle frame among the ones given to the video model. $q_{\boldsymbol{\Phi}}$ defines the distribution over renderings by the 4D scene with the learnable deformation field parameters $\boldsymbol{\Phi}$.
We could render videos from the 4D scene with a fixed camera, but in practice dynamic cameras, \textit{i.e.} varying $c_i$, help to learn more vivid 4D scenes, similar to \citet{singer2023mav3d}.\looseness=-1

Moreover, following \citet{singer2023mav3d}, our video DM is conditioned on the frame rate (fps) and we choose the times $0\le \tau_i \le 1$ accordingly by sampling fps $\in \{4,8,12\}$ and the starting time. 
We render videos from the 4D scene and condition the video DM with the sampled fps.
This helps generating not only sufficiently long but also temporally smooth 4D animations, as different fps correspond to long-term and short-term dynamics. Therefore, when rendering short but high fps videos they only span part of the entire length of the 4D sequence.
Also see Supp. Material.

Optimizing the deformation field while supervising both with a video and image DM is crucial. The video DM generates temporal dynamics, but text-to-video DMs are not as robust as general text-to-image DMs. Including the image DM during this stage ensures stable optimization and that high visual frame quality is maintained (ablations in \Cref{sec:experiments}).

A crucial advantage of the disentangled two stage design is that AYG's main 4D synthesis method---the main innovation of this work---could in the future in principle also be applied to 3D objects originating from other generation systems or even to synthetic assets created by digital artists.

\subsection{AYG's Score Distillation in Practice} \label{sec:ayd_score_distill}
Above, we have laid out AYG's general synthesis framework. The full stage 2 score distillation gradient including CFG can be expressed as (stage 1 proceeds analogously)
\begin{align}\label{eq:ayg_score}
& \nabla_{\boldsymbol{\Phi}}\mathcal{L}^{\textrm{AYG}}_{\textrm{SDS}}=\mathbb{E}_{t,\boldsymbol{\epsilon}^{\textrm{vid}},\boldsymbol{\epsilon}^{\textrm{im}}}\biggl[w(t)\biggl\{\gamma\biggl(\omega_{\textrm{vid}}\bigl[\hat{\boldsymbol{\epsilon}}^{\textrm{vid}}(\rmZ,v,t)- \hat{\boldsymbol{\epsilon}}^{\textrm{vid}}(\rmZ,t)\bigr] \nonumber \\
& \quad + \underbrace{\hat{\boldsymbol{\epsilon}}^{\textrm{vid}}(\rmZ,v,t)-\boldsymbol{\epsilon}^{\textrm{vid}}}_{\boldsymbol{\delta}_{\textrm{gen}}^{\textrm{vid}}}\biggr) + \kappa\biggl(\omega_{\textrm{im}}\bigl[\hat{\boldsymbol{\epsilon}}^{\textrm{im}}(\tilde{\rmZ},v,t)-\hat{\boldsymbol{\epsilon}}^{\textrm{im}}(\tilde{\rmZ},t)\bigr] \nonumber \\
& \quad + \underbrace{\hat{\boldsymbol{\epsilon}}^{\textrm{im}}(\tilde{\rmZ},v,t)-\boldsymbol{\epsilon}^{\textrm{im}}}_{\boldsymbol{\delta}_{\textrm{gen}}^{\textrm{im}}}\biggr)
\biggr\}\frac{\partial\{\mathbf{x}\}}{\partial\boldsymbol{\Phi}}\biggr],
\end{align}
where $\rmZ:=\{\rvz_{\tau_i}^{c_i}\}_F$, $\tilde{\rmZ}:=\{\tilde{\rvz}_{\tilde{\tau}_j}^{\tilde{c}_j}\}_M$, $\omega_\textrm{vid/im}$ are the CFG scales for the video and image DMs, $\hat{\boldsymbol{\epsilon}}^{\textrm{vid}}(\rmZ,v,t)$ and $\hat{\boldsymbol{\epsilon}}^{\textrm{im}}(\tilde{\rmZ},v,t)$ are the corresponding denoiser networks and $\boldsymbol{\epsilon}^{\textrm{vid}}$ and $\boldsymbol{\epsilon}^{\textrm{im}}$ are the diffusion noises (an analogous SDS gradient can be written for stage 1). Moreover, $\{\rvx\}$ denotes the set of all renderings from the 4D scene through which the SDS gradient is backpropagated, and which are diffused to produce $\rmZ$ and $\tilde{\rmZ}$. Recently, ProlificDreamer~\cite{wang2023prolificdreamer} proposed a scheme where the control variates $\boldsymbol{\epsilon}^{\textrm{vid/im}}$ above are replaced by DMs that model the rendering distribution, are initialized from the DMs guiding the synthesis ($\hat{\boldsymbol{\epsilon}}^{\textrm{vid}}(\rmZ,v,t)$ and $\hat{\boldsymbol{\epsilon}}^{\textrm{im}}(\tilde{\rmZ},v,t)$ here), and are then slowly fine-tuned on the diffused renderings ($\rmZ$ or $\tilde{\rmZ}$ here). This means that at the beginning of optimization the terms $\boldsymbol{\delta}_{\textrm{gen}}^{\textrm{vid/im}}$ in \Cref{eq:ayg_score} would be zero. Inspired by this observation and aiming to avoid ProlificDreamer's cumbersome fine-tuning, we instead propose to simply set $\boldsymbol{\delta}_{\textrm{gen}}^{\textrm{vid/im}}=0$ entirely and optimize with
\begin{align}\label{eq:ayg_score2}
\nabla_{\boldsymbol{\Phi}}\mathcal{L}^{\textrm{AYG}}_{\textrm{CSD}}&=\mathbb{E}_{t,\boldsymbol{\epsilon}^{\textrm{vid}},\boldsymbol{\epsilon}^{\textrm{im}}}\biggl[w(t)\biggl\{
\omega_{\textrm{vid}}\bigl[\underbrace{\hat{\boldsymbol{\epsilon}}^{\textrm{vid}}(\rmZ,v,t)- \hat{\boldsymbol{\epsilon}}^{\textrm{vid}}(\rmZ,t)}_{\boldsymbol{\delta}_{\textrm{cls}}^{\textrm{vid}}}\bigr] \nonumber \\
& + \omega_{\textrm{im}}\bigl[\underbrace{\hat{\boldsymbol{\epsilon}}^{\textrm{im}}(\tilde{\rmZ},v,t)-\hat{\boldsymbol{\epsilon}}^{\textrm{im}}(\tilde{\rmZ},t)}_{\boldsymbol{\delta}_{\textrm{cls}}^{\textrm{im}}}\bigr]
\biggr\}\frac{\partial\{\mathbf{x}\}}{\partial\boldsymbol{\Phi}}\biggr],
\end{align}
where we absorbed $\gamma$ and $\kappa$ into $\omega_{\textrm{vid/im}}$. Interestingly, this exactly corresponds to the concurrently proposed classifier score distillation (CSD)~\cite{yu2023csd}, which points out that the above two terms $\boldsymbol{\delta}_{\textrm{cls}}^{\textrm{vid/im}}$ in \Cref{eq:ayg_score2} correspond to implicit classifiers predicting $v$ from the video or images, respectively. CSD then uses only $\boldsymbol{\delta}_{\textrm{cls}}^{\textrm{vid/im}}$ for score distillation, resulting in improved performance over SDS. We discovered that scheme independently, while aiming to inherit ProlificDreamer's strong performance. Supp. Material for details.

\input{figures_tex/meta_comparison}
\subsection{Scaling Align Your Gaussians} \label{sec:scaling_ayg}
To scale AYG and achieve state-of-the-art text-to-4D performance, we introduce several further novel techniques.

\textbf{Distribution Regularization of 4D Gaussians.}
We developed a method to stabilize optimization and ensure realistic learnt motion. We calculate the means $\boldsymbol{\nu}_\tau$ and diagonal covariances $\boldsymbol{\Gamma}_\tau$ of the entire set of 3D Gaussians (using their means $\boldsymbol{\mu}_i$) at times $\tau$ along the 4D sequence (\Cref{fig:jsd_reg_figure}). Defining a Normal distribution $\mathcal{N}(\boldsymbol{\nu}_\tau,\boldsymbol{\Gamma}_\tau)$ with these means and covariances, we regularize with a modified version of the Jensen-Shannon divergence
$\textrm{JSD}\left(\mathcal{N}(\boldsymbol{\nu}_0,\boldsymbol{\Gamma}_0)||\mathcal{N}(\boldsymbol{\nu}_\tau,\boldsymbol{\Gamma}_\tau)\right)$
between the 3D Gaussians at the initial and later frames $\tau$ (see Supp. Material). This ensures that the mean and the diagonal covariance of the distribution of the Gaussians stay approximately constant and encourages AYG to generate meaningful and complex dynamics instead of simple global translations and object size changes.

\textbf{Extended Autoregressive Generation.}
By default, AYG produces relatively short 4D sequences, which is due to the guiding text-to-video model, which itself only generates short video clips (see \citet{blattmann2023videoldm}). To overcome this limitation, we developed a method to autoregressively extend the 4D sequences. We use the middle 4D frame from a first sequence as the initial frame of a second sequence, optimizing a second deformation field, optionally using a different text prompt. 
As the second sequence is initialized from the middle frame of the first sequence,
there is an overlap interval with length $0.5$ of the total length of each sequence.
When optimizing for the second deformation field, we smoothly interpolate between the first and second deformation fields for the overlap region (\Cref{fig:autoreg_concept}).
Specifically, we define
$\boldsymbol{\Delta}^{\textrm{interpol}}_{\boldsymbol{\Phi}_{12}} = (1-\chi(\tau))\boldsymbol{\Delta}_{\boldsymbol{\Phi}_1} + \chi(\tau) \boldsymbol{\Delta}_{\boldsymbol{\Phi}_2}$
where $\chi$ is a linear function with $\chi(\tau_{0.5}) = 0 $ and $\chi(\tau_{1.0}) = 1$, $\tau_{0.5}$ and $\tau_{1.0}$ represent the middle and last time frames of the first sequence,
$\boldsymbol{\Delta}^{\textrm{interpol}}_{\boldsymbol{\Phi}_{12}}$ is the interpolated deformation field, and $\boldsymbol{\Delta}_{\boldsymbol{\Phi}_1}$(kept fixed) and $\boldsymbol{\Delta}_{\boldsymbol{\Phi}_2}$ are the deformation fields of the first and second sequence, respectively.
We additionally minimize $\mathcal{L}_\textrm{Interpol-Reg.} = ||\boldsymbol{\Delta}_{\boldsymbol{\Phi}_1}-\boldsymbol{\Delta}^{\textrm{interpol}}_{\boldsymbol{\Phi}_{12}}||^2_2$ within the overlap region to regularize the optimization process of $\boldsymbol{\Delta}_{\boldsymbol{\Phi}_2}$.
For the non-overlap regions, we just use the corresponding $\boldsymbol{\Delta}_{\boldsymbol{\Phi}}$.
With this careful interpolation technique the deformation field smoothly transitions from the first sequence's into the second sequence's. Without it, we obtained abrupt, unrealistic transitions.

\input{tables_tex/meta_comparison}
\input{tables_tex/ablation_results}
\textbf{Motion Amplification.}
When a set of 4D scene renderings is given to the text-to-video model, it produces a (classifier) score distillation gradient for each frame $i$. We expect most motion when the gradient for each frame points into a different direction. With that in mind, we propose a motion amplification technique. We post-process the video model's individual frame scores ${\boldsymbol{\delta}_{\textrm{cls}}^{\textrm{vid}}}_i$ ($i\in\{1,...,F\}$) as
${\boldsymbol{\delta}_{\textrm{cls}}^{\textrm{vid}}}_i \rightarrow \left<{\boldsymbol{\delta}_{\textrm{cls}}^{\textrm{vid}}}_i\right> + \omega_\textrm{ma}\left({\boldsymbol{\delta}_{\textrm{cls}}^{\textrm{vid}}}_i-\left<{\boldsymbol{\delta}_{\textrm{cls}}^{\textrm{vid}}}_i\right>\right)$,
where $\left<{\boldsymbol{\delta}_{\textrm{cls}}^{\textrm{vid}}}_i\right>$
is the average score over the $F$ video frames and $\omega_{\textrm{ma}}$ is the motion amplifier scale. This scheme is inspired by CFG and reproduces regular video model scores for $\omega_{\textrm{ma}}{=}1$. For larger $\omega_{\textrm{ma}}$, the difference between the individual frames' scores and the average is amplified, thereby encouraging larger frame differences and more motion.

\textbf{View Guidance.}
In AYG's 3D stage, for the text-to-image model we use a new \textit{view guidance}. We construct an additional implicit classifier term $\omega_{\textrm{vg}}\bigl[\hat{\boldsymbol{\epsilon}}^{\textrm{im}}(\rvz,v^{\textrm{aug}},t)-\hat{\boldsymbol{\epsilon}}^{\textrm{im}}(\rvz,v,t)\bigr]$, where $v^{\textrm{aug}}$ denotes the original text prompt $v$ augmented with directional texts such as ``front view'' (see \Cref{sec:compo_gen}) and $\omega_{\textrm{vg}}$ is the guidance scale. View guidance amplifies the effect of directional text prompt augmentation.

\textbf{Negative Prompting.} We also use negative prompt guidance during both the 3D and 4D stages. During the 4D stage, we use \textit{``low motion, static statue, not moving, no motion''} to encourage AYG to generate more dynamic and vivid 4D scenes. Supp. Material for 3D stage and details.

%% file: figures_tex/jsd_reg_figure.tex
\begin{figure}[t!]
  \vspace{-0.5cm}
  \begin{minipage}[c]{0.18\textwidth}
    \includegraphics[width=0.9\textwidth]{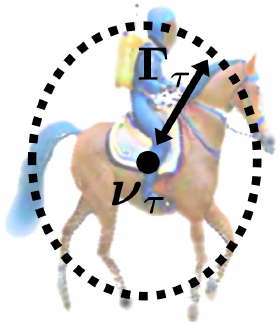}
  \end{minipage}\begin{minipage}[c]{0.295\textwidth}
  \vspace{3mm}
    \caption{\small \textbf{AYG's JSD-based regularization} of the evolving 4D Gaussians (see \Cref{sec:scaling_ayg}) calculates the 3D mean $\boldsymbol{\nu}_\tau$ and diagonal covariance matrix $\boldsymbol{\Gamma}_\tau$ of the set of dynamic 3D Gaussians at different times $\tau$ of the 4D sequence and regularizes them to not vary too much.}
    \label{fig:jsd_reg_figure}
  \end{minipage}
  \vspace{-7mm}
\end{figure}

%% file: figures_tex/autoreg_concept_figure.tex
\begin{figure}[t!]
  \vspace{-0.4cm}
    \begin{center}
    \includegraphics[width=0.44\textwidth]{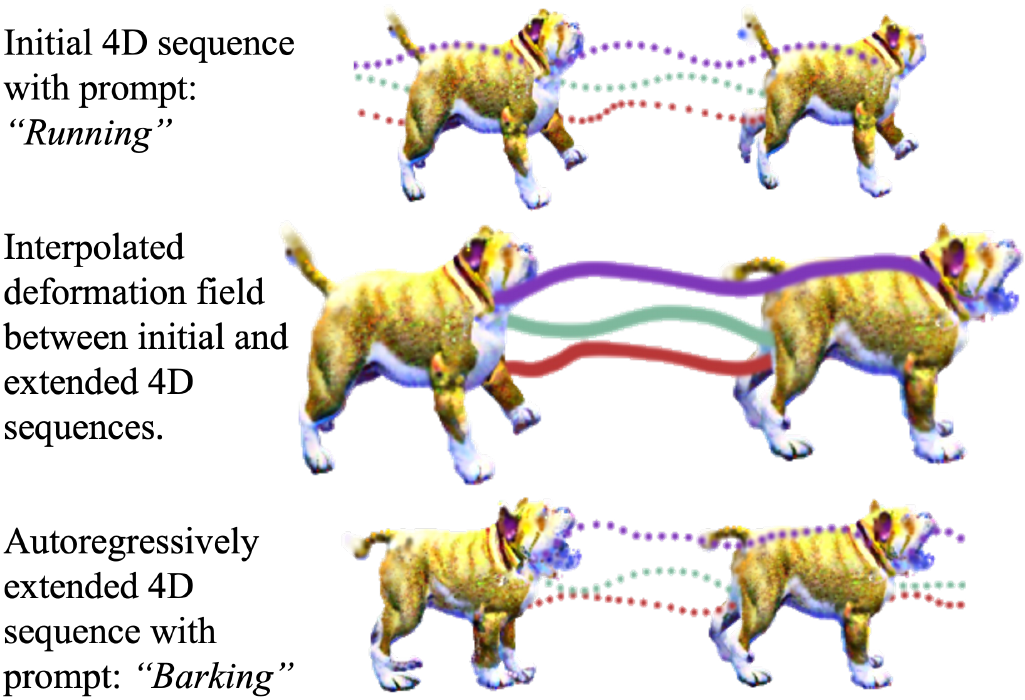}
    \end{center}
    \vspace{-0.5cm}
    \caption{\small \textbf{AYG's autoregressive extension scheme} interpolates the deformation fields of an initial and an extended 4D sequence within an overlap interval between the two sequences (\Cref{sec:scaling_ayg}).}
    \label{fig:autoreg_concept}
  \vspace{-5mm}
\end{figure}

%% file: figures_tex/main_results_figure.tex
\begin{figure*}[t!]
  \vspace{-0.4cm}
    \includegraphics[width=\textwidth]{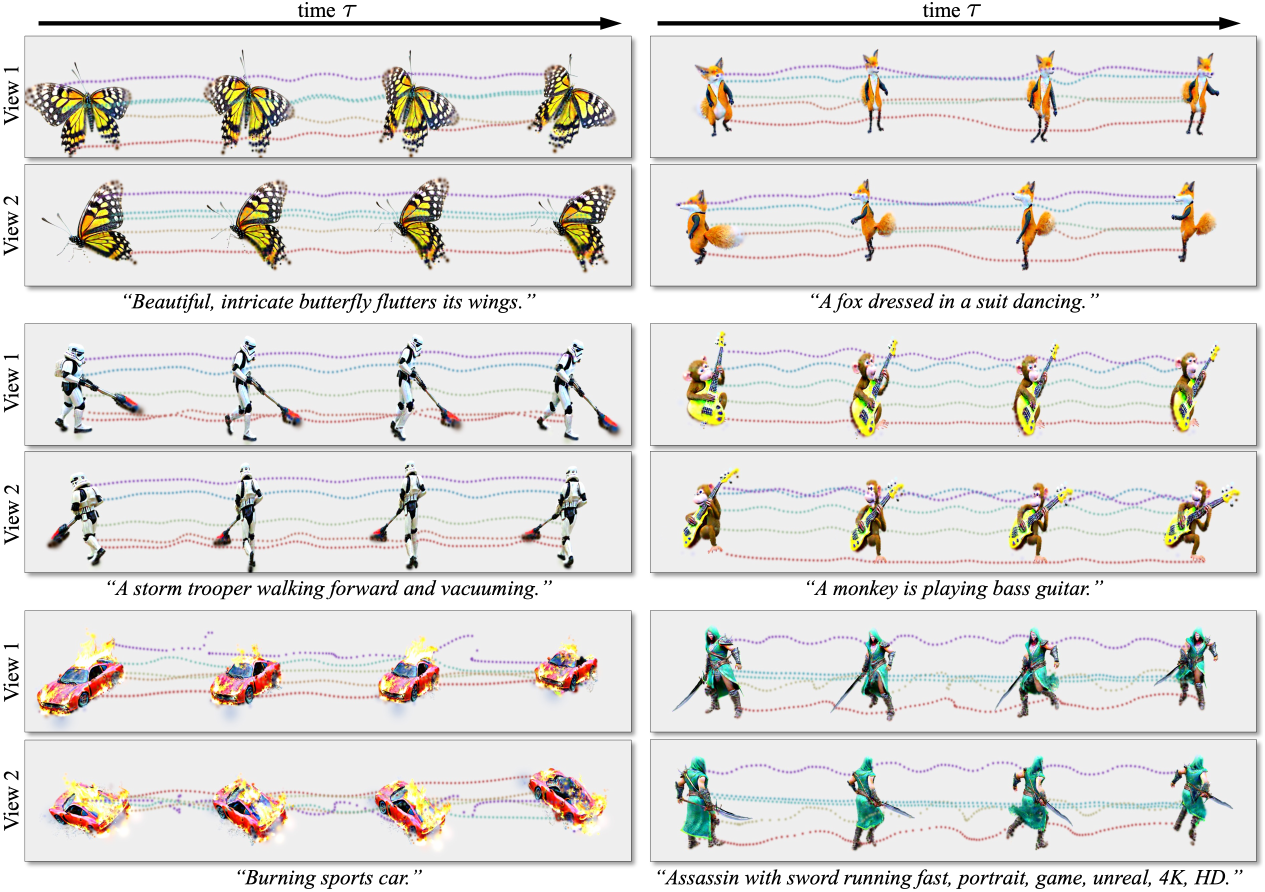}
    \vspace{-6mm}
    \caption{\small \textbf{Text-to-4D synthesis with AYG.} Various samples shown in two views each. Dotted lines denote deformation field dynamics.}
    \label{fig:main_results}
  \vspace{-5mm}
\end{figure*}

%% file: figures_tex/autoreg_loopy_results.tex
\begin{figure}[t!]
  \vspace{-0.4cm}
    \begin{center}
    \includegraphics[width=0.47\textwidth]{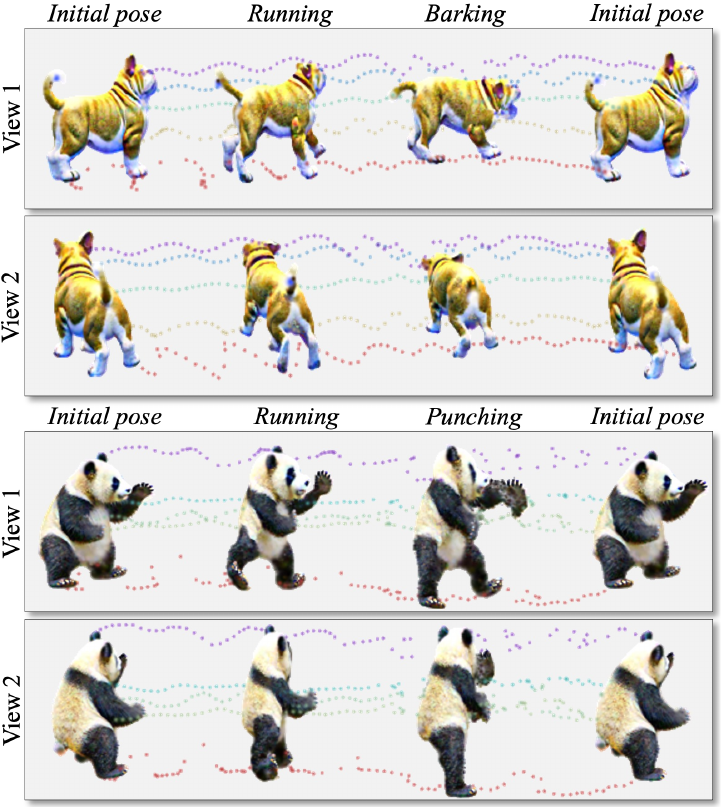}
    \end{center}
    \vspace{-0.7cm}
    \caption{\small \textbf{Autoregressively extended text-to-4D synthesis.} AYG is able to autoregressively extend dynamic 4D sequences, combine sequences with different text-guidance, and create looping animations, returning to the initial pose (also see Supp. Video).}
    \label{fig:autoreg_loopy_results}
  \vspace{-5mm}
\end{figure}

%% file: figures_tex/meta_comparison.tex
\begin{figure}[t!]
  \vspace{-0.4cm}
    \begin{center}
    \includegraphics[width=0.47\textwidth]{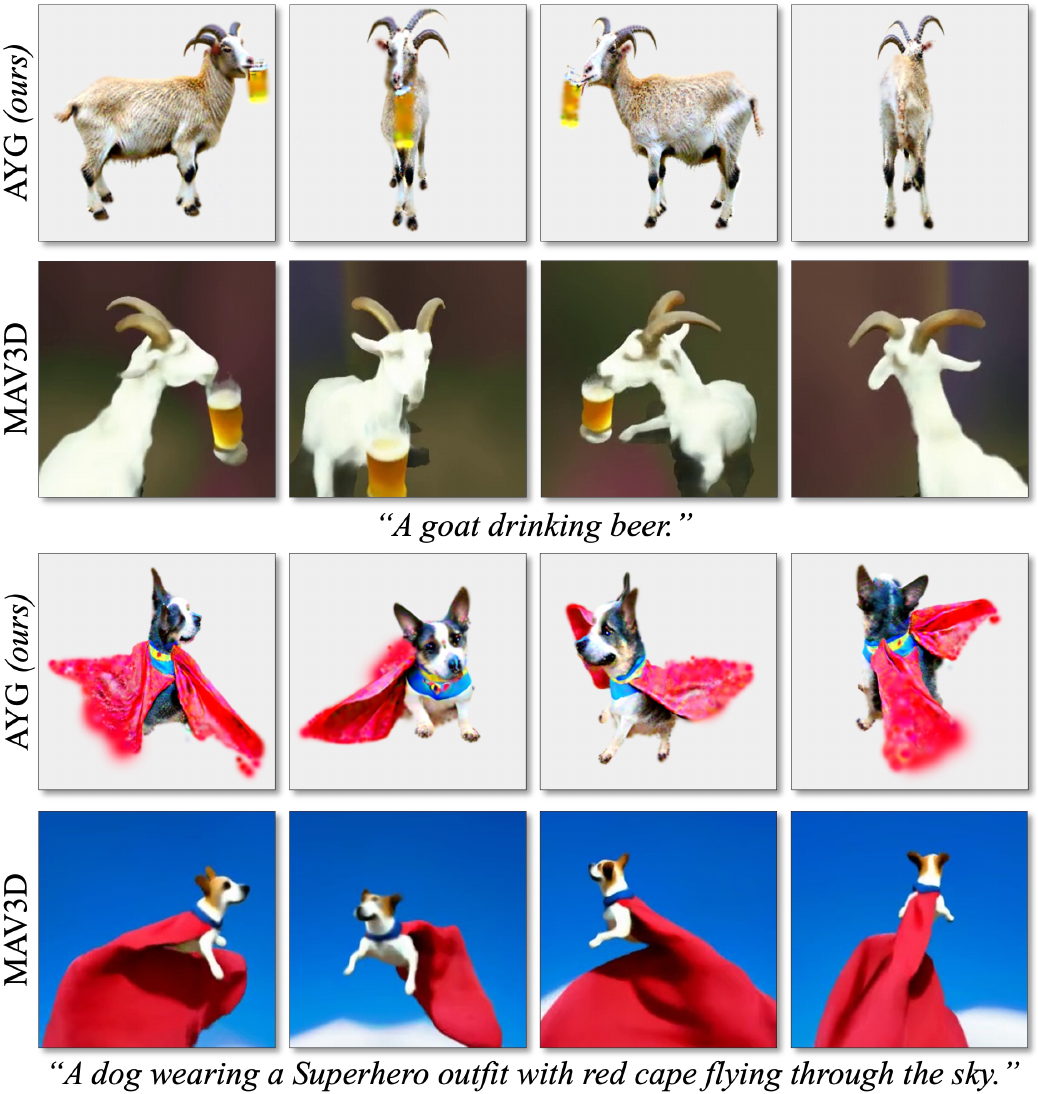}
    \end{center}
    \vspace{-0.6cm}
    \caption{\small \textbf{AYG (\textit{ours}) vs. MAV3D~\cite{singer2023mav3d}.} We show four 4D frames for different times and camera angles (also see Supp. Video).}
    \label{fig:meta_comparison}
  \vspace{-5mm}
\end{figure}

%% file: tables_tex/meta_comparison.tex
\begin{table}
\vspace{-4mm}
    \centering
        \caption{\small \textbf{Comparison to MAV3D~\cite{singer2023mav3d}} by user study on synthesized 4D scenes with 28 text prompts. Numbers are percentages.\vspace{-1em}}
    \label{tab:user_study_mav_comparison}
    \resizebox{ \linewidth}{!}{\rowcolors{2}{gray!15}{white}
    \begin{tabular}{l  c c c}
        \toprule
         Method &   AYG \textit{(ours)} &  MAV3D~\cite{singer2023mav3d}  & Equal  \\
        preference &   preferred &  preferred  &  preference \\
        \midrule
        Overall Quality & \textbf{53.6} & 38.8 & 7.6  \\
        3D Appearance & \textbf{47.4} & 37.2 & 15.4  \\
        3D Text Alignment & \textbf{45.9} & 38.8 & 15.3  \\
        Motion Amount & \textbf{45.9} & 38.8 & 15.3  \\
        Motion Text Alignment & \textbf{47.4} & 33.7 & 18.9  \\
        Motion Realism & \textbf{44.4} & 43.9 & 11.7  \\ \rowcolor{white}
        \bottomrule
        \vspace{-1.5em}
    \end{tabular}
    }
    \vspace{-5mm}
\end{table}

%% file: tables_tex/ablation_results.tex
\begin{table*}[h!]
\vspace{-4mm}
\begin{minipage}[c]{0.74\textwidth}
    \centering
    \scalebox{0.74}{\rowcolors{2}{gray!15}{white}
    \begin{tabular}{l  c c c c c c}
        \toprule
        \rowcolor{white}
        Align Your Gaussians  &   Overall &  3D  & 3D Text & Motion  & Motion Text  & Motion  \\
        \rowcolor{white}
        (full model) &   Quality & Appearance  & Alignment &  Amount & Alignment &  Realism \\
        \midrule

         v.s. w/o rigidity regularization & \textbf{45.8}/13.3 & \textbf{43.3}/19.2  & \textbf{38.3}/15.0  & \textbf{40.8}/15.0 & \textbf{42.5}/18.3 & \textbf{30.8}/26.7 \\ 
         v.s. w/o motion amplifier    &  \textbf{43.3}/23.3 & \textbf{37.5}/28.3  & \textbf{30.8}/26.7  & \textbf{45.8}/10.8 & \textbf{37.5}/26.7 & \textbf{33.3}/31.7 \\

          v.s. w/o initial 3D stage    &  \textbf{67.5}/15.0 & \textbf{57.5}/21.7  & \textbf{64.2}/15.0  & \textbf{60.8}/21.7 & \textbf{60.8}/20.8 & \textbf{59.2}/24.2 \\

           v.s. w/o JSD-based regularization    &  \textbf{40.0}/25.0 & \textbf{40.0}/27.5  & \textbf{36.7}/27.5  & \textbf{41.7}/24.2 & \textbf{39.2}/29.2 & \textbf{45.0}/24.2 \\

           v.s. w/o image DM score in 4D stage    &  \textbf{42.5}/22.5 & \textbf{39.2}/27.5  & \textbf{36.7}/25.8  & \textbf{33.3}/25.9 & \textbf{37.5}/30.0 & 27.5/\textbf{40.0} \\

          v.s. SDS instead of CSD    &  \textbf{44.2}/35.8 & \textbf{40.0}/27.5  & \textbf{35.8}/35.0  & \textbf{35.0}/27.5 & \textbf{35.0}/34.2 & 32.5/\textbf{35.8} \\

        v.s. 3D stage w/o MVDream    &  \textbf{66.7}/21.7 & \textbf{48.3}/34.2  & \textbf{38.3}/34.2  & \textbf{41.7}/22.5 &  \textbf{40.0}/27.5 & \textbf{40.8}/27.5 \\

        v.s. 4D stage with MVDream    &  \textbf{50.8}/27.5 &  \textbf{38.3}/34.2  &  \textbf{41.6}/29.2  & \textbf{39.2}/35.0 &  \textbf{44.2}/30.0 & \textbf{39.2}/31.7 \\

        v.s. video model with only fps 4   &  \textbf{46.7}/15.8 & 27.5/\textbf{36.7} & \textbf{30.0}/23.3  & \textbf{36.7}/30.0 & \textbf{31.7}/26.7 & \textbf{32.5}/28.3 \\

        v.s. video model with only fps 12    &  \textbf{48.3}/29.2 &  \textbf{30.8}/29.2 & \textbf{29.2}/28.3  & \textbf{35.0}/28.3 & \textbf{35.0}/30.0 & \textbf{39.2}/26.7 \\

        v.s. w/o dynamic cameras    &  \textbf{32.5}/25.0 &  \textbf{32.5}/31.7 & \textbf{35.0}/33.3  & \textbf{35.0}/32.5 &\textbf{ 35.8}/33.3 &  \textbf{32.5}/25.0 \\

         v.s. w/o negative prompting    &  \textbf{44.2}/28.3 &  \textbf{38.3}/32.5 & \textbf{31.7}/29.2  & 29.2/\textbf{31.6} & \textbf{33.3}/30.0 &  \textbf{37.5}/28.3 \\ \rowcolor{white}
        \bottomrule
        \vspace{-1.5em}
    \end{tabular}
    }
\end{minipage}\hfill
\begin{minipage}[c]{0.26\textwidth}
    \vspace{-4mm}
        \caption{\small \textbf{Ablation study} by user study on synthesized 4D scenes with 30 text prompts. For each pair of numbers, the left number is the percentage that the full AYG model is preferred and the right number indicates preference percentage for ablated model as described in left column. The numbers do not add up to $100$ and the difference is due to users voting ``no preference'' (details in Supp. Material). \vspace{-1em}}
    \label{tab:user_study_ablations}
    \vspace{-5mm}
\end{minipage}
\vspace{-4mm}
\end{table*}

%% file: sections/4_experiments.tex
\vspace{-1mm}
\section{Experiments}\label{sec:experiments}
\vspace{-1mm}
\textbf{Text-to-4D.} In \Cref{fig:main_results}, we show text-to-4D sequences generated by AYG (hyperparameters and details in Supp. Material). AYG can generate realistic, expressive, detailed and vivid dynamic 4D scenes (4D scenes can be rendered at varying speeds and frame rates). Importantly, our method demonstrates zero-shot generalization capabilities to creative text prompts corresponding to scenes that are unlikely to be found in the diffusion models' training images and videos. More results in Supp. Material and on project page.

To compare AYG to MAV3D~\cite{singer2023mav3d}, we performed a comprehensive user study where we took the 28 rendered videos from MAV3D's project page\footnote{\url{https://make-a-video3d.github.io/}} and compared them to corresponding generations from AYG with the same text prompts (\Cref{tab:user_study_mav_comparison}). We asked the users to rate overall quality, 3D appearance and text alignment, as well as motion amount, motion text alignment and motion realism (user study details in Supp. Material). AYG outperforms MAV3D on all metrics, achieving state-of-the-art text-to-4D performance (we also evaluated R-Precision~\cite{park2021benchmark,jain2022dreamfields} on a larger prompt set used by MAV3D~\cite{singer2023mav3d,singer2023makeavideo}, performing on par, see Supp. Mat.; however, R-Precision is a meaningless metric to evaluate \textit{dynamic} scenes). Qualitative comparisons are shown in \Cref{fig:meta_comparison} (more in Supp. Mat.). We see that AYG produces more detailed 4D outputs. Note that MAV3D uses an extra background model, while AYG does not. Adding a similar background model would be easy but is left to future work.

\textbf{Ablation Studies.} Next, we performed an ablation study on AYG's different components. We used a set of 30 text prompts and generated 4D scenes for versions of AYG with missing or modified components, see \Cref{tab:user_study_ablations}. Using the same categories as before, we asked users to rate preference of our full method vs. the ablated AYG variants. Some components have different effects with respect to 3D appearance and motion, but we generally see that all components matter significantly in terms of overall quality, \ie, for all ablations our full method is strongly preferred over the ablated AYG versions. This justifies AYG's design. A thorough discussion is presented in the Supp. Material, but we highlight some relevant observations. We see that our novel JSD-based regularization makes a major difference, and we also observe that the motion amplifier indeed has a strong effect for ``Motion Amount''. Moreover, our compositional approach is crucial. Running the 4D stage without image DM feedback produces much worse 3D and overall quality. Also the decomposition into two stages is important---carrying out 4D synthesis without initial 3D stage performs poorly.\looseness=-1

\textbf{Temporally Extended 4D Synthesis and Large Scene Composition.} In \Cref{fig:autoreg_loopy_results}, we show autoregressively extended text-to-4D results with changing text prompts (also see Supp. Video). AYG can realistically connect different 4D sequences and generate expressive animations with changing dynamics and behavior. We can also create sequences that loop endlessly by enforcing that the last frame of a later sequence matches the first frame of an earlier one and suppressing the deformation field there (similar to how we enforce zero deformation at $\tau{=}0$ in \Cref{sec:ayg_4drep}).
Finally, due to the explicit nature of the dynamic 3D Gaussians, AYG's 4D representation, multiple animated 4D objects can be easily composed into larger scenes, each shape with its own deformation field defining its dynamics. We show this in \Cref{fig:teaser}, where each dynamic object in the large scene is generated, except for the ground plane. These capabilities, not shown by previous work~\cite{singer2023mav3d}, are particularly promising for practical content creation applications.

%% file: sections/5_conclusions.tex
\vspace{-2mm}
\section{Conclusions}\label{sec:conclusions}
\vspace{-1mm}
We presented \textit{Align Your Gaussians} for expressive text-to-4D synthesis. AYG builds on dynamic 3D Gaussian Splatting with deformation fields as well as score distillation with multiple composed diffusion models. Novel regularization and guidance techniques allow us to achieve state-of-the-art dynamic scene generation and we also show temporally extended 4D synthesis as well as the composition of multiple dynamic objects within a larger scene. AYG has many potential applications for creative content creation and it could also be used in the context of synthetic data generation. For example, AYG would enable synthesis of videos and 4D sequences with exact tracking labels, useful for training discriminative models. AYG currently cannot easily produce topological changes of the dynamic objects. Overcoming this limitation would be an exciting avenue for future work. Other directions include scaling AYG beyond object-centric generation and personalized 4D synthesis. The initial 3D object could be generated from a personalized diffusion model (\textit{e.g.} DreamBooth3D~\cite{ruiz2023dreambooth,raj2023dreambooth3d}) or with image-to-3D methods~\cite{liu2023zero123,qian2023magic123,liu2023syncdreamer,hong2023lrm,liu2023one2345} and then animated with AYG.%

%% file: appendices/overview_supp_videos.tex
\section{Supplementary Videos} \label{app:supp_videos}
For fully rendered videos, we primarily refer the reader to our project page, \url{https://research.nvidia.com/labs/toronto-ai/AlignYourGaussians/}, which shows all our results in best quality.

Moreover, we include 3 videos in the following google drive folder: \url{https://drive.google.com/drive/folders/1I7e6aj-7BBrIVdePyHEQDGUxLEEls3-e}:
\begin{itemize}
\setlength\itemsep{0.4em}
    \item \texttt{ayg$\_$text$\_$to$\_$4d.mp4}: This video contains many text-to-4D generation results, comparisons to MAV3D~\cite{singer2023mav3d} and autoregressively extended and looping 4D sequences. Moreover, we show how we compose different 4D sequences into a large scene as in \Cref{fig:teaser}. 
    \item \texttt{ayg$\_$ablation$\_$study.mp4}: This video contains different generated text-to-4D sequences for our ablation study, comparing our main Align Your Gaussians model with the different modified versions in the ablations. Note that these ablation results only use 4,000 optimization steps in the second dynamic 4D optimization stage for efficiency (see \Cref{supp:extend_discussion_ablations}).
    \item \texttt{ayg$\_$new$\_$video$\_$model.mp4}: This video shows generated 2D video samples for different fps conditionings from our newly trained latent video diffusion model for this project.
\end{itemize}

\noindent Note that all videos shown on the project page leverage an additional fine-tuning stage with additional optimization steps compared to the results shown in the paper and in \texttt{ayg$\_$text$\_$to$\_$4d.mp4}. See \Cref{supp:extra_finetuning} for a discussion on additional fine-tuning and optimization.

%% file: appendices/realted_work_extended.tex
\section{Related Work---Extended Version}\label{sec:related_extended}
\textbf{Diffusion Models.} Align Your Gaussians leverages multiple different diffusion models (DMs)~\cite{ho2020ddpm,song2020score,sohl2015deep}. DMs have revolutionized deep generative modeling in the visual domain, most prominently for image synthesis~\cite{dhariwal2021diffusion,nichol2021improved,schwarz2023wildfusion,rombach2021highresolution,hoogeboom2023simplediffusion}. They leverage a forward diffusion process that gradually perturbs input data towards entirely random noise, while a denoiser neural network is learnt to reconstruct the input. Specifically, AYG builds on large-scale text-to-image~\cite{saharia2022photorealistic,rombach2021highresolution,ramesh2022dalle2,balaji2022ediffi,feng2023ernievilg,podell2023sdxl,dai2023emu,xue2023raphael,gu2023matryoshka}, text-to-video~\cite{blattmann2023videoldm,singer2023makeavideo,ho2022imagen,zhou2023magicvideo,wang2023videofactory,wang2023lavie,an2023latentshift,ge2022pyoco,wu2023tune,khachatryan2023text2videozero,guo2023animatediff,girdhar2023emu} and text-to-multiview-image DMs~\cite{liu2023zero123,qian2023magic123,liu2023syncdreamer,shi2023mvdream,shi2023zero123plus,liu2023one2345,xu2023dmv3d,long2023wonder3d}. A popular framework for efficient yet expressive generative modeling with DMs is the \textit{latent} DM approach, where data is mapped into a compressed latent space with an autoencoder and the DM is trained in a more efficient manner in this latent space~\cite{vahdat2021score,rombach2021highresolution}. The most prominent model of this type is Stable Diffusion~\cite{rombach2021highresolution}, and AYG uses exclusively latent DMs which are based on Stable Diffusion. Specifically, apart from Stable Diffusion 2.1, we retrain a VideoLDM~\cite{blattmann2023videoldm} for fps-conditioned text-to-video synthesis, which starts from a Stable Diffusion image generator as base model. Moreover, AYG uses MVDream~\cite{shi2023mvdream}, a text-guided multiview latent DM similarly fine-tuned from Stable Diffusion.

\textbf{Text-to-3D Generation with Score Distillation.} Text-conditioned image DMs are often trained on large-scale datasets consisting of hundreds of millions or billions of text-image pairs. However, such huge text-annotated datasets are not available for 3D data. While there exists a rich literature on 3D DMs trained on small explicit 3D or multiview image datasets~\cite{nichol2022pointe,zeng2022lion,kim2023nfldm,wang2023rodin,zhou2021pvd,luo2021diffusion,liu2023MeshDiffusion,kalischek2022tetrahedral,shue2023triplanediffusion,bautista2022gaudi}, the lack of large text-annotated 3D training datasets is a challenge for 3D generative modeling. Aiming to overcome these limitations, score distillation methods, introduced in the seminal work by \citet{poole2023dreamfusion}, use large-scale text-guided 2D diffusion models to distill 3D objects in a per-instance optimization process. A 3D scene, parametrized by learnable parameters, is rendered from different camera perspectives and the renderings are given to a 2D diffusion model, which can provide gradients to make the renderings look more realistic. These gradients can be backpropagated through the rendering process and used to update the 3D scene representation. This has now become a flourishing research direction for 3D generative modeling~\cite{lin2023magic3d,chen2023fantasia3d,wang2023score,tsalicoglou2024textmesh,zhu2023hifa,huang2023dreamtime,chen2023it3d,wang2023prolificdreamer,metzer2023latent,deng2023nerdi,xu2023neuralLift,lorraine2023att3d,feng2023metadreamer,sun2023dreamcraft3d,katzir2023noisefree,liang2023luciddreamer}. AYG also builds on the score distillation framework.

\textbf{Text-to-4D Generation.} The vast majority of papers on score distillation tackles static 3D object synthesis. To the best of our knowledge, there is only one previous paper that leverages score distillation for text-guided generation of dynamic 4D scenes, \textit{Make-A-Video3D (MAV3D)}~\cite{singer2023mav3d}. Hence, this is the most related work to AYG. However, MAV3D uses neural radiance fields~\cite{mildenhall2020nerf} with HexPlane~\cite{Cao2022hexplane} features as 4D representation, in contrast to AYG's dynamic 3D Gaussians, and it does not disentangle its 4D representation into a static 3D representation and a deformation field modeling dynamics. MAV3D's representation prevents it from composing multiple 4D objects into large dynamic scenes, which our 3D Gaussian plus deformation field representation easily enables, as we show. Moreover, MAV3D's sequences are limited in time, while we show a novel autoregressive generation scheme to extend our 4D sequences. AYG outperforms MAV3D qualitatively and quantitatively and synthesizes significantly higher-quality 4D scenes. Our novel compositional generation-based approach contributes to this, which MAV3D does not pursue. More specifically, in contrast to MAV3D, our AYG shows how to simultaneously leverage image and video diffusion models for improved synthesis in the 4D generation stage, and moreover leverages a 3D-aware multiview image diffusion model for improved 3D generation in the initial stage. Finally, instead of regular score distillation sampling~\cite{poole2023dreamfusion}, used by MAV3D, in practice AYG employs classifier score distillation~\cite{yu2023csd}. Note that MAV3D did not release any code or model checkpoints and its 4D score distillation leverages the large-scale Make-A-Video~\cite{singer2023makeavideo} text-to-video diffusion model, which also is not available publicly.

\textbf{3D Gaussian Splatting and Deformation Fields.} AYG leverages a 3D Gaussian Splatting-based 4D representation with deformation fields to model dynamics. 3D Gaussian Splatting~\cite{kerbl20233Dgaussians} has been introduced as an efficient 3D scene representation and, concurrently with our work, has also been employed for text-to-3D generation by DreamGaussian~\cite{tang2023dreamgaussian}, GSGEN~\cite{chen2023gsgen}, GaussianDreamer~\cite{yi2023gaussiandreamer}, LucidDreamer~\cite{liang2023luciddreamer} and another work also called LucidDreamer~\cite{chung2023luciddreamer}. However, these works synthesize only static 3D scenes, but do not consider dynamics. Deformation fields have been widely used for dynamic 3D scene reconstruction~\cite{pumarola2020d,park2021nerfies,Cai2022NDR,tretschk2021nonrigid,park2021hypernerf,mihajlovic2023ResFields,jiang2023consistent4d}. Concurrently with our work, also several papers on dynamic 3D Gaussian Splatting emerged~\cite{luiten2023dynamic,yang2023deformable,wu20234d,zielonka2023drivable,xie2023physgaussian}, similarly tackling the dynamic 3D scene reconstruction task. However, none of these works address generative modeling.

\textbf{Compositional Generation.} A crucial part of AYG is that it combines multiple diffusion models when performing score distillation. Specifically, in the 4D stage we combine a text-to-image with a text-to-video diffusion model in the form of a product distribution between the two and we then leverage the gradients of this product distribution for score distillation. This can be viewed as a form of compositional generation. Usually, in compositional generation, different diffusion models, or the same diffusion model with different text-conditionings, are combined such that multiple conditions are fulfilled during generation. Formally, this can be achieved by forming the product distribution of different models, see, for instance, \citet{liu2022compositional} and \citet{du2023reduce}. This has been used for controllable image generation. Analogously, we perform a form of compositional generation by composing an image and a video diffusion model to generate dynamic 4D assets whose renderings obey both the text-to-image and text-to-video model distributions simultaneously. Instead of aiming for controllability our goal is to simultaneously enforce smooth temporal dynamics (video model) and high individual frame quality (image model).

\textbf{Animation and Motion Synthesis.} AYG is also broadly related to the rich literature on character animation and motion synthesis with deep generative models; see, for example, the recent works by \citet{rempeluo2023tracepace} and \citet{yuan2023physdiff}. However, this line of work usually only considers the generation of joint configurations of human skeletons or similar low-dimensional simplified representations. An interesting direction for future work would be to combine these types of models with a method like our AYG. For instance, one could first synthesize a rough motion trajectory or motion path, and AYG could then generate a detailed character animation following this trajectory. There is also work on reconstructing animatable 3D assets directly from videos~\cite{yang2022banmo}. It would be interesting to extend AYG towards extracting assets from the synthesized 4D scenes which are simulation-ready and fully animatable using standard graphics software. Finally, AYG is also related to the broader literature on computer animation using classical methods without deep learning; see, for instance, \citet{kerlow2009book}.

\textbf{Further Concurrent Work.} Concurrently with us, Dream-in-4D~\cite{zheng2023unified} also developed a new text-to-4D synthesis scheme. In contrast to AYG, the work leverages a NeRF-based 4D representation and focuses on image-guided and personalized generation, instead of using dynamic 3D Gaussians and targeting temporally extended generation as well as the possibility to easily compose different 4D assets in large scenes (one of the goals of AYG). Hence, their direction is complementary to ours, and it would be interesting to also extend AYG to personalized and image-guided 4D generation. The image-to-4D task is also tackled by Animate124~\cite{zhao2023animate124}, which similarly leverages a dynamic NeRF-based representation. Moreover, 4D-fy~\cite{bahmani20234dfy} also addresses text-to-4D synthesis, combining supervision signals from image, video and 3D-aware diffusion models. However, like Dream-in-4D and Animate124, the work leverages a NeRF-based representation with a multi-resolution feature grid in contrast to AYG's dynamic 3D Gaussians. Moreover, 4D-fy does not explicitly disentangle shape and dynamics into a 3D component and a deformation field. Note that, in contrast to AYG, none of the mentioned works demonstrated temporally extended generation, the possibility to change the text prompt during synthesis or the ability to easily compose multiple 4D assets in large scenes.

%% file: appendices/extended_4d_rep_details.tex
\section{Details of Align Your Gaussians' 4D Representation and Optimization}

\subsection{3D Representation}
\label{sec:supp_3d_representation}
We initialize each scene with 1,000 3D Gaussians spread across a sphere with radius 0.3 located on the origin $(0,0,0)$.
Each Gaussian is initialized with a random RGB color and spherical harmonics coefficients of 0. 
As we use isotropic covariances, we only make use of a single scaling parameter per Gaussian, such that Gaussians are spherical. We follow \citet{kerbl20233Dgaussians} to initialize the scaling and opacity parameters. 
Similar to \citet{kerbl20233Dgaussians}, we add and delete Gaussians during the initial 3D optimization steps to densify regions requiring more detail and prune Gaussians that are not used for rendering. 
Gaussians with an average magnitude of position gradients above 0.002 are densified, and Gaussians with opacity less than 0.005 are pruned every 1,000 steps starting from the 500-th step. 
Additionally, we limit the number of Gaussians such that we only perform the densification step if the total number of Gaussians is less than 50,000, and we also reset the opacities to have the maximum value of 0.01 (after sigmoid) every 3,000 steps.

\subsection{Deformation Field}
After the 3D Gaussians are optimized into a 3D scene in the first stage, the scene dynamics are modeled by a deformation field $\boldsymbol{\Delta}_{\boldsymbol{\Phi}}(x,y,z,\tau)=(\Delta x,\Delta y,\Delta z)$, defined through a multi-layer perceptron (MLP) with parameters $\boldsymbol{\Phi}$.
We encode $(x,y,z,\tau)$ with the sinusoidal positional encoding~\cite{vaswani2017attention}. We use 4 frequencies (each with sine and cosine components), resulting in a 32-dimensional input.
The MLP with parameters $\boldsymbol{\Phi}$ consists of linear layers with ReLU~\cite{nair2010rectified} activation functions.
We include Layer Normalization~\cite{ba2016layer} before the ReLU every second layer, as we found it helped stabilize optimization.
The last layer of $\boldsymbol{\Phi}$ contains a linear layer that is initialized with zero weights, followed by a  soft clamping layer ($f(x) = \text{tanh}(x/0.5)*0.5$) so that it produces the 3-dimensional output $(\Delta x,\Delta y,\Delta z)$ clamped between $(-0.5, 0.5)$.
Furthermore, we preserve the initial 3D Gaussians for the first frame, \ie $\boldsymbol{\Delta}_{\boldsymbol{\Phi}}(x,y,z,0) = (0,0,0)$, by multipying 
the output with $\xi(\tau)$, and the final output is given as $(\xi(\tau)\Delta x,\xi(\tau)\Delta y,\xi(\tau)\Delta z)$
where $\xi(\tau)=\tau^{0.35}$ such that $\xi(0) = 0$ and $\xi(1) = 1$.
Given the deformation offsets $(\Delta x,\Delta y,\Delta z)$, we can visualize the dynamic 4D scene defined by Gaussians at a given time $\tau$ by running $\boldsymbol{\Phi}$ for all Gaussians and rendering them using the renderer from ~\citet{kerbl20233Dgaussians}.

\subsection{Frames-Per-Second (fps) Sampling}
At each 4D optimization step, we sample a fps $\in \{4,8,12\}$ from the distribution $p(\text{fps}=4) = 0.81, p(\text{fps}=8) = 0.14, p(\text{fps}=12) = 0.05$. 
This scheme samples lower fps more (faster video speed) to encourage our optimization process to converge to a 4D scene with larger motion.
Our time interval ranges from 0 to 1, and we set 0.75 as the length of the time interval covering a 16 frame 4 fps video (\ie $\tau \in [0,1]$ and 0.75 corresponds to a 4 seconds video). 
After an fps value is sampled, we sample the starting time $\tau_s \sim U(0, 1-3.0/\text{fps})$, and calculate the frame times for the 16 frames from $\tau_s$ and fps. This strategy is inspired by \citet{singer2023mav3d}.

\subsection{Rigidity Regularization} \label{supp:rigidity_reg}
Following \citet{luiten2023dynamic}, we regularize the deformation field such that nearby Gaussians deform similarly.
From the 3D optimized scene after the first stage, we pre-calculate 40 nearest neighbor Gaussians for each Gaussian.  
Then, at each 4D optimization step, we reduce the following loss:
\begin{equation}
\mathcal{L}_\textrm{Rigidity-Reg.}=\frac{1}{40} \sum_{i=1}^{40}||\boldsymbol{\Delta}_{\boldsymbol{\Phi}} - \boldsymbol{\Delta}_{\boldsymbol{\Phi}_{\text{NN}_i}} ||^2_2 
\end{equation}
where the term inside the summation denotes the L2 distance between the deformation values of Gaussians and their $i$-th nearest neighbor.

\subsection{JSD-based Regularization of the Evolving Distribution of the Dynamic 3D Gaussians} \label{supp:dyn_3dg_reg}
Here, we explicitly write out the regularization term introduced in \Cref{sec:scaling_ayg}, which is based on the Jensen-Shannon divergence and used to regularize the evolving distribution of the dynamic 3D Gaussians. It is used during the optimization of the deformation field when generating the temporal dynamics of the 4D sequences. The idea is to regularize the mean and the diagonal covariance of the distribution of the 3D Gaussians to stay approximately constant during the optimization of the temporal dynamics. The mean and the diagonal covariance capture an object's position (mean of the distribution of the Gaussians) and its size (variances of the distribution of the Gaussians in each $xyz$-direction, \ie, diagonal entries of the covariance matrix). Hence, keeping the mean and the diagonal covariance approximately constant ensures that the temporal dynamics of the 4D assets induced by the video model do not simply move the objects around or change them in size, but instead produce more complex and meaningful deformations and motions, while maintaining object size and position. The technique stabilizes the optimization of the deformation field and helps ensuring that the learnt motion is realistic.
Note that we also do not necessarily want the mean and covariances to be exactly the same across time, because some motion may naturally require them to vary (\eg a running horse would potentially be more realistic with an oscillating center of mass), hence we use a ``soft'' regularization strategy instead of, for instance, a hard re-centering and re-scaling during the optimization.\looseness=-1

Since we only want to regularize the mean and the variances in each $xyz$-direction, we can model the distribution of the 3D Gaussians at time $\tau$ by a Gaussian distribution $\mathcal{N}(\boldsymbol{\nu}_\tau,\boldsymbol{\Gamma}_\tau)$. We calculate the means $\boldsymbol{\nu}_\tau$ and diagonal covariances $\boldsymbol{\Gamma}_\tau$ of the entire set of 3D Gaussians (using their means $\boldsymbol{\mu}_i$) at times $\tau$ along the 4D sequence (see \Cref{fig:jsd_reg_figure}). As explained above, we would like to ensure that this distribution stays approximately the same during the optimization of the deformation field. 
To this end, we choose the Jensen--Shannon Divergence (JSD) as similarity metric measuring the distance between the distributions at different times $\tau$.
The JSD between $\mathcal{N}(\boldsymbol{\nu}_0,\boldsymbol{\Gamma}_0)$ at time $0$ and $\mathcal{N}(\boldsymbol{\nu}_\tau,\boldsymbol{\Gamma}_\tau)$ at time $\tau$ is
\begin{equation}
\begin{split}
\textrm{JSD}\left(\mathcal{N}(\boldsymbol{\nu}_0,\boldsymbol{\Gamma}_0)||\mathcal{N}(\boldsymbol{\nu}_\tau,\boldsymbol{\Gamma}_\tau)\right)&= \frac{1}{2} \textrm{KL}\left(\mathcal{N}(\boldsymbol{\nu}_0,\boldsymbol{\Gamma}_0) \bigg|\bigg|\frac{1}{2}\left(\mathcal{N}(\boldsymbol{\nu}_0,\boldsymbol{\Gamma}_0)+\mathcal{N}(\boldsymbol{\nu}_\tau,\boldsymbol{\Gamma}_\tau)\right)\right) \\ &\quad+ \frac{1}{2}\textrm{KL}\left(\mathcal{N}(\boldsymbol{\nu}_\tau,\boldsymbol{\Gamma}_\tau)\bigg|\bigg|\frac{1}{2}\left(\mathcal{N}(\boldsymbol{\nu}_0,\boldsymbol{\Gamma}_0)+\mathcal{N}(\boldsymbol{\nu}_\tau,\boldsymbol{\Gamma}_\tau)\right)\right).
\end{split}
\end{equation}
Unfortunately, the mixture distribution on the right hand side of the KL terms is generally not Gaussian. Therefore, it is generally not possible to derive a closed-form expression for the JSD between two Gaussians. However, we can make a simplifying modification here. Instead of calculating the mixture distribution we average the means and covariances from the two distributions $\mathcal{N}(\boldsymbol{\nu}_0,\boldsymbol{\Gamma}_0)$ and $\mathcal{N}(\boldsymbol{\nu}_\tau,\boldsymbol{\Gamma}_\tau)$ and construct a corresponding Gaussian distribution $\mathcal{N}\left(\frac{1}{2}\left(\boldsymbol{\nu}_0+\boldsymbol{\nu}_\tau\right),\frac{1}{2}\left(\boldsymbol{\Gamma}_0+\boldsymbol{\Gamma}_\tau\right)\right)$, which we use instead of the mixture distribution.
Hence, we have
\begin{equation}
\begin{split}
\textrm{JSD}\left(\mathcal{N}(\boldsymbol{\nu}_0,\boldsymbol{\Gamma}_0)||\mathcal{N}(\boldsymbol{\nu}_\tau,\boldsymbol{\Gamma}_\tau)\right)\rightarrow \mathcal{L}_\textrm{JSD-Reg.} &:= \frac{1}{2} \textrm{KL}\left(\mathcal{N}(\boldsymbol{\nu}_0,\boldsymbol{\Gamma}_0) \bigg|\bigg|\mathcal{N}\left(\frac{1}{2}\left(\boldsymbol{\nu}_0+\boldsymbol{\nu}_\tau\right),\frac{1}{2}\left(\boldsymbol{\Gamma}_0+\boldsymbol{\Gamma}_\tau\right)\right)\right) \\ &\quad+ \frac{1}{2}\textrm{KL}\left(\mathcal{N}(\boldsymbol{\nu}_\tau,\boldsymbol{\Gamma}_\tau)\bigg|\bigg|\mathcal{N}\left(\frac{1}{2}\left(\boldsymbol{\nu}_0+\boldsymbol{\nu}_\tau\right),\frac{1}{2}\left(\boldsymbol{\Gamma}_0+\boldsymbol{\Gamma}_\tau\right)\right)\right),
\end{split}
\end{equation}
which serves as our novel JSD-based regularization term $\mathcal{L}_\textrm{JSD-Reg.}$ to regularize the evolving distribution of the dynamic 3D Gaussians. Specifically, the deformation field is regularized with $\mathcal{L}_\textrm{JSD-Reg.}$ such that the mean and the variances in $xyz$-direction of the distribution of the 3D Gaussians at time $\tau$ do not deviate significantly from the corresponding mean and variances at time $0$ (recall that the initial Gaussians at $\tau=0$ are not subject to any deformation and hence fixed).

We can write out the above as
\begin{equation}
\begin{split}
\mathcal{L}_\textrm{JSD-Reg.} &:= \frac{1}{2} \textrm{KL}\left(\mathcal{N}(\boldsymbol{\nu}_0,\boldsymbol{\Gamma}_0) \bigg|\bigg|\mathcal{N}\left(\frac{1}{2}\left(\boldsymbol{\nu}_0+\boldsymbol{\nu}_\tau\right),\frac{1}{2}\left(\boldsymbol{\Gamma}_0+\boldsymbol{\Gamma}_\tau\right)\right)\right) \\ &\quad+ \frac{1}{2}\textrm{KL}\left(\mathcal{N}(\boldsymbol{\nu}_\tau,\boldsymbol{\Gamma}_\tau)\bigg|\bigg|\mathcal{N}\left(\frac{1}{2}\left(\boldsymbol{\nu}_0+\boldsymbol{\nu}_\tau\right),\frac{1}{2}\left(\boldsymbol{\Gamma}_0+\boldsymbol{\Gamma}_\tau\right)\right)\right) \\
&=\sum_{i\in\{x,y,z\}}\left[-\frac{1}{2}\log\left[2\right] + \frac{1}{2}\log\left[\boldsymbol{\Gamma}^i_0+\boldsymbol{\Gamma}^i_\tau\right] - \frac{1}{4}\log\left[\boldsymbol{\Gamma}^i_0\right] - \frac{1}{4}\log\left[\boldsymbol{\Gamma}^i_\tau\right] + \frac{1}{4}\frac{(\boldsymbol{\nu}^i_\tau-\boldsymbol{\nu}^i_0)^2}{\boldsymbol{\Gamma}^i_0+\boldsymbol{\Gamma}^i_\tau}
\right],
\end{split}
\end{equation}
for the three dimensions $i\in\{x,y,z\}$ and where $\boldsymbol{\Gamma}^i$ denotes the $i$-th diagonal entry in the diagonal covariance matrix $\boldsymbol{\Gamma}$ (since we use diagonal covariance matrices, we obtain this simple factorized form).
The gradients with respect to $\boldsymbol{\nu}_\tau$ and $\boldsymbol{\Gamma}_\tau$, which are calculated from the learnt distribution of the evolving 3D Gaussians, are
\begin{align}
\boldsymbol{\nabla}_{\boldsymbol{\nu}^i_\tau}\mathcal{L}_\textrm{JSD-Reg.}&=\frac{1}{2}\frac{\boldsymbol{\nu}^i_\tau-\boldsymbol{\nu}^i_0}{\boldsymbol{\Gamma}^i_0+\boldsymbol{\Gamma}^i_\tau}, \\
\boldsymbol{\nabla}_{\boldsymbol{\Gamma}^i_\tau}\mathcal{L}_\textrm{JSD-Reg.}&=\frac{1}{2}\frac{1}{\boldsymbol{\Gamma}^i_0+\boldsymbol{\Gamma}^i_\tau}-\frac{1}{4}\frac{1}{\boldsymbol{\Gamma}^i_\tau}-\frac{1}{4}\frac{(\boldsymbol{\nu}^i_\tau-\boldsymbol{\nu}^i_0)^2}{(\boldsymbol{\Gamma}^i_0+\boldsymbol{\Gamma}^i_\tau)^2}.
\end{align}
By analyzing the first and second derivatives it is easy to see that $\mathcal{L}_\textrm{JSD-Reg.}$ has a unique minimum when
\begin{equation}
\boldsymbol{\nu}^i_\tau = \boldsymbol{\nu}^i_0
\end{equation}
and
\begin{equation}
\boldsymbol{\Gamma}^i_\tau = \frac{1}{2}\left[(\boldsymbol{\nu}^i_\tau-\boldsymbol{\nu}^i_0)^2 + \sqrt{(\boldsymbol{\nu}^i_\tau-\boldsymbol{\nu}^i_0)^4+4{\boldsymbol{\Gamma}^i_0}^2} \right] \;\stackrel{\boldsymbol{\nu}^i_\tau=\boldsymbol{\nu}^i_0}{=}\;\boldsymbol{\Gamma}^i_0
\end{equation}
for all $i\in\{x,y,z\}$. This implies that $\mathcal{L}_\textrm{JSD-Reg.}$ is a meaningful regularization term for the means and variances in $xyz$-direction of the distribution of the evolving dynamic 3D Gaussians. It ensures that the distribution's mean $\boldsymbol{\nu}_\tau$ and diagonal covariance $\boldsymbol{\Gamma}_\tau$ remain close to the initial $\boldsymbol{\nu}_0$ and $\boldsymbol{\Gamma}_0$ during the optimization of the deformation field. That said, note that $\mathcal{L}_\textrm{JSD-Reg.}$ is not necessarily an accurate approximation of the true JSD. However, we found that this approach worked very well in practice during score distillation.

We opted for the JSD-based approach, because we wanted to use a symmetric distribution similarity metric. Potentially, we could have formulated our regularization also based on a Wasserstein distance, symmetrized KL divergence or regular non-symmetric KL divergence. However, we implemented the JSD-based approach above first and it worked right away and significantly improved AYG's optimization behavior and the 4D results. Therefore, we decided to leave the exploration of differently defined regularizations for the evolving distribution of the dynamic 3D Gaussians to future research.

\subsection{Camera Distribution} In the initial 3D stage, we follow MVDream~\cite{shi2023mvdream} to sample random cameras. 
At each optimization step, we sample a field-of-view $fov \sim U(15, 60)$, an elevation angle $elv \sim U(10, 45)$ and an azimuth angle $azm \sim U(0, 360)$. 
The camera distance is calculated as $cam\_d = s  / \tan(\dfrac{fov}{2}*\dfrac{\pi}{180})$ where $s \sim U(0.8, 1.0)$ is a random scaling factor.
The sampled camera looks towards the origin (0,0,0) and the images are rendered at $256\times256$ resolution.

At each 4D optimization step, we sample $fov \sim U(40, 70)$, $elv \sim U(-10, 45)$, $azm \sim U(0, 360)$, and $cam\_d \sim U(1.5, 3.0)$. 
We use \textit{dynamic} cameras where the camera location changes across the temporal time domain. 
For this purpose, we additionally sample offset values $elv\_offset \sim U(-13.5, 30)$ and $azm\_offset \sim U(-45, 45)$ and construct 16 cameras (to be used for rendering, and consequently, to be used as input to the video diffusion model that expects a 16-frame input) by setting $elv_i = elv + elv\_offset * i/15$ and $azm_i = azm + azm\_offset * i/15$ for $i \in \{0, ..., 15\}$, where $elv_i$ and $azm_i$ corresponds to the elevation and azimuth angles of the $i$-th frame's camera. We use the same field-of-view and camera distance across the 16 frames. 
We render frames at $160\times256$ resolution, which conforms to the aspect ratio that the video diffusion model is trained on.

\subsection{Diffusion Models} \label{sec:supp_diff_models}
As discussed, AYG leverages a latent text-to-image diffusion model~\cite{rombach2021highresolution}, a latent text-to-video diffusion model~\cite{blattmann2023videoldm}, and a latent text-to-multiview-image diffusion model~\cite{shi2023mvdream} in its compositional 4D score distillation framework.

For the latent text-to-multiview-image diffusion model we use MVDream~\cite{shi2023mvdream}.\footnote{\url{https://huggingface.co/MVDream/MVDream}} For the latent text-to-video diffusion model, we train our own model, largely following VideoLDM~\cite{blattmann2023videoldm} but with more training data and additional fps conditioning, see \Cref{sec:supp_video_model_training}. For the latent text-to-image diffusion model we use the spatial backbone of our newly trained video diffusion model, removing its temporal layers. This spatial backbone corresponds to a version of Stable Diffusion 2.1~\cite{rombach2021highresolution}\footnote{\url{https://huggingface.co/stabilityai/stable-diffusion-2-1}} that was fine-tuned for generation at resolution $320\times512$ (see details in \citet{blattmann2023videoldm}).

\subsection{Rendering Resolution}
In the initial 3D synthesis stage, we render the scenes at resolution $256\times256$. This is the resolution at which MVDream operates, \ie, its encoder takes images of this resolution. We additionally perform bilinear rescaling to $320\times512$ to calculate our text-to-image model's score distillation gradient. The text-to-image model's encoder accepts images of this resolution after fine-tuning (see previous subsection).
In the main 4D synthesis stage, we render at resolution $256\times160$ and perform bilinear rescaling to $320\times512$ to calculate both the text-to-image model's and the text-to-video model's score distillation gradients. They share the same encoder, which was fine-tuned to operate at this resolution.
For evaluation and visualization, we render at $512\times320$ resolution and pad to $512\times512$ if necessary.

\subsection{Additional Fine-tuning and Optimization} \label{supp:extra_finetuning}
We explored adding more Gaussians to further improve the quality of the generated 4D scenes.
As mentioned in \Cref{sec:supp_3d_representation}, we only perform densification steps if the total number of Gaussians is less than 50,000 for the initial 3D synthesis stage, and then optimize the deformation field in our 4D synthesis stage, as we found that this strategy gave us the best motion. 
Once optimized, we can further fine-tune the dynamic 4D scenes by doing a similar two-stage optimization. 
In the first additional stage, we continue optimizing the 3D Gaussians (not the deformation field) and do further densification steps even when the number of Gaussians is more than 50,000. On average, we end up with 150,000 Gaussians in this stage.
After that, in the second fine-tuning stage, we continue optimizing the deformation field, which has already been optimized well previously, with all the Gaussians including the newly added ones.
We use the same hyperparameters as in the initial 3D/4D stages except that we reduce the learning rates by 1/5 and increase the rendering image resolution to 512$\times$512 for the additional 3D optimization stage. For the two additional stages, we optimize for 7,000 steps and 3,000 steps on average, respectively. Also see the hyperparameters in \Cref{supp:hyperparams_text_to_4d}.

All results shown on the project page, \url{https://research.nvidia.com/labs/toronto-ai/AlignYourGaussians/}, correspond to 4D assets subject to this additional fine-tuning. All results shown in the paper itself as well as in the supplementary video \texttt{ayg$\_$text$\_$to$\_$4d.mp4} only use a single 3D plus 4D optimization run without this additional fine-tuning. Also the quantitative and qualitative comparisons to MAV3D~\cite{singer2023mav3d} in \Cref{sec:experiments,sec:supp_mav3d_comp} only use a single 3D and 4D optimization. This implies that with the additional fine-tuning described here and shown on the project page, we would possibly outperform MAV3D by an even larger margin in the user study. Also the ablation studies were carried out with only a single 3D plus 4D optimization without fine-tuning, and in that case we additionally only optimized for a total of 4,000 steps in the second 4D optimization stage in the interest of efficiency (see \Cref{sec:experiments,supp:extend_discussion_ablations}).

%% file: appendices/extended_composit_gen_details.tex
\section{Align Your Gaussian's Synthesis Framework}
Here, we discuss AYG's compositional generation framework in more detail and derive AYG's score distillation scheme.

\subsection{AYG's Compositional Generation Framework} 
Our goal during AYG's main 4D distillation is to simultaneously synthesize smooth and realistic temporal dynamics while also maintaining a high individual frame quality. Formally, this means that we would like to fulfill two objectives at once:

\begin{enumerate}
\setlength\itemsep{0.4em}
    \item A set of 2D frames $\{\rvx_{\tau_i}^{c_i}\}_F$ rendered from the 4D sequence at consecutive time steps $\tau_i$ and smoothly changing camera positions $c_i$ (dynamic cameras) should form a realistic dynamic video and should be high probability under the text-to-video model $p_\textrm{vid}(\{\rvz_{\tau_i}^{c_i}\}_F)$.
    \item Individual frames $\tilde{\rvx}_{\tilde{\tau}_j}^{\tilde{c}_j}$ at possibly different time steps $\tilde{\tau}_j$ and camera positions $\tilde{c}_j$ should, each individually, be high probability under the text-to-image model $p_\textrm{im}(\tilde{\rvx}_{\tilde{\tau}_j}^{\tilde{c}_j})$.
\end{enumerate}

\noindent If we want to ensure that both criteria are fulfilled simultaneously, it means that we are interested in generating 4D sequences whose renderings obey \textit{the product distribution} of $p_\textrm{vid}$ and $p_\textrm{im}$, \ie, $p_\textrm{vid}\,p_\textrm{im}$. This exactly corresponds to a form of compositional generation~\cite{liu2022compositional,du2023reduce}. In compositional generation, one often combines different diffusion models, or the same diffusion model with different text-conditionings, such that multiple conditions are fulfilled during generation, usually with the goal of controllable image generation or image editing. In that case, one leverages the product distribution of the corresponding diffusion model distributions with the different conditionings. Conceptually, this is analogous to what we are doing in AYG, just that we are doing a version of compositional generation by composing an image and a video diffusion model to generate dynamic 4D assets whose renderings obey both the text-to-image and text-to-video model distributions simultaneously. Instead of aiming for controllability our goal is to simultaneously enforce smooth temporal dynamics (video model) and high individual frame quality (image model).

One may ask, why do we even need the additional text-to-image model? Should the video model itself not also enforce high individual frame quality? The number of text-video pairs used to train text-to-video models is typically significantly smaller than the huge text-image datasets utilized when training text-to-image models. This makes text-to-video models, including the one we trained, often slightly less robust and more unstable than text-to-image models and they yield noisier gradients in the score distillation setting. Generally, video diffusion models are not as mature as image diffusion models yet. This motivates us to compose both, the image and the video model, where the video model can synthesize temporal dynamics and the image model ensures that high individual frame quality is maintained despite potentially somewhat noisy gradients of the video model. Looking at it from another perspective, the image model essentially stabilizes the optimization process.

We have explained this for the 4D generation stage here, but something very similar occurs also in AYG's initial 3D generation stage, where a text-conditioned multiview-image diffusion model~\cite{shi2023mvdream} is composed with a regular large-scale text-to-image model~\cite{rombach2021highresolution} in the form of their product distribution for score distillation. The motivation is almost the same. The multiview image model is able to generate a set of multiview-consistent images, similar to the set of consecutive frames of the video model, but it was trained only on a limited 3D dataset and therefore does encode the biases of this small 3D dataset. Including an additional general text-to-image prior can stabilize optimization and boost performance. Moreover, by augmenting the image model's text conditioning with view prompts, we can easily break the symmetry inherent in the multiview diffusion model, which does not generate objects in a canonical orientation. Note, however, that our main contribution is the compositional generation approach for AYG's 4D stage and this has been our main focus in this paper. AYG's 4D stage could also be combined with other methods to generate the initial static 3D assets.

\subsection{AYG's Score Distillation Scheme}
Let us now formalize this, while focusing on the 4D stage (derivations for the 3D stage proceed analogously). Following the above motivation, when optmizing the deformation field $\boldsymbol{\Phi}$ through score distillation we seek to minimize a reverse Kullback-Leibler divergence (KLD) from $q_{\boldsymbol{\Phi}}$, the distribution over the 4D scene's renderings defined through the deformation field $\boldsymbol{\Phi}$, to the product distribution $p_\textrm{vid}\,p_\textrm{im}$ of the diffusion models used as teachers, \ie, $\textrm{KL}\bigl(q_{\boldsymbol{\Phi}}\big|\big| p_\textrm{vid}p_\textrm{im}\bigr)$. Since $p_\textrm{vid}$ and $p_\textrm{im}$ are defined through diffusion models, in practice we minimize the KL not just for the clean renderings, but for diffused renderings at all diffusion time steps $t$, \ie,
\begin{equation}
    \textrm{KL}\biggl(q_{\boldsymbol{\Phi}}\left(\{\rvz_{\tau_i}^{c_i}\}_F,\{\tilde{\rvz}_{\tilde{\tau}_j}^{\tilde{c}_j}\}_M\right)\bigg|\bigg| p^\gamma_\textrm{vid}\left(\{\rvz_{\tau_i}^{c_i}\}_F\right)
    \prod_{j=1}^M p^\kappa_\textrm{im}\left(\tilde{\rvz}_{\tilde{\tau}_j}^{\tilde{c}_j}\right)\biggr),
\end{equation}
which is exactly the equation from the main paper in \Cref{sec:compo_gen}. Here, we now added some details: In general we can always consider a set of $F+M$ rendered frames, where $\{\rvz_{\tau_i}^{c_i}\}_F$ are the $F$ consecutive frames given to the video model and $\{\tilde{\rvz}_{\tilde{\tau}_j}^{\tilde{c}_j}\}_M$ are another $M$ independent frames given to the image model (since these are independent, the image model operates on them independently; hence the product in the equation). Following the notation in the main paper, $\rvz$ denotes diffused renderings and we omit diffusion time subscripts $t$. Note that, as discussed in the main paper, in practice the $M$ frames given to the image model are just a subset of the $F$ frames given to the video model. However, the framework is general. Moreover, $\gamma$ and $\kappa$ are temperatures which scale the width of the distributions $p_\textrm{vid}$ and $p_\textrm{im}$. They can act as hyperparameters controlling the relative strength of the gradients coming from the video and image model, respectively. Note that for now we omit explicitly indicating the text conditioning in the equations in the interest of brevity.

The derivation of our score distillation framework through minimizing the reverse KLD between the rendering distribution and the teacher distribution follows \citet{poole2023dreamfusion} (Appendix A.4). In practice, one typically gives different weights to the KL minimization for different diffusion times $t$, and the full SDS gradient can be written as
\begin{equation}\label{eq:supp:kl2}
\nabla_{\boldsymbol{\Phi}}\mathcal{L}^{\textrm{AYG}}_{\textrm{SDS}}(\{\mathbf{x}\}=g(\boldsymbol{\Phi}))=\mathbb{E}_{t,\boldsymbol{\epsilon}}\biggl[w(t)\frac{\sigma_t}{\alpha_t}\nabla_{\boldsymbol{\Phi}}\textrm{KL}\biggl(q_{\boldsymbol{\Phi}}\left(\{\rvz_{\tau_i}^{c_i}\}_F,\{\tilde{\rvz}_{\tilde{\tau}_j}^{\tilde{c}_j}\}_M\right)\bigg|\bigg| p^\gamma_\textrm{vid}\left(\{\rvz_{\tau_i}^{c_i}\}_F\right)
    \prod_{j=1}^M p^\kappa_\textrm{im}\left(\tilde{\rvz}_{\tilde{\tau}_j}^{\tilde{c}_j}\right)\biggr)\biggr],
\end{equation}
where $w(t)$ is a weighting function and $\alpha_t$ and $\sigma_t$ are the diffusion models' noise schedule (note that all our diffusion models use the same noise schedules). Specifically, the diffused renderings $\rvz$ are given as $\rvz=\alpha_t\rvx + \sigma_t\boldsymbol{\epsilon}$, where $\rvx$ is a clean rendering, $\boldsymbol{\epsilon}\sim \mathcal{N}(\mathbf{0},\mI)$ is noise from a standard normal distribution, and the parameters $\alpha_t$ and $\sigma_t$ are defined such the signal-to-noise ratio $\alpha_t/\sigma_t$ is strictly decreasing as a function of the diffusion time $t$. Moreover, usually $\alpha_t\in[0,1]$ and $\sigma_t\in[0,1]$ and in our case we use a ``variance-preserving'' noise schdule~\cite{song2020score,ho2020ddpm} such that $\alpha_t^2+\sigma_t^2=1$. Furthermore,
in \Cref{eq:supp:kl2} above, $g$ denotes the differentiable rendering function of the 4D scene, which produces all clean renderings $\{\rvx\}_{F+M}$ (omitting camera and 4D sequence time sub- and superscripts to keep notation concise), which are then diffused into $\{\rvz_{\tau_i}^{c_i}\}_F$ and $\{\tilde{\rvz}_{\tilde{\tau}_j}^{\tilde{c}_j}\}_M$.

We can now expand the KL divergence, similar to \citet{poole2023dreamfusion}:
\begin{equation}\label{eq:supp:kl3}
\nabla_{\boldsymbol{\Phi}}\mathcal{L}^{\textrm{AYG}}_{\textrm{SDS}}(\{\mathbf{x}\}_{F+M}=g(\boldsymbol{\Phi}))=\mathbb{E}_{t,\boldsymbol{\epsilon}}\biggl[w(t)\frac{\sigma_t}{\alpha_t}\biggl(\nabla_{\boldsymbol{\Phi}}\log q_{\boldsymbol{\Phi}}\left(\{\rvz_{\tau_i}^{c_i}\}_F,\{\tilde{\rvz}_{\tilde{\tau}_j}^{\tilde{c}_j}\}_M\right)-\nabla_{\boldsymbol{\Phi}}\log \biggl[p^\gamma_\textrm{vid}\left(\{\rvz_{\tau_i}^{c_i}\}_F\right)
    \prod_{j=1}^M p^\kappa_\textrm{im}\left(\tilde{\rvz}_{\tilde{\tau}_j}^{\tilde{c}_j}\right)\biggr]\biggr)\biggr].
\end{equation}
Let us now introduce some abbreviations for brevity, as in the main paper, and define $\rmZ:=\{\rvz_{\tau_i}^{c_i}\}_F$, $\tilde{\rmZ}:=\{\tilde{\rvz}_{\tilde{\tau}_j}^{\tilde{c}_j}\}_M$ and $\rmX:=\{\rvx\}_{F+M}$. Moreover, for a concise notation we write the product over the image distributions as one distribution over all the different diffused renderings with a slight abuse of notation, \ie,
\begin{equation}
    \prod_{j=1}^M p^\kappa_\textrm{im}\left(\tilde{\rvz}_{\tilde{\tau}_j}^{\tilde{c}_j}\right) = p^\kappa_\textrm{im}\left(\{\tilde{\rvz}_{\tilde{\tau}_j}^{\tilde{c}_j}\}_M\right)=p^\kappa_\textrm{im}\left(\tilde{\rmZ}\right).
\end{equation}
We can keep in mind that $p^\kappa_\textrm{im}\left(\tilde{\rmZ}\right)$ decomposes into a product, which will lead to different independent gradients. 
Now we have
\begin{equation}\label{eq:supp:kl4}
\begin{split}
\nabla_{\boldsymbol{\Phi}}\mathcal{L}^{\textrm{AYG}}_{\textrm{SDS}}(\rmX=g(\boldsymbol{\Phi}))&=\mathbb{E}_{t,\boldsymbol{\epsilon}}\biggl[w(t)\frac{\sigma_t}{\alpha_t}\biggl(\nabla_{\boldsymbol{\Phi}}\log q_{\boldsymbol{\Phi}}\left(\rmZ,\tilde{\rmZ}\right)-\nabla_{\boldsymbol{\Phi}}\log \biggl[p^\gamma_\textrm{vid}\left(\rmZ\right)
    p^\kappa_\textrm{im}\left(\tilde{\rmZ}\right)\biggr]\biggr)\biggr] \\
    &=\mathbb{E}_{t,\boldsymbol{\epsilon}}\biggl[w(t)\frac{\sigma_t}{\alpha_t}\biggl(\nabla_{\boldsymbol{\Phi}}\log q_{\boldsymbol{\Phi}}\left(\rmZ\right)-\gamma\,\nabla_{\boldsymbol{\Phi}}\log p_\textrm{vid}\left(\rmZ\right) \\
    &\qquad\qquad\qquad+\nabla_{\boldsymbol{\Phi}}\log q_{\boldsymbol{\Phi}}\left(\tilde{\rmZ}\right)-\kappa\,\nabla_{\boldsymbol{\Phi}}\log 
    p_\textrm{im}\left(\tilde{\rmZ}\right)\biggr)\biggr],
\end{split}
\end{equation}
where we decomposed the $\log$-terms. We can now analyze the individual terms, analogous to \citet{poole2023dreamfusion}. We have
\begin{equation}
    \nabla_{\boldsymbol{\Phi}}\log p_\textrm{vid}\left(\rmZ\right) = \nabla_{\rmZ}\log p_\textrm{vid}\left(\rmZ\right) \frac{\partial \rmZ}{\partial \boldsymbol{\Phi}} = -\frac{1}{\sigma_t} \hat{\boldsymbol{\epsilon}}^{\textrm{vid}}(\rmZ,v,t) \frac{\partial \rmZ}{\partial \boldsymbol{\Phi}}=-\frac{\alpha_t}{\sigma_t} \hat{\boldsymbol{\epsilon}}^{\textrm{vid}}(\rmZ,v,t) \frac{\partial \rmX}{\partial \boldsymbol{\Phi}},
\end{equation}
where we inserted the text-to-video model's denoiser neural network $\hat{\boldsymbol{\epsilon}}^{\textrm{vid}}(\rmZ,v,t)$, which represents the diffusion model's score $\nabla_{\rmZ}\log p_\textrm{vid}\left(\rmZ\right)$ as~\cite{ho2020ddpm,song2020score,vahdat2021score}
\begin{equation}
    \nabla_{\rmZ}\log p_\textrm{vid}\left(\rmZ\right) \sim -\frac{1}{\sigma_t} \hat{\boldsymbol{\epsilon}}^{\textrm{vid}}(\rmZ,v,t),
\end{equation}
and we now also explicitly indicate the model's conditioning on the diffusion time $t$ and a text prompt $v$. The image model term $\nabla_{\boldsymbol{\Phi}}\log p_\textrm{im}\left(\tilde{\rmZ}\right)$ can be written similarly.
Moreover, we have 
\begin{equation} \label{eq:supp:kl6}
\begin{split}
    \nabla_{\boldsymbol{\Phi}}\log q_{\boldsymbol{\Phi}}\left(\rmZ\right) & = \left(\frac{\partial \log q_{\boldsymbol{\Phi}}\left(\rmZ\right)}{\partial \rmX} + \frac{\partial\log q_{\boldsymbol{\Phi}}\left(\rmZ\right)}{\partial \rmZ} \frac{\partial \rmZ}{\partial \rmX}\right)\frac{\partial \rmX}{\partial \boldsymbol{\Phi}} \\
    & = \left(-\frac{\partial}{\partial \rmX}(\alpha_t\rmX - \rmZ)^2 - \alpha_t \frac{\partial}{\partial \rmZ}(\alpha_t\rmX - \rmZ)^2 \right) \frac{1}{2\sigma_t^2}\frac{\partial \rmX}{\partial \boldsymbol{\Phi}} \\
    & = \left(-(\alpha_t\rmX - \rmZ) + (\alpha_t\rmX - \rmZ) \right) \frac{\alpha_t}{\sigma_t^2}\frac{\partial \rmX}{\partial \boldsymbol{\Phi}} \\
    & = \left(-(\alpha_t\rmX - \alpha_t\rmX - \sigma_t \boldsymbol{\epsilon}) + (\alpha_t\rmX - \alpha_t\rmX - \sigma_t \boldsymbol{\epsilon}) \right) \frac{\alpha_t}{\sigma_t^2}\frac{\partial \rmX}{\partial \boldsymbol{\Phi}} \\
    & = \left(\frac{\alpha_t}{\sigma_t}\boldsymbol{\epsilon}-\frac{\alpha_t}{\sigma_t}\boldsymbol{\epsilon}\right)\frac{\partial \rmX}{\partial \boldsymbol{\Phi}} \\
    & = 0
\end{split}
\end{equation}
which follows exactly \citet{poole2023dreamfusion}, Appendix A.4, and analogously for $\nabla_{\boldsymbol{\Phi}}\log q_{\boldsymbol{\Phi}}\left(\tilde{\rmZ}\right)$. The first term in \Cref{eq:supp:kl6} corresponds to the gradient directly with respect to the variational parameters $\boldsymbol{\Phi}$ through the renderings $\rmX$, while the second term is the path derivative through the diffused samples $\rmZ$~\cite{roeder2017sticking}. Note that in score distillation, we take an expectation over $\boldsymbol{\epsilon}$ and after such expectation both terms are individually zero as the noise $\boldsymbol{\epsilon}$ has $\mathbf{0}$ mean.

We can now write our SDS gradient as
\begin{equation}\label{eq:supp:kl7}
\nabla_{\boldsymbol{\Phi}}\mathcal{L}^{\textrm{AYG}}_{\textrm{SDS}}(\rmX=g(\boldsymbol{\Phi}))=\mathbb{E}_{t,\boldsymbol{\epsilon}}\biggl[w(t)\biggl(\gamma\,\hat{\boldsymbol{\epsilon}}^{\textrm{vid}}(\rmZ,v,t) + \kappa\,\hat{\boldsymbol{\epsilon}}^{\textrm{im}}(\tilde{\rmZ},v,t)\biggr)\biggr]\frac{\partial \rmX}{\partial \boldsymbol{\Phi}}.
\end{equation}
However, in SDS one typically includes the path derivative gradient (the second term in \Cref{eq:supp:kl6}) as a zero mean control variate to reduce the variance of the SDS gradient when using a sample-based approximation of the expectation, following \citet{roeder2017sticking}. In other words, since
\begin{equation}
    \mathbb{E}_{t,\boldsymbol{\epsilon}}\biggl[-w(t)\frac{\alpha_t}{\sigma_t}\boldsymbol{\epsilon}\biggr]\frac{\partial \rmX}{\partial \boldsymbol{\Phi}} = 0
\end{equation}
after the expectation of the diffusion noise $\boldsymbol{\epsilon}$, we can freely include such terms in the SDS gradient as control variates and scale with the temperatures $\gamma$ and $\kappa$ as needed. Now explicitly defining the different noise values $\boldsymbol{\epsilon}^{\textrm{vid}}$ and $\boldsymbol{\epsilon}^{\textrm{im}}$ as the diffusion noises for the perturbed renderings given to the video and the image model, respectively, we can write
\begin{equation}\label{eq:supp:kl8}
\begin{split}
\nabla_{\boldsymbol{\Phi}}\mathcal{L}^{\textrm{AYG}}_{\textrm{SDS}}(\rmX=g(\boldsymbol{\Phi}))&=\mathbb{E}_{t,\boldsymbol{\epsilon}}\biggl[w(t)\biggl(\gamma\,\hat{\boldsymbol{\epsilon}}^{\textrm{vid}}(\rmZ,v,t) + \kappa\,\hat{\boldsymbol{\epsilon}}^{\textrm{im}}(\tilde{\rmZ},v,t)\biggr)\biggr]\frac{\partial \rmX}{\partial \boldsymbol{\Phi}} \\
&=\mathbb{E}_{t,\boldsymbol{\epsilon}^{\textrm{vid}},\boldsymbol{\epsilon}^{\textrm{im}}}\biggl[w(t)\biggl(\gamma\,(\hat{\boldsymbol{\epsilon}}^{\textrm{vid}}(\rmZ,v,t)-\boldsymbol{\epsilon}^{\textrm{vid}}) + \kappa\,(\hat{\boldsymbol{\epsilon}}^{\textrm{im}}(\tilde{\rmZ},v,t)-\boldsymbol{\epsilon}^{\textrm{im}})\biggr)\biggr]\frac{\partial \rmX}{\partial \boldsymbol{\Phi}}.
\end{split}
\end{equation}
Note that we have been using different noise values to perturb the different renderings from the very beginning of the derivation, but we have not been explicit about it in the notation so far in the interest of conciseness.

In practice, one typically employs classifier-free guidance (CFG)~\cite{ho2021classifierfree} with guidance weights $\omega_\textrm{vid/im}$ for the video and image model, respectively. Hence, we have that
\begin{equation}
    \hat{\boldsymbol{\epsilon}}^{\textrm{vid}}(\rmZ,v,t) \rightarrow \hat{\boldsymbol{\epsilon}}^{\textrm{vid}}(\rmZ,v,t) + \omega_\textrm{vid}\left[\hat{\boldsymbol{\epsilon}}^{\textrm{vid}}(\rmZ,v,t)-\hat{\boldsymbol{\epsilon}}^{\textrm{vid}}(\rmZ,t)\right]
\end{equation}
for the video model, and for the image model analogously ($\hat{\boldsymbol{\epsilon}}^{\textrm{vid}}(\rmZ,t)$ indicates the denoiser prediction without text conditioning). The no-CFG setting is recovered for $\omega_\textrm{vid/im}=0$. Inserting above, we obtain
\begin{equation}\label{eq:supp:kl9}
\begin{split}
\nabla_{\boldsymbol{\Phi}}\mathcal{L}^{\textrm{AYG}}_{\textrm{SDS}}(\rmX=g(\boldsymbol{\Phi}))&=\mathbb{E}_{t,\boldsymbol{\epsilon}^{\textrm{vid}},\boldsymbol{\epsilon}^{\textrm{im}}}\biggl[w(t)\biggl\{\gamma\biggl(\omega_{\textrm{vid}}\bigl[\hat{\boldsymbol{\epsilon}}^{\textrm{vid}}(\rmZ,v,t)- \hat{\boldsymbol{\epsilon}}^{\textrm{vid}}(\rmZ,t)\bigr] + \underbrace{\hat{\boldsymbol{\epsilon}}^{\textrm{vid}}(\rmZ,v,t)-\boldsymbol{\epsilon}^{\textrm{vid}}}_{\boldsymbol{\delta}_{\textrm{gen}}^{\textrm{vid}}}\biggr) \\
& \qquad\qquad\quad + \kappa\biggl(\omega_{\textrm{im}}\bigl[\hat{\boldsymbol{\epsilon}}^{\textrm{im}}(\tilde{\rmZ},v,t)-\hat{\boldsymbol{\epsilon}}^{\textrm{im}}(\tilde{\rmZ},t)\bigr] + \underbrace{\hat{\boldsymbol{\epsilon}}^{\textrm{im}}(\tilde{\rmZ},v,t)-\boldsymbol{\epsilon}^{\textrm{im}}}_{\boldsymbol{\delta}_{\textrm{gen}}^{\textrm{im}}}\biggr)
\biggr\}\frac{\partial \rmX}{\partial\boldsymbol{\Phi}}\biggr],
\end{split}
\end{equation}
which is exactly \Cref{eq:ayg_score} from the main paper.

As discussed in the main paper, recently, ProlificDreamer~\cite{wang2023prolificdreamer} proposed a scheme where the noise-based control variate $\boldsymbol{\epsilon}$ ($\boldsymbol{\epsilon}^{\textrm{vid}}$ and $\boldsymbol{\epsilon}^{\textrm{im}}$ here) is replaced by a separate diffusion model that models the rendering distribution. More specifically, ProlificDreamer initializes this separate diffusion model from the diffusion model guiding the synthesis ($\hat{\boldsymbol{\epsilon}}^{\textrm{vid}}(\rmZ,v,t)$ and $\hat{\boldsymbol{\epsilon}}^{\textrm{im}}(\tilde{\rmZ},v,t)$ here), and then slowly fine-tunes on the diffused renderings ($\rmZ$ or $\tilde{\rmZ}$ here). In our setting, one would initialize from the video and the image model, respectively, in the two terms.
This means that, at the beginning of optimization, the terms $\boldsymbol{\delta}_{\textrm{gen}}^{\textrm{vid/im}}$ in \Cref{eq:supp:kl9} would be zero. Inspired by this observation and aiming to avoid ProlificDreamer's cumbersome fine-tuning, we instead propose to simply set $\boldsymbol{\delta}_{\textrm{gen}}^{\textrm{vid/im}}=0$ entirely, and simply optimize with
\begin{equation}\label{eq:supp:kl10}
\begin{split}
\nabla_{\boldsymbol{\Phi}}\mathcal{L}^{\textrm{AYG}}_{\textrm{CSD}}(\rmX=g(\boldsymbol{\Phi}))&=\mathbb{E}_{t,\boldsymbol{\epsilon}^{\textrm{vid}},\boldsymbol{\epsilon}^{\textrm{im}}}\biggl[w(t)\biggl\{
\omega_{\textrm{vid}}\bigl[\underbrace{\hat{\boldsymbol{\epsilon}}^{\textrm{vid}}(\rmZ,v,t)- \hat{\boldsymbol{\epsilon}}^{\textrm{vid}}(\rmZ,t)}_{\boldsymbol{\delta}_{\textrm{cls}}^{\textrm{vid}}}\bigr] + \omega_{\textrm{im}}\bigl[\underbrace{\hat{\boldsymbol{\epsilon}}^{\textrm{im}}(\tilde{\rmZ},v,t)-\hat{\boldsymbol{\epsilon}}^{\textrm{im}}(\tilde{\rmZ},t)}_{\boldsymbol{\delta}_{\textrm{cls}}^{\textrm{im}}}\bigr]
\biggr\}\frac{\partial\rmX}{\partial\boldsymbol{\Phi}}\biggr] \\
& = \mathbb{E}_{t,\boldsymbol{\epsilon}^{\textrm{vid}},\boldsymbol{\epsilon}^{\textrm{im}}}\biggl[w(t)\biggl\{
\omega_{\textrm{vid}}\boldsymbol{\delta}_{\textrm{cls}}^{\textrm{vid}} + \omega_{\textrm{im}}\boldsymbol{\delta}_{\textrm{cls}}^{\textrm{im}}
\biggr\}\frac{\partial\rmX}{\partial\boldsymbol{\Phi}}\biggr],
\end{split}
\end{equation}
which is exactly \Cref{eq:ayg_score2} from the main paper. Moreover, we have absorbed $\gamma$ and $\kappa$ into $\omega_{\textrm{vid/im}}$. Interestingly, this exactly corresponds to the concurrently proposed classifier score distillation (CSD)~\cite{yu2023csd}, which points out that the above two terms $\boldsymbol{\delta}_{\textrm{cls}}^{\textrm{vid/im}}$ correspond to implicit classifiers predicting the text $v$ from the video or images, respectively. CSD then proposes to use only $\boldsymbol{\delta}_{\textrm{cls}}^{\textrm{vid/im}}$ for score distillation, resulting in improved performance over SDS. We discovered that scheme independently, while aiming to inherit ProlificDreamer's strong performance.

The score for the first 3D stage looks exactly analogous, just that instead of the composition of the video and image diffusion model, we have the composition of the MVDream multiview image diffusion model and the regular text-to-image model.

\subsection{AYG's Parameter Gradients---Putting it All Together}
In practice, AYG additionally uses negative prompt guidance, motion amplification, JSD regularization, and rigidity regularization in the 4D stage, as well as view guidance in the initial 3D stage, see \Cref{sec:scaling_ayg}. Let us now put everything together. 

\textbf{Stage 1: 3D Synthesis.} In the 3D stage, the entire gradient backpropagated into the 3D Gaussian splatting representation $\boldsymbol{\theta}$ is
\begin{equation}
\begin{split}
\nabla_{\boldsymbol{\theta}}\mathcal{L}^{\textrm{AYG}}_{\textrm{3D-stage}} & = \mathbb{E}_{t,\boldsymbol{\epsilon}^{\textrm{3D}},\boldsymbol{\epsilon}^{\textrm{im}}}\biggl[w(t)\biggl\{
\omega_{\textrm{3D}}\boldsymbol{\delta}_{\textrm{cls}}^{\textrm{3D}} + \omega_{\textrm{im}}\boldsymbol{\delta}_{\textrm{cls}}^{\textrm{im}} \\ 
& \qquad\qquad\quad + \omega_{\textrm{vg}} \left(\hat{\boldsymbol{\epsilon}}^{\textrm{im}}(\tilde{\rmZ},v^\textrm{aug},t)-\hat{\boldsymbol{\epsilon}}^{\textrm{im}}(\tilde{\rmZ},v,t)\right) \\
& \qquad\qquad\quad + \omega_{\textrm{neg}} \left(\hat{\boldsymbol{\epsilon}}^{\textrm{3D}}(\rmZ,t)-\hat{\boldsymbol{\epsilon}}^{\textrm{3D}}(\rmZ,v^\textrm{neg},t)\right)
\biggr\}\frac{\partial\rmX}{\partial\boldsymbol{\theta}}\biggr],
\end{split}
\end{equation}
where $\boldsymbol{\delta}_{\textrm{cls}}^{\textrm{3D}}$ corresponds to the implicit classifier score for the text-conditioned 3D-aware MVDream multiview diffusion model, analogous to the implicit classifier score for the video model above. Moreover, we added the view guidance and negative prompting terms~\cite{yu2023csd}. Our negative prompt $v^\textrm{neg}$ in the 3D stage is \textit{``ugly, bad anatomy, blurry, pixelated obscure, unnatural colors, poor lighting, dull, and unclear, cropped, lowres, low quality, artifacts, duplicate, morbid, mutilated, poorly drawn face, deformed, dehydrated, bad proportions''}, following \citet{shi2023mvdream}. We apply negative prompting only to the MVDream 3D multiview diffusion model (the negative prompt is taken from MVDream, after all). 
For view guidance, we attach ``, front view'' ($azm \le 30^{\circ}$ or $azm \ge 330^{\circ}$),  ``, back view'' ($150^{\circ} \le azm \le 210^{\circ}$),  ``, overhead view'' ($elv \ge 60^{\circ}$) and ``, side view'' (otherwise) at the end of a text prompt to construct the augmented prompt $v^\textrm{aug}$. Furthermore, $\boldsymbol{\epsilon}^{\textrm{3D}}$ and $\boldsymbol{\epsilon}^{\textrm{im}}$ denote the diffusion noise values used to perturb the renderings given to the 3D-aware diffusion model and the regular text-to-image model, respectively.

\textbf{Stage 2: Adding Dynamics for 4D Synthesis.} In the 4D stage, we apply our novel motion amplification mechanism (\Cref{sec:scaling_ayg}) after we combined the regular implicit classifier score of the video model ${\boldsymbol{\delta}_{\textrm{cls}}^{\textrm{vid}}}$ with the additional negative prompting term. Therefore, let us define
\begin{equation}
    \hat{\boldsymbol{\delta}}_{\textrm{cls+neg}}^{\textrm{vid}}:={\boldsymbol{\delta}_{\textrm{cls}}^{\textrm{vid}}}+ \omega_{\textrm{neg}} \left(\hat{\boldsymbol{\epsilon}}^{\textrm{vid}}(\rmZ,t)-\hat{\boldsymbol{\epsilon}}^{\textrm{vid}}(\rmZ,v^\textrm{neg},t)\right).
\end{equation}
We can now write out the entire gradient used to update the deformation field $\boldsymbol{\Phi}$ in AYG's main 4D stage in a concise manner as
\begin{equation}
\begin{split}
\nabla_{\boldsymbol{\Phi}}\mathcal{L}^{\textrm{AYG}}_{\textrm{4D-stage}} & = \mathbb{E}_{t,\boldsymbol{\epsilon}^{\textrm{vid}},\boldsymbol{\epsilon}^{\textrm{im}}}\biggl[w(t)\biggl\{
\omega_{\textrm{vid}}\biggl(\left<\hat{\boldsymbol{\delta}}_{\textrm{cls+neg}}^{\textrm{vid}}\right>_{\textrm{frame-avg.}} + \omega_\textrm{ma}\left[\hat{\boldsymbol{\delta}}_{\textrm{cls+neg}}^{\textrm{vid}} -\left<\hat{\boldsymbol{\delta}}_{\textrm{cls+neg}}^{\textrm{vid}}\right>_{\textrm{frame-avg.}}\right]\biggr) + \omega_{\textrm{im}}\boldsymbol{\delta}_{\textrm{cls}}^{\textrm{im}}
\biggr\}\frac{\partial\rmX}{\partial\boldsymbol{\Phi}}\biggr] \\
& \quad + \lambda_{\textrm{JSD}}\ \nabla_{\boldsymbol{\Phi}}\mathcal{L}_\textrm{JSD-Reg.} + \lambda_{\textrm{Rigidity}}\nabla_{\boldsymbol{\Phi}}\mathcal{L}_{\textrm{Rigidity-Reg.}},
\end{split}
\end{equation}
where we inserted the motion amplification mechanism with scale $\omega_\textrm{ma}$, negative prompting with weight $\omega_\textrm{neg}$, as well as the additional regularizers. We only do negative prompting for the video model, using \textit{``low motion, static statue, not moving, no motion''} as the negative prompt $v^\textrm{neg}$.

\textbf{Use of Latent Diffusion Models.} We would like to remind the reader that in the above derivations we have not explicitly included our diffusion models' encoders. All employed diffusion models are \textit{latent} diffusion models~\cite{rombach2021highresolution,vahdat2021score}, see \Cref{sec:supp_diff_models}. In practice, all score distillation gradients are calculated in latent space on encodings of the 4D scene renderings, and the gradients are backpropagated through the models' encoders before further backpropagated through the differentiable rendering process into the 4D scene representation. However, AYG's synthesis framework is general.

%% file: appendices/experiment_details.tex
\section{Experiment Details}

\subsection{Video Diffusion Model Training}\label{sec:supp_video_model_training}
We train a text-to-video diffusion model, following VideoLDM \cite{blattmann2023videoldm}. VideoLDM is a latent diffusion model that builds on Stable Diffusion~\cite{rombach2021highresolution} as text-to-image backbone model and fine-tunes it on text-video datasets by inserting additional temporal layers into the architecture. For details, see \citet{blattmann2023videoldm}. Here, we add conditioning on the frames-per-second (fps) frame rate, following~\cite{singer2023makeavideo,singer2023mav3d}, scale the number of frames from 8 to 16 frames, and train the model on a larger dataset. Note that our video model was trained only for the purpose of 4D score distillation and is not meant to be a standalone video generation system, which would require additional upsampling and interpolation modules.

\textbf{Datasets.} Our training initially utilizes the WebVid-10M dataset \cite{bain21frozen}, which comprises 10 million text-video pairs. In our analysis of the WebVid dataset, we noted the presence of many slow-motion videos. To address this, we implement a simple filtering algorithm to exclude such data. Initially, we calculate optical flow at 2 fps utilizing the iterative Lucas-Kanade method~\cite{lucas1981iterative}.
Following this, we filter out any videos where the average optical flow magnitude falls below a specified threshold. Specifically, we set this threshold to be 10. 
We further filter data with aesthetic score~\cite{schuhmann2022laion} lower than 4.0 and train our model for 100K iterations. 
To enhance the model's generalization capabilities, we incorporate the HDVG-130M dataset \cite{wang2023videofactory} in a subsequent fine-tuning phase for an additional 100K iterations. 
During fine-tuning, we specifically exclude data containing keywords such as `talk', `chat', `game', `screenshot', `newspaper', `screen', and `microphone' in text captions. 
These keywords often indicate video recordings of interviews showing little scene dynamics or computer games with very unusual, non-photorealistic visual appearance. 
Since there are many such videos in the dataset and we are not interested in generating such videos or 4D scenes in this paper, we remove these videos from the training data.
We also increase the optical flow slow motion filtering~\cite{lucas1981iterative} threshold to 20.
Moreover, since WebVid-10M is much smaller than HDVG-130M, but provides valuable training data with high-quality text annotations by human annotators, we slightly oversample the filtered WebVid-10M data. Specifically, we do not sample training data as if we just merged WebVid and HDVG and then randomly drew samples from the resulting dataset to form training batches. Instead, we sample in such a way as if we tripled the filtered WebVid data and only then merged with the filtered HDVG and sampled from the full training dataset randomly to construct training batches.
We train the model on 128 NVIDIA-80G-A100 GPUs, utilizing fp16 precision and a gradient accumulation strategy with a batch size of 4 per-GPU. The total batch size, distributed across all GPUs, is 2048, achieved after complete gradient accumulation. 

\textbf{Frames-per-second (fps) Conditioning.} We condition the model on the frames-per-second (fps) frame rate (by using cross-attention to the corresponding sinusoidal embedding). During training, the fps values are randomly sampled within a range of 2 to 16. Given that lower fps often indicates more motion and presents a greater learning challenge, we sample low fps more often.  Specifically, we sample the training video fps based on the probability distribution  $p(\textrm{fps}) \sim {1 / \textrm{fps}^c }$, where $c$ is a hyperparameter. We choose $c=1.6$ in our experiments. 

Otherwise, our training exactly follows VideoLDM~\cite{blattmann2023videoldm}. Samples from AYG's latent video diffusion model are shown in \Cref{fig:supp_video_samples1,fig:supp_video_samples2,fig:supp_video_samples3} in \Cref{sec:supp:video_model_samples} as well as in the supplementary video \texttt{ayg$\_$new$\_$video$\_$model.mp4}, demonstrating the effect of the fps conditioning.

It is worth noting that we also explored training a larger latent video diffusion model with more parameters, improved temporal attention layers and additional cross-frame attention~\cite{khachatryan2023text2videozero,wu2023tune} for higher performance. While this model indeed was able to generate noticeably higher quality videos, this did not translate into improved 4D distillation performance when used during score distillation. It would be valuable to study the effect of different video diffusion models during text-to-4D synthesis via score distillation in more detail. Based on these observations, we kept using the more memory efficient model described in this section above, which more directly follows the architecture from \citet{blattmann2023videoldm}, apart from the fps conditioning.

\subsection{Text-to-4D Hyperparameters}\label{supp:hyperparams_text_to_4d}
\textbf{Stage 1.} Table~\ref{tab:supp_hyperparam_3d} summarizes the hyperparameters for the 3D optimization stage. $bs$ denotes the batch size, and $lr_{\text{position}}$, $lr_{\text{rgb}}$, $lr_{\text{sh}}$, $lr_{\text{opacity}}$ and $lr_{\text{scaling}}$ denote the learning rates for the 3D Gaussians' position, color, spherical harmonics, opacity and scaling parameters, respectively. We use a single GPU for the optimization in the first 3D synthesis stage and a batch size of 4. This means that we use 4 independent sets of 4 images each, given to MVDream, which takes sets of 4 images as input. Each set of 4 images consists of renders with the correct relative camera angles for the multiview diffusion model MVDream. The 4 sets of images are then fed to MVDream, and all images are additionally fed to the regular text-to-image diffusion model (Stable Diffusion). Hence, in each optimization step, 16 different images are rendered in total from the 3D scene.
For $lr_{\text{position}}$, we start from 0.001 and decay to the specified value by the 500-th iteration. We note that we followed the learning rate schedules used by DreamGaussian~\cite{tang2023dreamgaussian}.
$\omega_{\textrm{vg}}$ and $\omega_{\textrm{neg}}$ denote the view guidance scale and negative prompt guidance scale, respectively. 
$\omega_{\textrm{3D}}$ is the weighting factor for the classifier score distillation term from MVDream~\cite{shi2023mvdream} ($p_{\text{3D}}$) and  $\omega_{\textrm{im}}$ is the weighting factor for the classifier score distillation term from the image diffusion model ($p_{\text{im}}$) (see Section~\ref{sec:compo_gen}).
We sample the diffusion time $t$ in the range $[0.02, 0.98]$ at the start of optimization and decay the range to $[0.02, 0.5]$ by the 6,000-th iteration for the image diffusion model ($p_{\text{im}}$).
For MVDream~\cite{shi2023mvdream} ($p_{\text{3D}}$), we directly follow their schedule which samples $t$ from $[0.98,0.98]$ at the start of optimization and decay the range to $[0.02, 0.5]$ by the 8,000-th iteration. We randomly choose black or white background during training. 
We run 10,000 optimization steps on average for this stage.

\begin{table}
\centering
\caption{Hyperparameters for the first stage (3D synthesis).}
\vspace{-0.2cm}
\label{tab:supp_hyperparam_3d}
\scalebox{0.95}{\rowcolors{1}{white}{gray!15}
\begin{tabular}{c c c c c c c c c c c c}
    \toprule
     $bs$ per GPU & \# GPUs & \# renders & $lr_{\text{position}}$ &  $lr_{\text{rgb}}$ &  $lr_{\text{sh}}$ & $lr_{\text{opacity}}$ & $lr_{\text{scaling}}$ & $\omega_{\textrm{vg}}$ & $\omega_{\textrm{neg}}$ & $\omega_{\textrm{3D}}$ & $\omega_{\textrm{im}}$ \\ 
    \midrule
     4 & 1 & 16 & 0.0002 & 0.01 & 0.0005 & 0.05 & 0.005 & 3.0 & 0.8 & 1.6 & 0.4 \\
    \bottomrule
\end{tabular}}
\end{table}
\begin{table}
\centering
\caption{Hyperparameters for the second stage (dynamic 4D synthesis).}
\vspace{-0.2cm}
\label{tab:supp_hyperparam_4d}
\scalebox{0.95}{\rowcolors{1}{white}{gray!15}
\begin{tabular}{c c c c c c c c c c c c}
    \toprule
     $bs$ per GPU & \# GPUs & \# renders & $lr_{\boldsymbol{\Phi}}$ & num. hidden $\boldsymbol{\Phi}$& num. layers $\boldsymbol{\Phi}$ & $\lambda_{\textrm{Rigidity}}$ &  $\lambda_{\textrm{JSD}}$  & $\omega_{\textrm{ma}}$ &  $\omega_{\textrm{neg}}$ &
     $\omega_{\textrm{vid}}$ & $\omega_{\textrm{im}}$ \\
    \midrule
     1 & 4 & 64 & 0.001 & 128 & 5 & 100.0 & 30.0 & 24.0 & 0.8 & 1.0 & 1.0 \\
    \bottomrule
\end{tabular}}
\end{table}

\textbf{Stage 2.} Table~\ref{tab:supp_hyperparam_4d} summarizes the hyperparameters for the dynamic 4D optimization stage. $bs$ again denotes the batch size, and $lr_{\boldsymbol{\Phi}}$, ``num. hidden $\boldsymbol{\Phi}$" and ``num. layers $\boldsymbol{\Phi}$"  denote the learning rate, number of hidden units and number of layers for the deformation MLP, respectively.
$\lambda_{\textrm{Rigidity}}$ and $\lambda_{\textrm{JSD}}$ denote the weighting terms for the rigidity regularization (\Cref{supp:rigidity_reg}) and the JSD-based dynamic 3D Gaussian distribution regularization (\Cref{supp:dyn_3dg_reg}).
$\omega_{\textrm{ma}}$ specifies the motion amplification scale, while $\omega_{\textrm{neg}}$ is the negative prompt guidance scale for the video diffusion model.
$\omega_{\textrm{vid}}$ denotes the weighting term for the video DM's ($p_{\text{vid}}$) classifier score distillation term and  $\omega_{\textrm{im}}$ denotes the weighting term for the image model's ($p_{\text{im}}$) classifier score distillation term (see Section~\ref{sec:compo_gen}).
Here, we used 4 GPUs per optimization with a batch size of 1 on each GPU. This means that on each GPU a single batch of 16 consecutive 2D images is rendered, consistent with the video diffusion model, which models 16-frame videos. Recall that, as discussed in \Cref{sec:compo_gen}, only four of those frames are also given to the regular text-to-image diffusion model. Hence, in each optimization step, 64 different images are rendered in total from the dynamic 4D scene.
We sample the diffusion time $t$ in the range $[0.02, 0.98]$ throughout the optimization process for the second stage.
We also run 10,000 optimization steps on average for this stage.

\begin{figure}[t!]
    \includegraphics[width=0.3\textwidth]{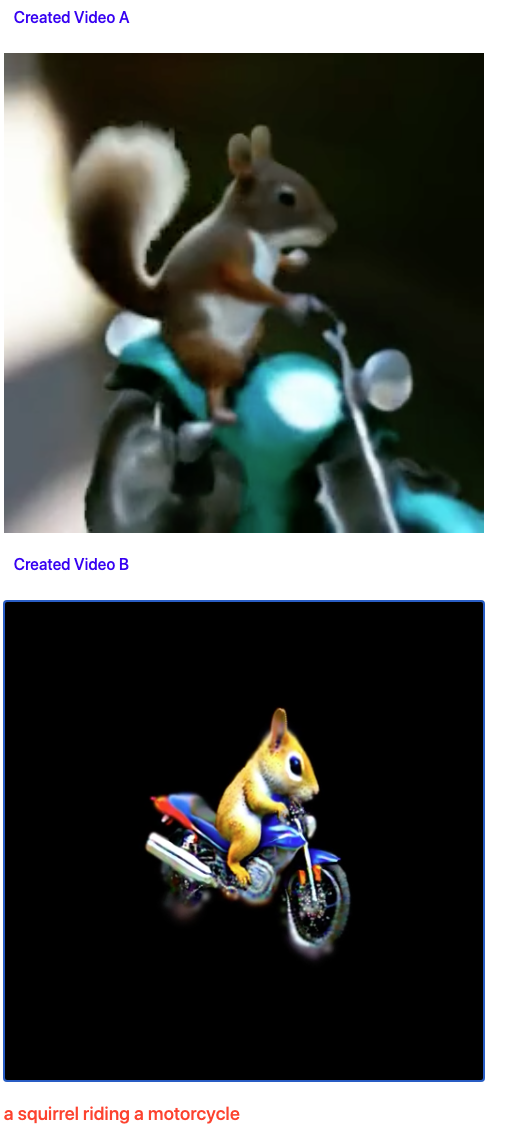}\\
    \includegraphics[width=1.03\textwidth]{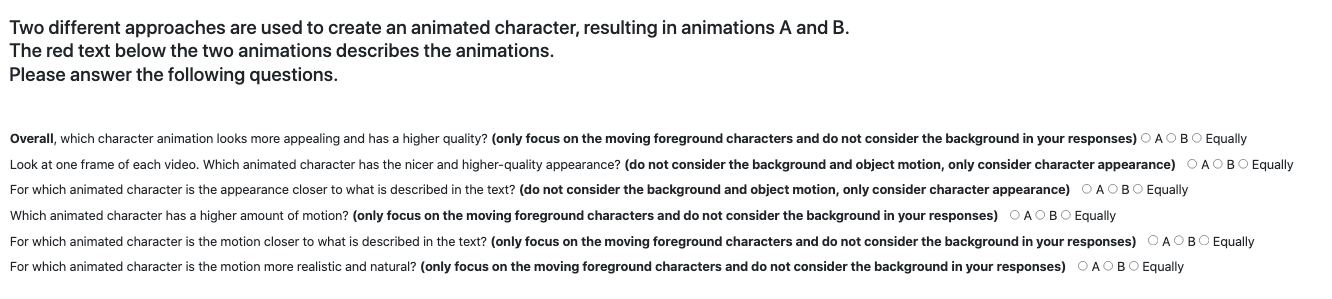}
    \vspace{-0.4cm}\caption{\small Screenshot of instructions provided to participants of the user studies for comparing AYG and MAV3D~\cite{singer2023mav3d} as well as for the ablation studies.}
    \label{fig:user_study_supp}
\end{figure}
\subsection{Evaluation Prompts} \label{sec:supp_prompts}
\textbf{For the baseline comparison} to MAV3D~\cite{singer2023mav3d}, we used all the 28 prompts from MAV3D's project page: 

\textit{``An alien playing the piano.";
``Shark swimming in the desert.";
``A dog wearing a Superhero outfit with red cape flying through the sky.";
``A monkey eating a candy bar.";
``A squirrel DJing.";
``A cat singing.";
``A bear driving a car.";
``Chihuahua running on the grass.";
``A human skeleton drinking wine.";
``A yorkie dog eating a donut";
``A baby panda eating ice cream";
``A kangaroo cooking a meal.";
``A humanoid robot playing the violin.";
``A squirrel playing the saxophone.";
``An octopus is underwater.";
``A silver humanoid robot flipping a coin.";
``A goat drinking beer.";
``A squirrel playing on a swing set.";
``A panda playing on a swing set.";
``A crocodile playing a drum set.";
``A squirrel riding a motorcycle.";
``3D rendering of a fox playing videogame.";
``A dog riding a skateboard.";
``An emoji of a baby panda reading a book.";
``Clown fish swimming through the coral reef.";
``A space shuttle launching.";
``A corgi playing with a ball.";
``A panda dancing."}

\textbf{For the ablation study}, we selected the following 30 text prompts:

\textit{``A cat singing.";
``A corgi playing with a ball.";
``A cow running fast.";
``A dog wearing a Superhero outfit with red cape flying through the sky.";
``A fox dressed in a suit dancing.";
``A monkey eating a candy bar.";
``A monkey is playing bass guitar.";
``A panda dancing.";
``A panda surfing a wave.";
``A pig running fast.";
``A purple unicorn flying.";
``A space shuttle launching.";
``A squirrel DJing.";
``A squirrel playing on a swing set.";
``A squirrel playing the saxophone.";
``A squirrel riding a motorcycle.";
``A storm trooper walking forward and vacuuming.";
``An alien playing the piano.";
``an astronaut is playing the electric guitar.";
``An astronaut riding a horse.";
``An astronaut riding motorcycle.";
``A panda reading a book.";
``Beer pouring into a glass.";
``Chihuahua running.";
``Clown fish swimming.";
``A dog riding a skateboard.";
``Flying dragon on fire.";
``Intricate butterfly flutters its wings.";
``Waves crashing against a lighthous.";
``Wood on fire."}

\subsection{User Study Details} \label{supp:user_study_details}
We conducted human evaluations (user studies) through Amazon Mechanical Turk to assess the quality of our generated 4D scenes, comparing them with MAV3D~\cite{singer2023mav3d} and performing ablation studies. 

For the MAV3D comparison, we used the 28 rendered videos from MAV3D's project page (\url{https://make-a-video3d.github.io/}) and compared them against our model (AYG) using identical text prompts (see \Cref{sec:supp_prompts} above). We rendered our dynamic 4D scenes from similar camera perspectives and created similar videos. We then asked the participants to compare the two videos with respect to 6 different categories and indicate preference for one of the methods with an option to vote for `equally good' in a non-forced-choice format. The 6 categories measure overall quality, 3D appearance and 3D text alignment, as well as motion amount, motion text alignment and motion realism (see questions in \Cref{fig:user_study_supp}).

In the ablation studies, we proceeded similarly. We showed participants 4D animations for 30 text prompts (see \Cref{sec:supp_prompts} above) generated by the full AYG model and by the modified, ablated models. Again, participants were asked to choose the more favorable 4D character from each pair, with an option to vote for `equally good'. 

For a visual reference, see \Cref{fig:user_study_supp} for a screenshot of the evaluation interface. In all user studies, the order of video pairs (A-B) was randomized for each question. Note that since MAV3D uses an extra background model, while AYG does not, we asked participants to focus only on the moving foreground characters and to not consider the background in their responses. In all user studies, each video pair was evaluated by seven participants, totaling 196 responses for the MAV3D comparison and 210 for each setup in the ablation study. We selected participants based on specific criteria: they had to be from English-speaking countries, have an acceptance rate above 95\%, and have completed over 1000 approved tasks on the platform. 

%% file: appendices/more_quantitative_results.tex
\section{Additional Quantitative Results}
\begin{table}[t!]
    \centering
        \caption{\small \textbf{R-Precision comparison to MAV3D~\cite{singer2023mav3d}} with the 300 text prompts also used by \citet{singer2023makeavideo} and \citet{singer2023mav3d}. \vspace{-0.2cm}}
    \label{tab:r-precision}
    \scalebox{0.95}{\rowcolors{2}{gray!15}{white}
    \begin{tabular}{l  c c c c}
        \toprule
         Method &  AYG (\textit{ours}) &   AYG (\textit{ours})  & MAV3D~\cite{singer2023mav3d}  & MAV3D~\cite{singer2023mav3d}   \\
          &  3D-stage &   4D-stage  & 3D-stage  & 4D-stage   \\
        \midrule
       R-Precision $\uparrow$ & 82.2 & 81.7  &82.4&  83.7  \\
        \bottomrule
    \end{tabular}}
\end{table}
\subsection{Comparisons to MAV3D and R-Precision Evaluation}\label{sec:supp_mav3d_comp}
The important MAV3D baseline~\cite{singer2023mav3d} did not release any code or models and its 4D score distillation leverages the large-scale Make-A-Video~\cite{singer2023makeavideo} text-to-video diffusion model, which is also not available publicly. This makes comparisons to MAV3D somewhat difficult and this is why we compared to MAV3D by using the available results on their project page. As reported in the main text, we performed a user study comparing all their generated 4D scenes with AYG's generated scenes with the same text prompts. 
AYG outperforms MAV3D in our user study in all categories (see \Cref{tab:user_study_mav_comparison} in main text). Moreover, our comparisons to MAV3D do not leverage any fine-tuning as discussed in \Cref{supp:extra_finetuning}. With this fine-tuning, our quality improvements over MAV3D are even larger, which would likely be reflected in an even higher preference for our 4D scenes in the user study.

MAV3D also reports R-Precision~\cite{park2021benchmark,jain2022dreamfields} in their paper.
R-Precision is commonly used in the text-to-3D literature as an evaluation metric. In the R-Precision calculation, 2D renderings of the scene are given to a CLIP~\cite{radford2021clip} image encoder and the CLIP encoding is then used to retrieve the closest text prompt among a set of text prompts used in the evaluation. The R-Precision value is then the top-1 retrieval accuracy, \ie, the percentage of correctly retrieved text prompts (this is, the text prompt which was used to generate the 3D scene is correctly retrieved). However, R-Precision measures only 3D quality (and more specifically the alignment of the 3D appearance with the text prompt) and does not in any way capture dynamics at all. It is therefore a meaningless metric to evaluate \textit{dynamic} scenes, which is the focus of our and also MAV3D's work.

Nevertheless, for future reference and simply to follow MAV3D we also provide R-Precision evaluation results in \Cref{tab:r-precision}. We obtained the list of 300 prompts used in MAV3D's R-Precision evaluation by the authors of MAV3D. Note that this list of prompts was originally collected not for the evaluation of synthesized 4D dynamic scenes but for the evaluation of video diffusion models in Make-A-Video~\cite{singer2023makeavideo}. To calculate our R-Precision scores with the 300 text prompts, we used the evaluation protocol from \url{https://github.com/Seth-Park/comp-t2i-dataset}. Similarly to MAV3D, we evaluated R-Precision both after the initial 3D stage and at random times $\tau$ of the dynamic 4D scene after the full 4D stage optimization (MAV3D similarly first optimizes a static 3D scene and only then adds an additional temporal dimension to generate a full dynamic 4D scene). Specifically, for 3D objects, we render with 20 different azimuth angles with a fixed elevation of 15 degree, camera distance of 3 and field-of-view of 40. For dynamic 4D scenes, we sample 20 times $\tau$ together with 20 different azimuth angles. We use majority voting over the 20 views as top-1 retrieval results for the R-Precision calculation. The results are shown in \Cref{tab:r-precision}.

We see that the two methods perform on par. %
Importantly, we do not know the exact evaluation protocol MAV3D used (\eg camera poses used for rendering), and in our experience these details matter and can influence the metric non-negligibly. Hence, considering that the results of the two methods are extremely close, we conclude that the two methods perform approximately similarly with respect to R-Precision. We also see that, for both methods, performance does not meaningfully differ between the 3D and 4D stage. This means that both methods preserve the overall 3D object well when learning dynamics in their main 4D stage.

We would like to stress again that R-Precision is in the end not very useful for the evaluation of dynamic 4D scenes, as it completely misses the important temporal aspect and does not judge scene dynamics and object motion. We believe that user studies are a more meaningful way to evaluate dynamic 4D scenes (also MAV3D performs various user studies in their paper). Recall that in our user studies, we outperform MAV3D on all categories. 

\subsection{Extended Discussion of Ablation Studies}\label{supp:extend_discussion_ablations}
\input{tables_tex/ablation_results_supp}
Here, we provide an extended discussion of our main ablation studies, originally presented in \Cref{tab:user_study_ablations} in \Cref{sec:experiments}. We have carried out an extensive number of ablations and there is not enough space in the paper's main text to discuss all of them in detail. For convenience, we copied the table here, see \Cref{tab:user_study_ablations_supp}, and we will now discuss the different settings one by one. Also see the supplementary video \texttt{ayg$\_$ablation$\_$study.mp4}, which shows dynamic 4D scenes for all ablations. Note, however, that these 4D scenes were optimized with only 4,000 optimization steps in the second dynamic 4D optimization stage in the interest of efficiency, considering that we had to optimize many 4D scenes for all ablations. The quality of the shown 4D scenes is therefore somewhat lower than that of our main results shown elsewhere in the paper and on the project page.

\textbf{Full AYG v.s. w/o rigidity regularization.} We can see a clear preference for the full AYG model compared to a variant without rigidity regularization. Users strongly prefer the full model in all categories in \Cref{tab:user_study_ablations_supp}. In the supplementary video we see unrealistic distortions of the object for the baseline without rigidity regularization. Such distortions are prevented by the regularization, as expected.

\textbf{Full AYG v.s. w/o motion amplifier.} Users prefer the full AYG variant that leverages the motion amplifier for all categories. The difference in preference is most pronounced in the ``Motion Amount'' category, which validates that the motion amplifier indeed amplifies motion. In the supplementary video, we can clearly observe enhanced motion compared to the baseline without the motion amplifier.

\textbf{Full AYG v.s. w/o initial 3D stage.} Without the initial 3D stage, simultaneously optimizing the 3D object itself and also distilling motion into the 4D scene results in unstable optimization behavior and no coherent dynamic 4D scenes can be produced. This is visible in the supplementary video and also validated in the user study. The full AYG model with the initial 3D stage is strongly preferred.\looseness=-1

\textbf{Full AYG v.s. w/o JSD-based regularization.} We can also observe a clear preference for the full AYG model including the JSD-based regularization of the distribution of the dynamic 3D Gaussians over an ablated model that does not use it. This is visible in all categories in \Cref{tab:user_study_ablations_supp}. In the supplementary video, we can see that without JSD-based regularization the 4D sequences show little motion and only some slow floating of the entire objects can be observed. We hypothesize that this slow global motion represents a local minimum of the video diffusion model, whose gradients are used to learn the dynamics. Falling into this local minimum is prevented through the JSD-based regularization. With JSD-based regularization, more complex, local motion is learnt instead of global translations or object size changes. 

\textbf{Full AYG v.s. w/o image DM score in 4D stage.} A central design decision of AYG is the simultaneous use of both a text-to-image diffusion model and a text-to-video diffusion model for score distillation in the main 4D optimization stage. Hence, we compared to a variant that only uses the video model, but we find that the full AYG approach including the image diffusion model score in the 4D stage is preferred, except for ``Motion Realism''. However, this is a somewhat ambiguous category, as people might have subjective opinions on what constitutes a better motion, such as preferring slow but consistent motion versus large but artifact-prone motion.
The full model wins on all other categories. The margin between the full and ablated model is especially large on the 3D appearance and 3D text alignment categories. This is exactly what we expected and why the text-to-image model is used, \ie, to ensure that high visual and 3D quality is maintained while the video model is responsible for the optimization of the temporal dynamics. In line with that, in the supplementary video we can observe some degradation in the 3D quality without the text-to-image diffusion model in the 4D stage. This justifies AYG's design composing a text-to-image and a text-to-video diffusion model for score distillation in the 4D stage.

\textbf{Full AYG v.s. SDS instead of CSD.} We have a somewhat similar observation when replacing classifier score distillation (CSD) with regular score distillation sampling (SDS). The full model is preferred for all categories, except for the more ambiguous ``Motion Realism''. In the supplementary video we see slightly more distorted 3D objects and less motion when using SDS instead of CSD.

\textbf{Full AYG v.s. 3D stage w/o MVDream.} Another central design decision of AYG is the use of the 3D-aware multiview-image diffusion model MVDream~\cite{shi2023mvdream} in its first 3D stage. In this ablation, we removed MVDream from the 3D stage and optimized the 3D assets only with the text-to-image Stable Diffusion model~\cite{rombach2021highresolution}, and then performed the dynamic 4D distillation. We find that users prefer the full model by a large margin on all categories. In short, AYG is not able to produce high-quality static 3D assets without MVDream, and then also the dynamic 4D optimization struggles because of the poor 3D initialization. It is worth pointing out, however, that there is a rich literature on score distillation of 3D assets using only text-to-image models and arguably some of these techniques could help. As pointed out previously, the initial 3D assets used by AYG in its main 4D stage could potentially also be produced by other methods.

\textbf{Full AYG v.s. 4D stage with MVDream.} The previous ablation makes one wonder whether it would help to also include MVDream in the 4D optimization stage to ensure that geometric and multiview consistency is maintained during 4D optimization at all times $\tau$ of the 4D sequences. However, we find that this is not the case. Users prefer the dynamic 4D scenes generated by the regular AYG model over the one that also includes MVDream in the 4D stage for all categories in the table. In the supplementary video, we see that including MVDream in the 4D stage can lead to odd motion patterns or overall reduced motion. We hypothesize that the video diffusion model and the MVDream multiview diffusion model produce somewhat conflicting gradients, harming the 4D optimization process and suppressing the learning of smooth 4D dynamics. Hyperparameter tuning and a carefully chosen weighting between the MVDream and video models in the composed score distillation scheme could potentially address this, but we were not able to make this setting work better than the regular AYG approach without MVDream in the 4D stage. Together with the previous ablation, we conclude that MVDream is a crucial component of AYG, but only in its initial 3D stage.

\textbf{Full AYG v.s. video model with only fps 4.} Our newly trained latent video diffusion model is conditioned on the frames-per-second (fps) frame rate and when optimizing dynamic scenes during AYG's 4D stage we sample different fps values. In this ablation, we ask what would happen if we only sampled the low $\textrm{fps}=4$ value, which corresponds to videos that show a lot of motion, but therefore are less smooth temporally (see the attached supplementary video \texttt{ayg$\_$new$\_$video$\_$model.mp4}). We find that the full AYG model that samples different fps conditionings during score distillation is overall preferred, although there is an outlier in the 3D appearance category. However, the varying fps conditionings in AYG primarily aim at producing better motion and not at improving 3D quality, and for the motion categories the main AYG model is generally preferred over the ablated version with only one $\textrm{fps}=4$ value. Visually, in the attached ablations video we observe slightly lower quality motion, in line with the results from the user study.

\textbf{Full AYG v.s. video model with only fps 12.} Here, we instead only use $\textrm{fps}=12$, corresponding to videos with less but smoother motion. We again see that the full AYG model is preferred over this ablated variant, this time in all categories including the 3D ones. In the supplementary video we can see significantly reduced motion in the dynamic 4D scenes when using this ablated AYG version. 

Note that these observations are in line with \citet{singer2023mav3d}, who also used an fps-conditioned video diffusion model and different fps during score distillation. Sampling different fps during 4D optimization helps both them and our AYG to produce better dynamic 4D scenes with higher-quality motion.

\textbf{Full AYG v.s. w/o dynamic cameras.} We can also observe the benefit of dynamic cameras. The full AYG model that includes dynamic cameras is preferred in all categories over the ablated model that uses static cameras when rendering the video frames given to the video diffusion model during score distillation in the 4D stage. In the supplementary video, we see that the dynamic 4D sequences generated without dynamic cameras have less motion. This is also consistent with \citet{singer2023mav3d}, who also observed a similar benefit of dynamic cameras.

\textbf{Full AYG v.s. w/o negative prompting.} Finally, we also studied the effect of negative prompt guidance during the 4D stage. Overall, the main model that includes negative prompting is preferred, although on ``Motion Amount'' the baseline without the negative prompt guidance is preferred. In the attached video, we observe lower quality dynamics without negative prompting.

\textbf{Conclusions.} Overall, our ablation studies show that all of AYG's components are important for generating high quality dynamic 4D scenes. On ``Overall Quality'', the main AYG model wins in all ablations with large margins. This justifies AYG's design choices.

\begin{table}[t!]
    \centering
        \caption{\small \textbf{Ablation study on view guidance} by user study on synthesized static 3D scenes from AYG's initial 3D stage. We used 30 text prompts, the same as in the other ablation studies. Numbers are percentages.} \vspace{-0.2cm}
    \label{tab:view_guidance_ablation}
    \scalebox{0.95}{\rowcolors{2}{gray!15}{white}
    \begin{tabular}{l  c c c}
        \toprule
         Method &  AYG w/ view guidance &  AYG w/o view guidance  & Equal  \\
        preference &   preferred &  preferred  &  preference \\
        \midrule
        Overall Quality & 33.3 & 33.8 & 32.9  \\
        3D Appearance & 28.6 & 29.0 & 42.4 \\
        3D Text Alignment & 30.0 & 34.8 & 35.2  \\
        \bottomrule
    \end{tabular}}
\end{table}
\subsection{View-guidance Ablation Study}
Our main ablation studies focused primarily on the 4D stage and we wanted to study the effects of the different components when learning dynamic 4D scenes, which is the main novelty of our paper. Here, we show an additional ablation study that analyzes the effect of view-guidance (\Cref{sec:scaling_ayg}), which we used only in AYG's initial 3D stage. We performed a user study using the same prompts and following the exact same protocol as in our other user studies for the other ablations (see \Cref{supp:user_study_details}). However, since view-guidance has only been used in the 3D stage, we only asked the users about overall quality, 3D appearance and 3D text alignment of the static 3D scenes synthesized after the initial 3D stage. We showed the users 3D objects generated with and without view guidance and asked them to compare and rate them. The results are shown in \Cref{tab:view_guidance_ablation}.
We see that AYG with or without view-guidance in its 3D stage performs approximately similar according to the user ratings and there is no clear winner in any of the categories.

We nevertheless used view guidance in AYG, as we subjectively found in early experiments that it sometimes helped with overall 3D appearance and led to a small reduction of the Janus-face problem. However, as the results of the user study here demonstrate, view guidance is certainly not one of the crucial components of AYG that make or break synthesis. However, to the best of our knowledge view guidance is a new idea and maybe it can find applications in future work.

%% file: tables_tex/ablation_results_supp.tex
\begin{table}[t!]
    \centering
    \caption{\small \textbf{Ablation study} by user study on synthesized 4D scenes with 30 text prompts. For each pair of numbers, the left number is the percentage that the full AYG model is preferred and the right number indicates preference percentage for ablated model as described in left column. The numbers do not add up to $100$ and the difference is due to users voting ``no preference'' (table copied here from main paper for extended discussion in \Cref{supp:extend_discussion_ablations}).} \vspace{-0.2cm}
    \label{tab:user_study_ablations_supp}
    \scalebox{0.95}{\rowcolors{2}{gray!15}{white}
    \begin{tabular}{l  c c c c c c}
        \toprule
        \rowcolor{white}
        Align Your Gaussians  &   Overall &  3D  & 3D Text & Motion  & Motion Text  & Motion  \\
        \rowcolor{white}
        (full model) &   Quality & Appearance  & Alignment &  Amount & Alignment &  Realism \\
        \midrule

         v.s. w/o rigidity regularization & \textbf{45.8}/13.3 & \textbf{43.3}/19.2  & \textbf{38.3}/15.0  & \textbf{40.8}/15.0 & \textbf{42.5}/18.3 & \textbf{30.8}/26.7 \\ 
         v.s. w/o motion amplifier    &  \textbf{43.3}/23.3 & \textbf{37.5}/28.3  & \textbf{30.8}/26.7  & \textbf{45.8}/10.8 & \textbf{37.5}/26.7 & \textbf{33.3}/31.7 \\

          v.s. w/o initial 3D stage    &  \textbf{67.5}/15.0 & \textbf{57.5}/21.7  & \textbf{64.2}/15.0  & \textbf{60.8}/21.7 & \textbf{60.8}/20.8 & \textbf{59.2}/24.2 \\

           v.s. w/o JSD-based regularization    &  \textbf{40.0}/25.0 & \textbf{40.0}/27.5  & \textbf{36.7}/27.5  & \textbf{41.7}/24.2 & \textbf{39.2}/29.2 & \textbf{45.0}/24.2 \\

           v.s. w/o image DM score in 4D stage    &  \textbf{42.5}/22.5 & \textbf{39.2}/27.5  & \textbf{36.7}/25.8  & \textbf{33.3}/25.9 & \textbf{37.5}/30.0 & 27.5/\textbf{40.0} \\

          v.s. SDS instead of CSD    &  \textbf{44.2}/35.8 & \textbf{40.0}/27.5  & \textbf{35.8}/35.0  & \textbf{35.0}/27.5 & \textbf{35.0}/34.2 & 32.5/\textbf{35.8} \\

        v.s. 3D stage w/o MVDream    &  \textbf{66.7}/21.7 & \textbf{48.3}/34.2  & \textbf{38.3}/34.2  & \textbf{41.7}/22.5 &  \textbf{40.0}/27.5 & \textbf{40.8}/27.5 \\

        v.s. 4D stage with MVDream    &  \textbf{50.8}/27.5 &  \textbf{38.3}/34.2  &  \textbf{41.6}/29.2  & \textbf{39.2}/35.0 &  \textbf{44.2}/30.0 & \textbf{39.2}/31.7 \\

        v.s. video model with only fps 4   &  \textbf{46.7}/15.8 & 27.5/\textbf{36.7} & \textbf{30.0}/23.3  & \textbf{36.7}/30.0 & \textbf{31.7}/26.7 & \textbf{32.5}/28.3 \\

        v.s. video model with only fps 12    &  \textbf{48.3}/29.2 &  \textbf{30.8}/29.2 & \textbf{29.2}/28.3  & \textbf{35.0}/28.3 & \textbf{35.0}/30.0 & \textbf{39.2}/26.7 \\

        v.s. w/o dynamic cameras    &  \textbf{32.5}/25.0 &  \textbf{32.5}/31.7 & \textbf{35.0}/33.3  & \textbf{35.0}/32.5 &\textbf{ 35.8}/33.3 &  \textbf{32.5}/25.0 \\

         v.s. w/o negative prompting    &  \textbf{44.2}/28.3 &  \textbf{38.3}/32.5 & \textbf{31.7}/29.2  &29.2/\textbf{31.6} & \textbf{33.3}/30.0 &  \textbf{37.5}/28.3 \\
        \bottomrule
    \end{tabular}
    }
\end{table}

%% file: appendices/more_qualitative_results.tex
\section{Additional Qualitative Results---More AYG Samples}
Here, we show additional generated dynamic 4D scene samples from AYG. We also refer the reader to our supplementary video \texttt{ayg$\_$text$\_$to$\_$4d.mp4}, which shows almost all of our dynamic 4D scene samples. We also share videos generated by AYG's newly trained latent video diffusion model for this work.

\subsection{Text-to-4D Samples}
\input{figures_tex/supp_main_text_to_4d_results_part1}
\input{figures_tex/supp_main_text_to_4d_results_part3}
\input{figures_tex/supp_autoreg}
\input{figures_tex/supp_main_text_to_4d_results_part2}
In \Cref{fig:supp_ayg_samples1,fig:supp_ayg_samples2,fig:supp_ayg_samples3}, we show additional text-to-4D samples from AYG, similar to \Cref{fig:main_results} in the main paper.

\subsection{Autoregressively Extended and Looping Text-to-4D Synthesis}
In \Cref{fig:supp_autoreg_loopy_results}, we show additional samples from AYG that are autoregressively extended to form longer sequences while changing the text prompt and that return to the initial pose to enable looping animations (similar to \Cref{fig:autoreg_loopy_results} in the main paper).
For the first two rows (the \textit{assassin}) in \Cref{fig:supp_autoreg_loopy_results}, we use the following sequence of text prompts:
``Assassin with sword running fast, portrait, game, unreal, 4K, HD.'',
``Assassin walking, portrait, game, unreal, 4K, HD.'' and
``Assassin dancing, portrait, game, unreal, 4K, HD.''.
For the next two rows (the \textit{lion}), we use the following sequence of text prompts:
``A lion running fast.'',
``A lion is jumping.'' and
``A lion is eating.''.

For reference, we also provide the prompts used for \Cref{fig:autoreg_loopy_results} in the main paper. 
For the first two rows (the \textit{bulldog}), we use the following sequence of text prompts: 
``A bulldog is running fast.'' and
``A bulldog barking loudly''.
For the next two rows (the \textit{panda}), we use the following sequence of text prompts: 
``A panda running.'' and
``A panda is boxing and punching.''.

\subsection{More Comparisons to Make-A-Video3D}
\input{figures_tex/supp_meta_comparison}
In \Cref{fig:supp_meta_comparison}, we show additional visual comparisons to MAV3D~\cite{singer2023mav3d}, similar to \Cref{fig:meta_comparison} in the main paper.

\subsection{Videos Generated by AYG's fps-conditioned Video Diffusion Model} \label{sec:supp:video_model_samples}
In \Cref{fig:supp_video_samples1,fig:supp_video_samples2,fig:supp_video_samples3}, we present videos generated by AYG's latent video diffusion model, showing the effect of the fps conditioning. We also refer to the attached supplementary video \texttt{ayg$\_$new$\_$video$\_$model.mp4}, which shows more samples.
\input{figures_tex/supp_videomodel_1}
\input{figures_tex/supp_videomodel_2}
\input{figures_tex/supp_videomodel_3}

%% file: figures_tex/supp_main_text_to_4d_results_part1.tex
\begin{figure}[t!]
  \vspace{-0cm}
    \begin{center}
    \includegraphics[width=0.65\textwidth]{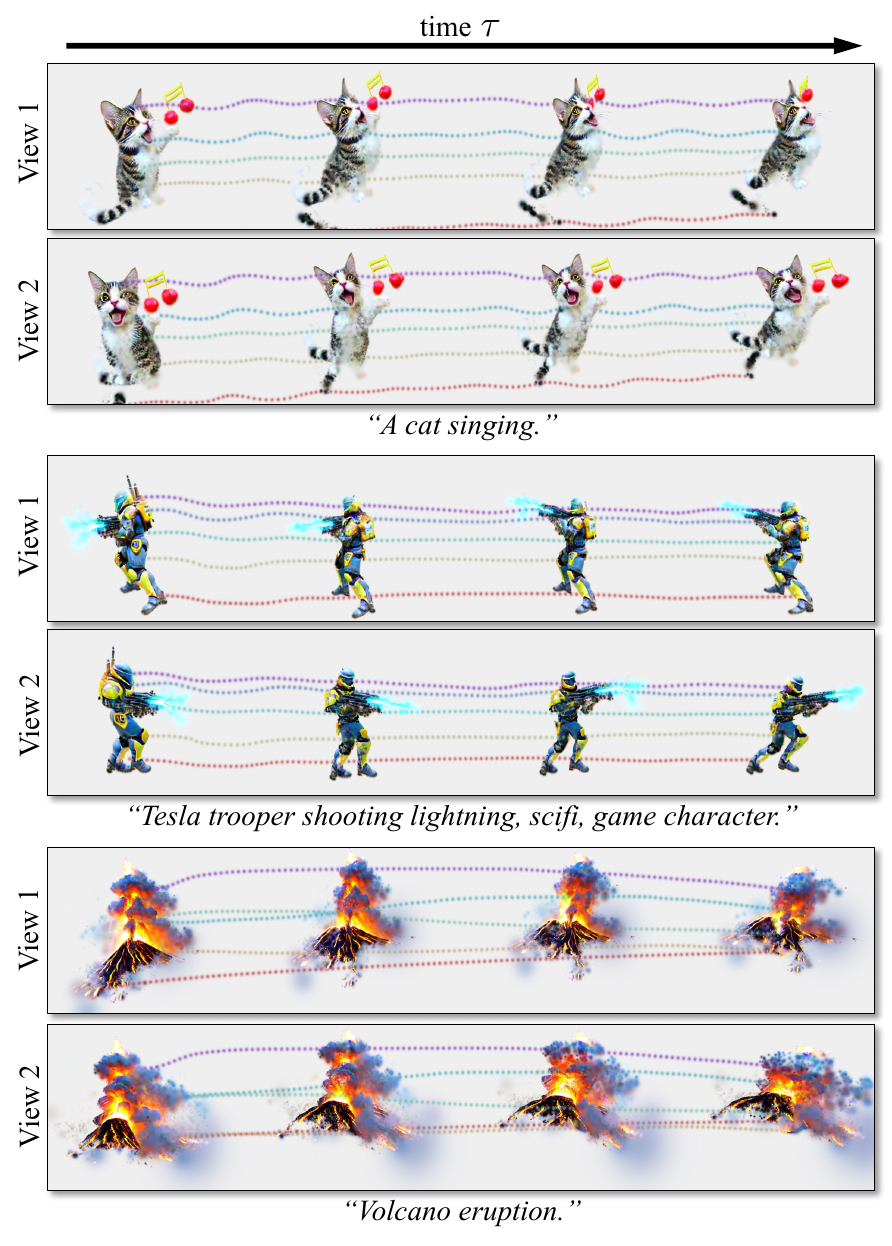}
    \end{center}
    \vspace{-3mm}
    \caption{\small \textbf{Text-to-4D synthesis with AYG.} Various samples shown in two views each. Dotted lines denote deformation field dynamics (also see supplementary video \texttt{ayg$\_$text$\_$to$\_$4d.mp4}, where the dynamics are much better visible).}
    \label{fig:supp_ayg_samples1}
  \vspace{-0mm}
\end{figure}

%% file: figures_tex/supp_main_text_to_4d_results_part3.tex
\begin{figure}[t!]
  \vspace{-0cm}
    \begin{center}
    \includegraphics[width=0.65\textwidth]{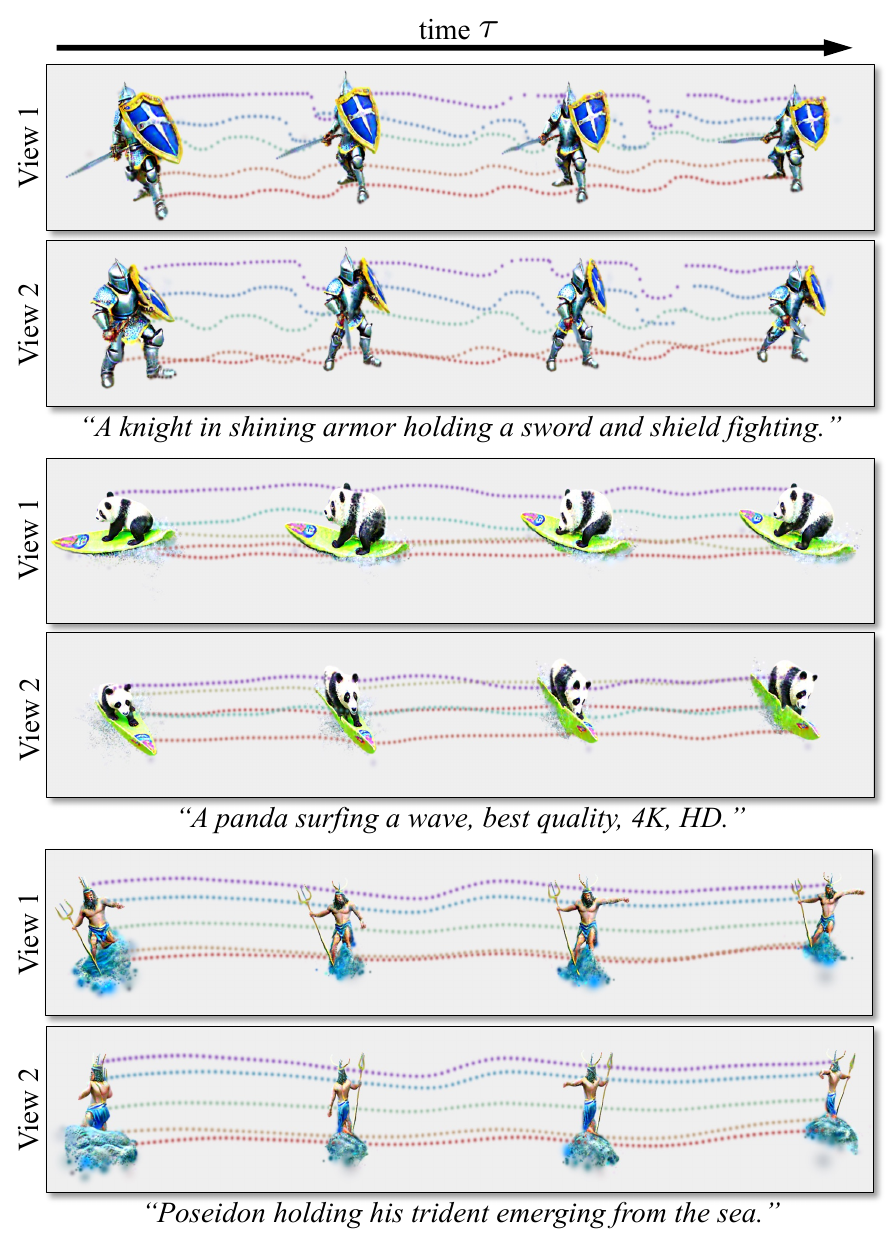}
    \end{center}
    \vspace{-3mm}
    \caption{\small \textbf{Text-to-4D synthesis with AYG.} Various samples shown in two views each. Dotted lines denote deformation field dynamics (also see supplementary video \texttt{ayg$\_$text$\_$to$\_$4d.mp4}, where the dynamics are much better visible).}
    \label{fig:supp_ayg_samples3}
  \vspace{-0mm}
\end{figure}

%% file: figures_tex/supp_autoreg.tex
\begin{figure}[t!]
  \vspace{-0cm}
    \begin{center}
    \includegraphics[width=0.8\textwidth]{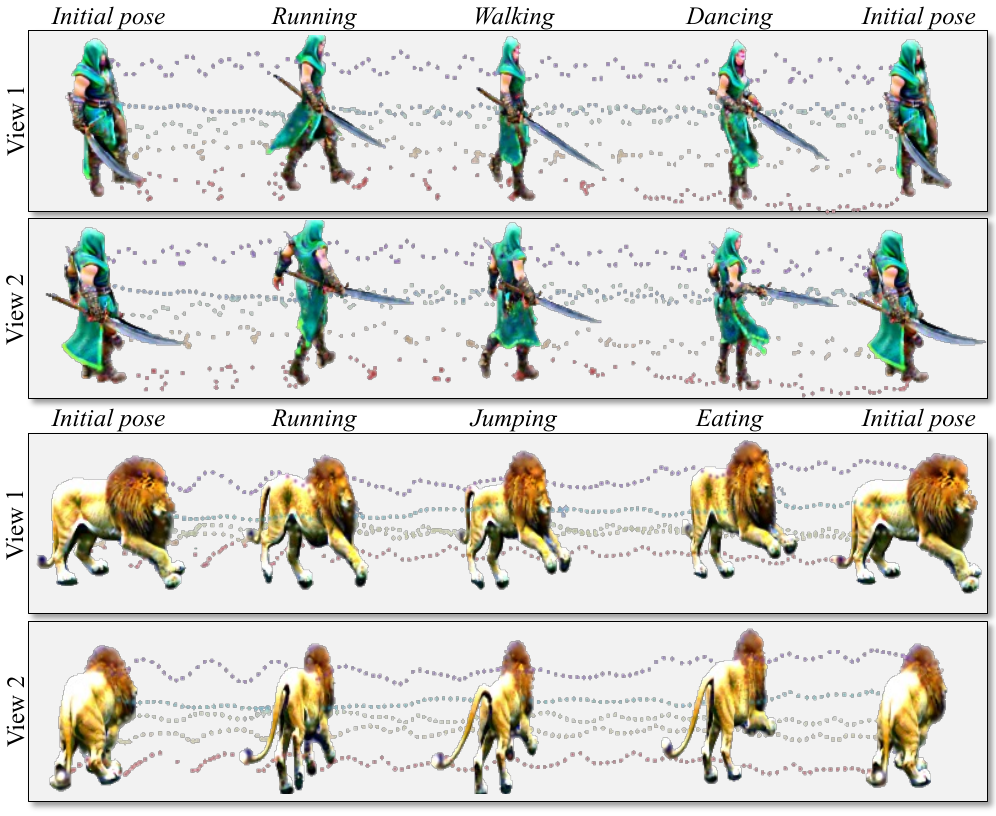}
    \end{center}
    \vspace{-3mm}
    \caption{\small \textbf{Autoregressively extended text-to-4D synthesis.} AYG is able to autoregressively extend dynamic 4D sequences, combine sequences with different text-guidance, and create looping animations, returning to the initial pose (also see supplementary video \texttt{ayg$\_$text$\_$to$\_$4d.mp4}, where the different actions are much better visible).}
    \label{fig:supp_autoreg_loopy_results}
  \vspace{-0mm}
\end{figure}

%% file: figures_tex/supp_main_text_to_4d_results_part2.tex
\begin{figure}[t!]
  \vspace{-0cm}
    \begin{center}
    \includegraphics[width=0.65\textwidth]{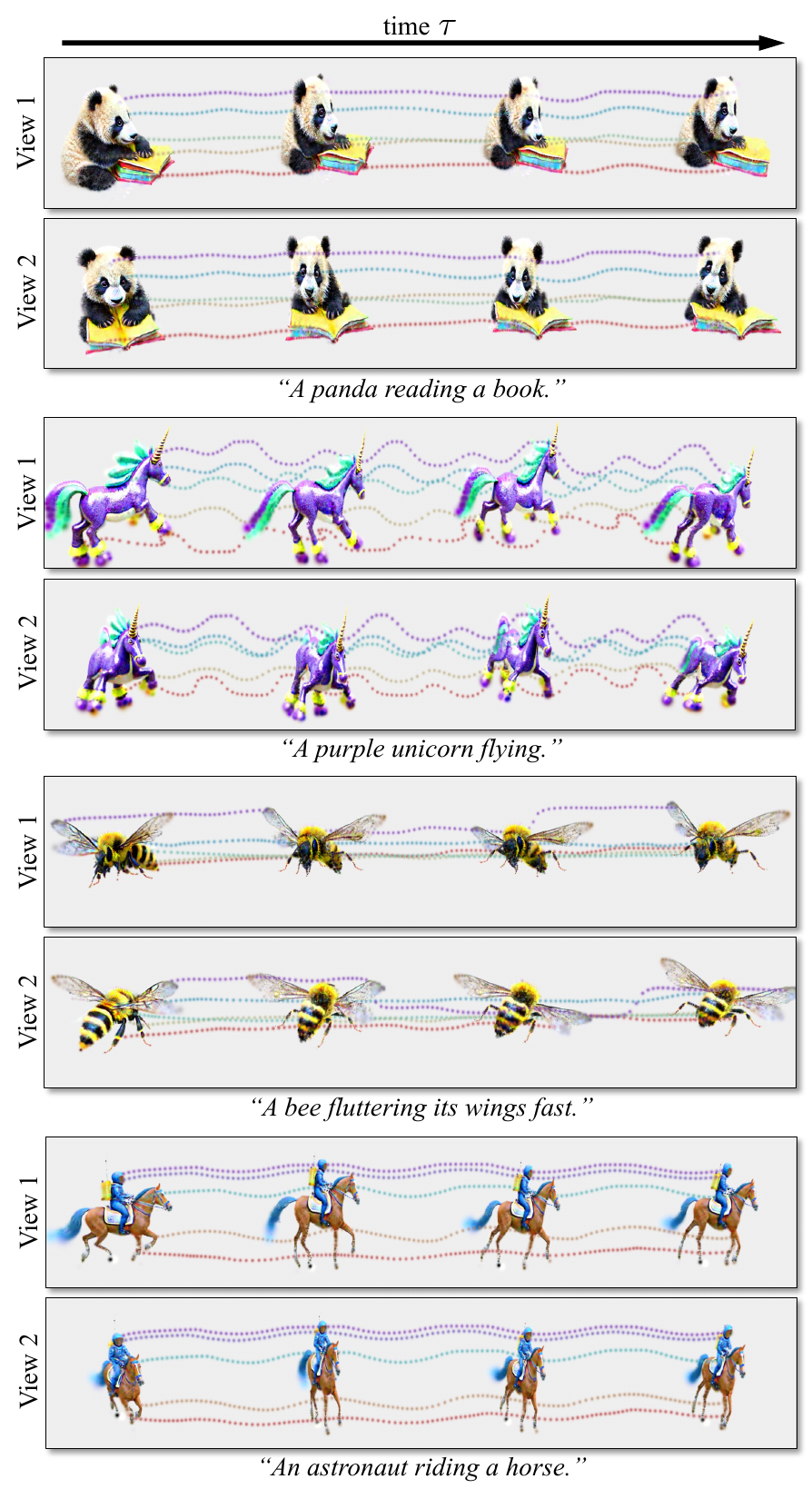}
    \end{center}
    \vspace{-3mm}
    \caption{\small \textbf{Text-to-4D synthesis with AYG.} Various samples shown in two views each. Dotted lines denote deformation field dynamics (also see supplementary video \texttt{ayg$\_$text$\_$to$\_$4d.mp4}, where the dynamics are much better visible).}
    \label{fig:supp_ayg_samples2}
  \vspace{-0mm}
\end{figure}

%% file: figures_tex/supp_meta_comparison.tex
\begin{figure}[t!]
  \vspace{-0cm}
    \begin{center}
    \includegraphics[width=0.48\textwidth]{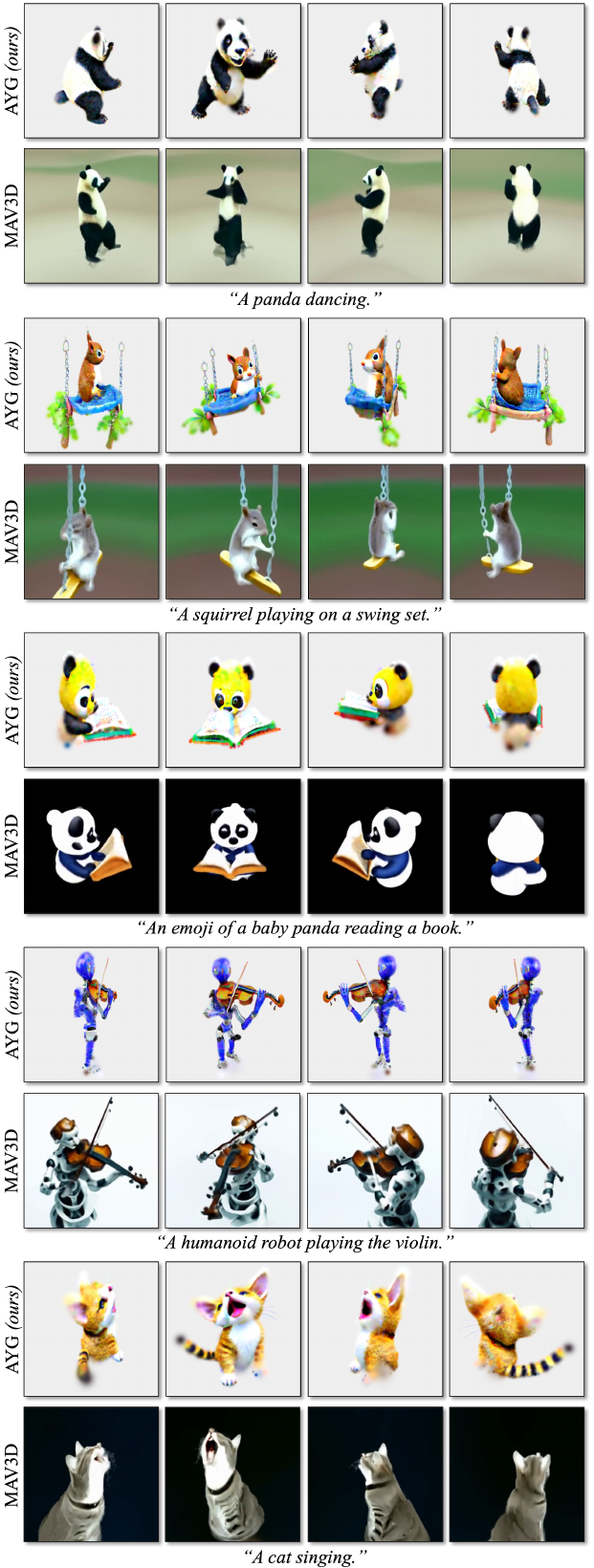}
    \end{center}
    \vspace{-5mm}
    \caption{\small \textbf{AYG (\textit{ours}) vs. MAV3D~\cite{singer2023mav3d}.} We show four 4D frames for different times and camera angles (also see supplementary video \texttt{ayg$\_$text$\_$to$\_$4d.mp4}, where we also show comparisons to MAV3D and where the dynamics are much better visible).}
    \label{fig:supp_meta_comparison}
  \vspace{-0mm}
\end{figure}

%% file: figures_tex/supp_videomodel_1.tex
\begin{figure}[t!]
  \vspace{-0cm}
    \begin{center}
    \includegraphics[width=0.7\textwidth]{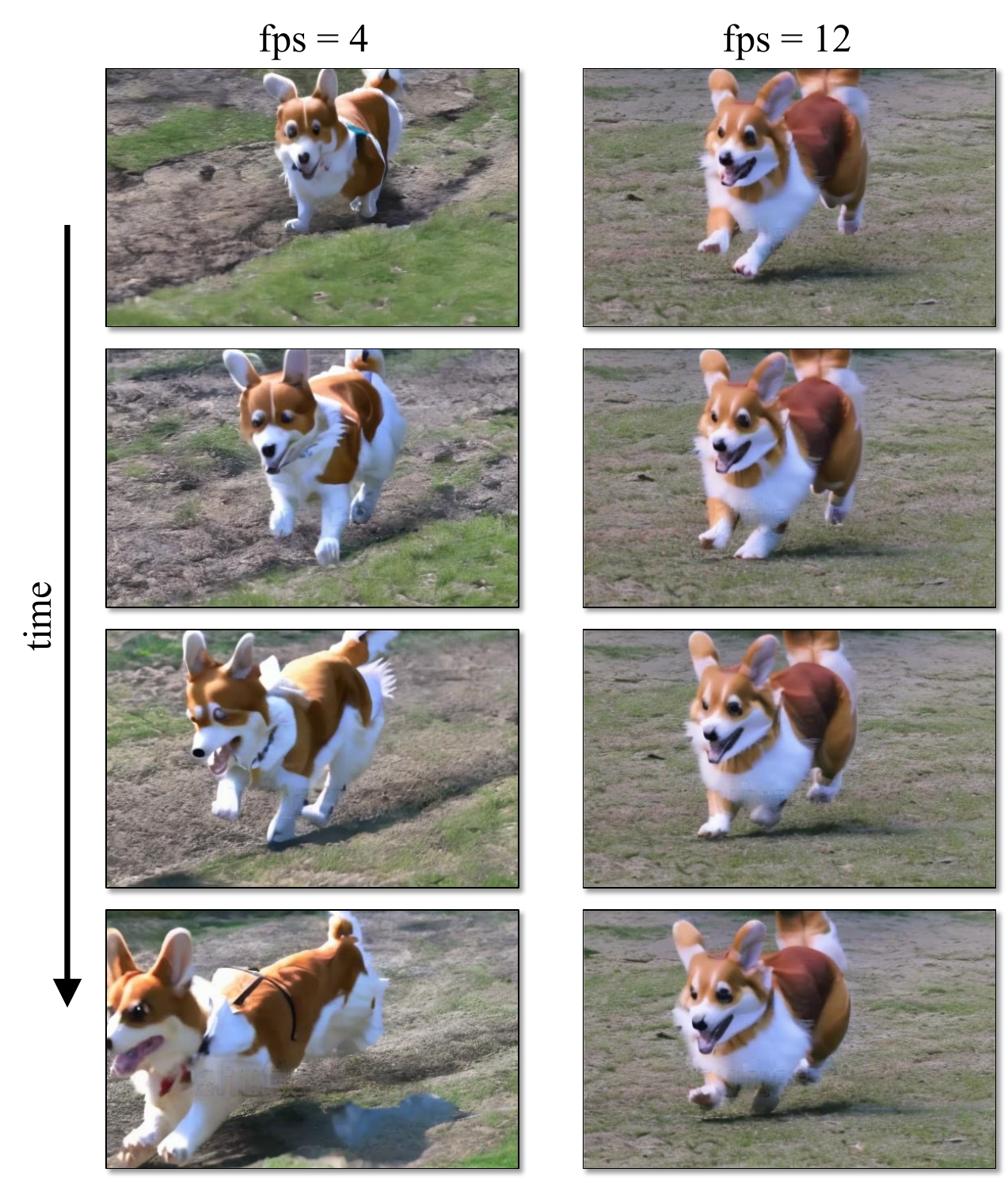}
    \end{center}
    \vspace{-2mm}
    \caption{\small \textbf{Two video samples from AYG's newly trained latent text-to-video diffusion model} for the same text prompt \textit{``A corgi running.''} but with different fps conditionings $\textrm{fps}=4$ and $\textrm{fps}=12$. We see that, as expected, conditioning on the lower fps value generates a video with more motion for the same 4 frames (the model synthesizes 16 frames and we show the 1st, the 6th, the 11th, and the 16th frame). Conditioning on the higher fps value results in a video with less motion but good temporal consistency.}
    \label{fig:supp_video_samples1}
  \vspace{-0mm}
\end{figure}

%% file: figures_tex/supp_videomodel_2.tex
\begin{figure}[t!]
  \vspace{-0cm}
    \begin{center}
    \includegraphics[width=0.7\textwidth]{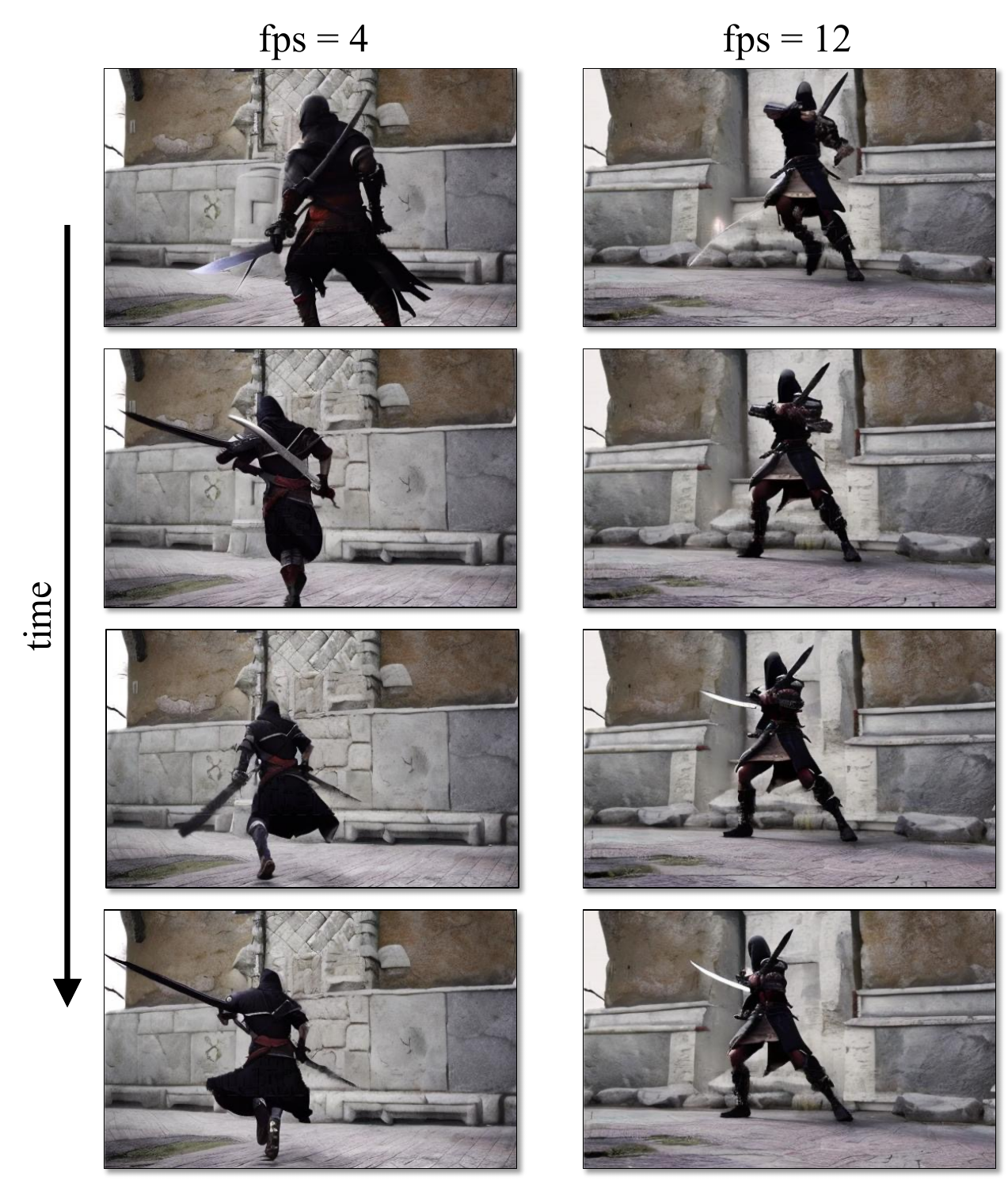}
    \end{center}
    \vspace{-2mm}
    \caption{\small \textbf{Two video samples from AYG's newly trained latent text-to-video diffusion model} for the same text prompt \textit{``Assassin with sword running fast, portrait, game, unreal, 4K, HD.''} but with different fps conditionings $\textrm{fps}=4$ and $\textrm{fps}=12$. We see that, as expected, conditioning on the lower fps value generates a video with more motion for the same 4 frames (the model synthesizes 16 frames and we show the 1st, the 6th, the 11th, and the 16th frame). Conditioning on the higher fps value results in a video with less motion but good temporal consistency.}
    \label{fig:supp_video_samples2}
  \vspace{-0mm}
\end{figure}

%% file: figures_tex/supp_videomodel_3.tex
\begin{figure}[t!]
  \vspace{-0cm}
    \begin{center}
    \includegraphics[width=0.7\textwidth]{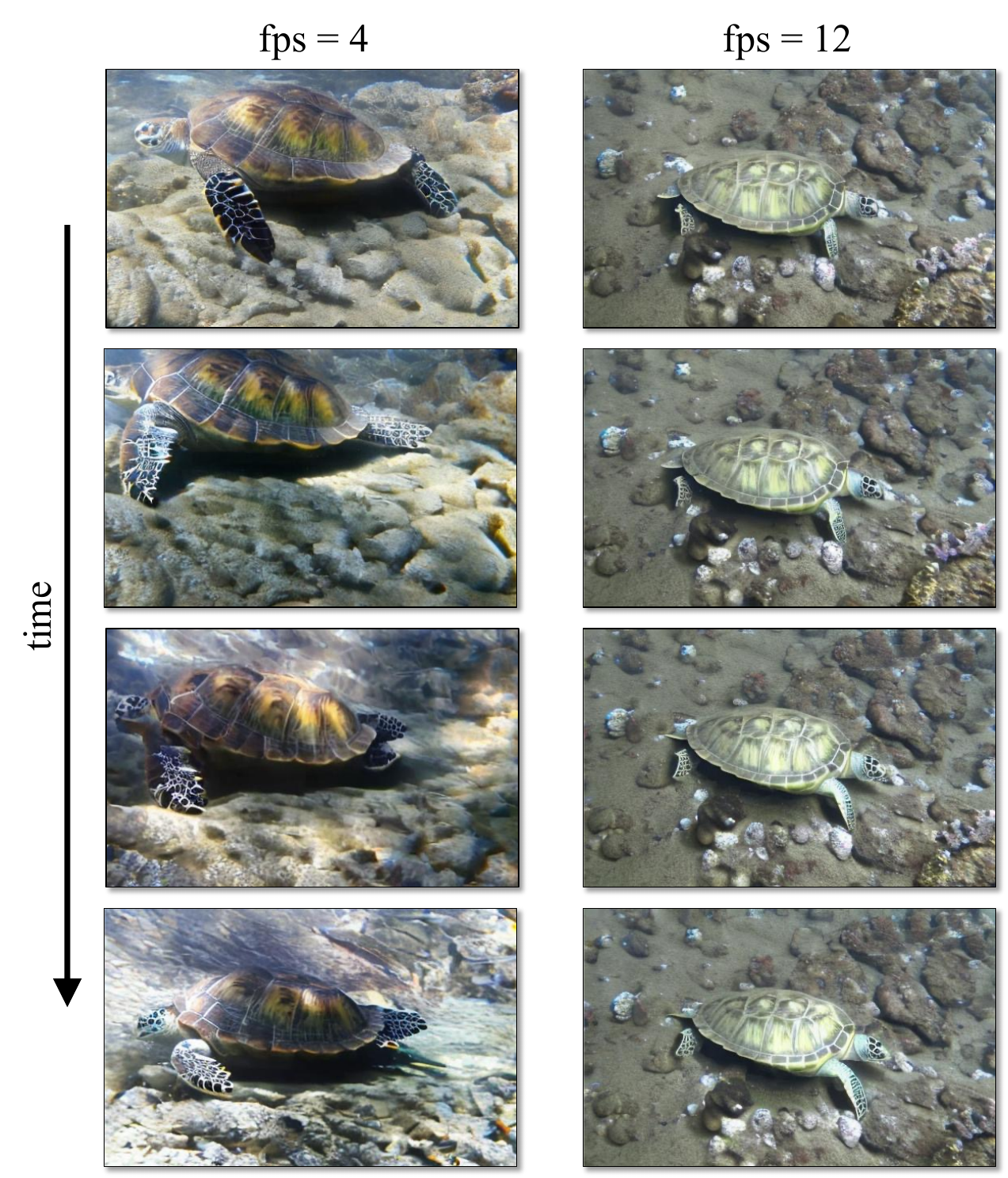}
    \end{center}
    \vspace{-2mm}
    \caption{\small \textbf{Two video samples from AYG's newly trained latent text-to-video diffusion model} for the same text prompt \textit{``A turtle swimming.''} but with different fps conditionings $\textrm{fps}=4$ and $\textrm{fps}=12$. We see that, as expected, conditioning on the lower fps value generates a video with more motion for the same 4 frames (the model synthesizes 16 frames and we show the 1st, the 6th, the 11th, and the 16th frame). Conditioning on the higher fps value results in a video with less motion but good temporal consistency.}
    \label{fig:supp_video_samples3}
  \vspace{-0mm}
\end{figure}

%% file: ms.bbl
\begin{thebibliography}{111}
\providecommand{\natexlab}[1]{#1}
\providecommand{\url}[1]{\texttt{#1}}
\expandafter\ifx\csname urlstyle\endcsname\relax
  \providecommand{\doi}[1]{doi: #1}\else
  \providecommand{\doi}{doi: \begingroup \urlstyle{rm}\Url}\fi

\bibitem[An et~al.(2023)An, Zhang, Yang, Gupta, Huang, Luo, and
  Yin]{an2023latentshift}
Jie An, Songyang Zhang, Harry Yang, Sonal Gupta, Jia-Bin Huang, Jiebo Luo, and
  Xi Yin.
\newblock {Latent-Shift: Latent Diffusion with Temporal Shift for Efficient
  Text-to-Video Generation}.
\newblock \emph{arXiv preprint arXiv:2304.08477}, 2023.

\bibitem[Ba et~al.(2016)Ba, Kiros, and Hinton]{ba2016layer}
Jimmy~Lei Ba, Jamie~Ryan Kiros, and Geoffrey~E Hinton.
\newblock {Layer Normalization}.
\newblock \emph{arXiv preprint arXiv:1607.06450}, 2016.

\bibitem[Bahmani et~al.(2023)Bahmani, Skorokhodov, Rong, Wetzstein, Guibas,
  Wonka, Tulyakov, Park, Tagliasacchi, and Lindell]{bahmani20234dfy}
Sherwin Bahmani, Ivan Skorokhodov, Victor Rong, Gordon Wetzstein, Leonidas
  Guibas, Peter Wonka, Sergey Tulyakov, Jeong~Joon Park, Andrea Tagliasacchi,
  and David~B. Lindell.
\newblock {4D-fy: Text-to-4D Generation Using Hybrid Score Distillation
  Sampling}.
\newblock \emph{arXiv preprint arXiv:2311.17984}, 2023.

\bibitem[Bain et~al.(2021)Bain, Nagrani, Varol, and Zisserman]{bain21frozen}
Max Bain, Arsha Nagrani, G{\"u}l Varol, and Andrew Zisserman.
\newblock {Frozen in Time: A Joint Video and Image Encoder for End-to-End
  Retrieval}.
\newblock In \emph{Proceedings of the IEEE/CVF International Conference on
  Computer Vision (ICCV)}, 2021.

\bibitem[Balaji et~al.(2022)Balaji, Nah, Huang, Vahdat, Song, Zhang, Kreis,
  Aittala, Aila, Laine, Catanzaro, Karras, and Liu]{balaji2022ediffi}
Yogesh Balaji, Seungjun Nah, Xun Huang, Arash Vahdat, Jiaming Song, Qinsheng
  Zhang, Karsten Kreis, Miika Aittala, Timo Aila, Samuli Laine, Bryan
  Catanzaro, Tero Karras, and Ming-Yu Liu.
\newblock {eDiff-I: Text-to-Image Diffusion Models with Ensemble of Expert
  Denoisers}.
\newblock \emph{arXiv preprint arXiv:2211.01324}, 2022.

\bibitem[Bautista et~al.(2022)Bautista, Guo, Abnar, Talbott, Toshev, Chen,
  Dinh, Zhai, Goh, Ulbricht, Dehghan, and Susskind]{bautista2022gaudi}
Miguel~{\'A}ngel Bautista, Pengsheng Guo, Samira Abnar, Walter Talbott,
  Alexander~T Toshev, Zhuoyuan Chen, Laurent Dinh, Shuangfei Zhai, Hanlin Goh,
  Daniel Ulbricht, Afshin Dehghan, and Joshua~M. Susskind.
\newblock {{GAUDI}: A Neural Architect for Immersive 3D Scene Generation}.
\newblock In \emph{Advances in Neural Information Processing Systems
  (NeurIPS)}, 2022.

\bibitem[Blattmann et~al.(2023)Blattmann, Rombach, Ling, Dockhorn, Kim, Fidler,
  and Kreis]{blattmann2023videoldm}
Andreas Blattmann, Robin Rombach, Huan Ling, Tim Dockhorn, Seung~Wook Kim,
  Sanja Fidler, and Karsten Kreis.
\newblock {Align your Latents: High-Resolution Video Synthesis with Latent
  Diffusion Models}.
\newblock In \emph{Proceedings of the IEEE/CVF Conference on Computer Vision
  and Pattern Recognition (CVPR)}, 2023.

\bibitem[Cai et~al.(2022)Cai, Feng, Feng, Wang, and Zhang]{Cai2022NDR}
Hongrui Cai, Wanquan Feng, Xuetao Feng, Yan Wang, and Juyong Zhang.
\newblock {Neural Surface Reconstruction of Dynamic Scenes with Monocular RGB-D
  Camera}.
\newblock In \emph{Thirty-sixth Conference on Neural Information Processing
  Systems (NeurIPS)}, 2022.

\bibitem[Cao and Johnson(2023)]{Cao2022hexplane}
Ang Cao and Justin Johnson.
\newblock {HexPlane: A Fast Representation for Dynamic Scenes}.
\newblock In \emph{Proceedings of the IEEE/CVF Conference on Computer Vision
  and Pattern Recognition (CVPR)}, 2023.

\bibitem[Chen et~al.(2023{\natexlab{a}})Chen, Chen, Jiao, and
  Jia]{chen2023fantasia3d}
Rui Chen, Yongwei Chen, Ningxin Jiao, and Kui Jia.
\newblock {Fantasia3D: Disentangling Geometry and Appearance for High-quality
  Text-to-3D Content Creation}.
\newblock In \emph{Proceedings of the IEEE/CVF International Conference on
  Computer Vision (ICCV)}, 2023{\natexlab{a}}.

\bibitem[Chen et~al.(2023{\natexlab{b}})Chen, Zhang, Yang, Cai, Yu, Yang, and
  Lin]{chen2023it3d}
Yiwen Chen, Chi Zhang, Xiaofeng Yang, Zhongang Cai, Gang Yu, Lei Yang, and
  Guosheng Lin.
\newblock {IT3D: Improved Text-to-3D Generation with Explicit View Synthesis}.
\newblock \emph{arXiv preprint arXiv:2308.11473}, 2023{\natexlab{b}}.

\bibitem[Chen et~al.(2023{\natexlab{c}})Chen, Wang, and Liu]{chen2023gsgen}
Zilong Chen, Feng Wang, and Huaping Liu.
\newblock {Text-to-3D using Gaussian Splatting}.
\newblock \emph{arXiv preprint arXiv:2309.16585}, 2023{\natexlab{c}}.

\bibitem[Chung et~al.(2023)Chung, Lee, Nam, Lee, and
  Lee]{chung2023luciddreamer}
Jaeyoung Chung, Suyoung Lee, Hyeongjin Nam, Jaerin Lee, and Kyoung~Mu Lee.
\newblock {LucidDreamer: Domain-free Generation of 3D Gaussian Splatting
  Scenes}.
\newblock \emph{arXiv preprint arXiv:2311.13384}, 2023.

\bibitem[Dai et~al.(2023)Dai, Hou, Ma, Tsai, Wang, Wang, Zhang, Vandenhende,
  Wang, Dubey, Yu, Kadian, Radenovic, Mahajan, Li, Zhao, Petrovic, Singh,
  Motwani, Wen, Song, Sumbaly, Ramanathan, He, Vajda, and Parikh]{dai2023emu}
Xiaoliang Dai, Ji Hou, Chih-Yao Ma, Sam Tsai, Jialiang Wang, Rui Wang, Peizhao
  Zhang, Simon Vandenhende, Xiaofang Wang, Abhimanyu Dubey, Matthew Yu,
  Abhishek Kadian, Filip Radenovic, Dhruv Mahajan, Kunpeng Li, Yue Zhao, Vladan
  Petrovic, Mitesh~Kumar Singh, Simran Motwani, Yi Wen, Yiwen Song, Roshan
  Sumbaly, Vignesh Ramanathan, Zijian He, Peter Vajda, and Devi Parikh.
\newblock {Emu: Enhancing Image Generation Models Using Photogenic Needles in a
  Haystack}.
\newblock \emph{arXiv preprint arXiv:2309.15807}, 2023.

\bibitem[Deitke et~al.(2023{\natexlab{a}})Deitke, Liu, Wallingford, Ngo,
  Michel, Kusupati, Fan, Laforte, Voleti, Gadre, VanderBilt, Kembhavi,
  Vondrick, Gkioxari, Ehsani, Schmidt, and Farhadi]{deitke2023objaverseXL}
Matt Deitke, Ruoshi Liu, Matthew Wallingford, Huong Ngo, Oscar Michel, Aditya
  Kusupati, Alan Fan, Christian Laforte, Vikram Voleti, Samir~Yitzhak Gadre,
  Eli VanderBilt, Aniruddha Kembhavi, Carl Vondrick, Georgia Gkioxari, Kiana
  Ehsani, Ludwig Schmidt, and Ali Farhadi.
\newblock {Objaverse-XL: A Universe of 10M+ 3D Objects}.
\newblock \emph{arXiv preprint arXiv:2307.05663}, 2023{\natexlab{a}}.

\bibitem[Deitke et~al.(2023{\natexlab{b}})Deitke, Schwenk, Salvador, Weihs,
  Michel, VanderBilt, Schmidt, Ehsani, Kembhavi, and
  Farhadi]{deitke2023objaverse}
Matt Deitke, Dustin Schwenk, Jordi Salvador, Luca Weihs, Oscar Michel, Eli
  VanderBilt, Ludwig Schmidt, Kiana Ehsani, Aniruddha Kembhavi, and Ali
  Farhadi.
\newblock {Objaverse: A Universe of Annotated 3D Objects}.
\newblock In \emph{Proceedings of the IEEE/CVF Conference on Computer Vision
  and Pattern Recognition (CVPR)}, 2023{\natexlab{b}}.

\bibitem[Deng et~al.(2023)Deng, Jiang, Qi, Yan, Zhou, Guibas, and
  Anguelov]{deng2023nerdi}
C. Deng, C. Jiang, C.~R. Qi, X. Yan, Y. Zhou, L. Guibas, and D. Anguelov.
\newblock {NeRDi: Single-View NeRF Synthesis with Language-Guided Diffusion as
  General Image Priors}.
\newblock In \emph{Proceedings of the IEEE/CVF Conference on Computer Vision
  and Pattern Recognition (CVPR)}, 2023.

\bibitem[Dhariwal and Nichol(2021)]{dhariwal2021diffusion}
Prafulla Dhariwal and Alexander~Quinn Nichol.
\newblock {Diffusion Models Beat {GAN}s on Image Synthesis}.
\newblock In \emph{Advances in Neural Information Processing Systems}, 2021.

\bibitem[Du et~al.(2023)Du, Durkan, Strudel, Tenenbaum, Dieleman, Fergus,
  Sohl-Dickstein, Doucet, and Grathwohl]{du2023reduce}
Yilun Du, Conor Durkan, Robin Strudel, Joshua~B. Tenenbaum, Sander Dieleman,
  Rob Fergus, Jascha Sohl-Dickstein, Arnaud Doucet, and Will Grathwohl.
\newblock {Reduce, Reuse, Recycle: Compositional Generation with Energy-Based
  Diffusion Models and MCMC}.
\newblock In \emph{Proceedings of the 40th International Conference on Machine
  Learning}, 2023.

\bibitem[Feng et~al.(2023{\natexlab{a}})Feng, Wang, Wang, Xu, and
  Liu]{feng2023metadreamer}
Lincong Feng, Muyu Wang, Maoyu Wang, Kuo Xu, and Xiaoli Liu.
\newblock {MetaDreamer: Efficient Text-to-3D Creation With Disentangling
  Geometry and Texture}.
\newblock \emph{arXiv preprint arXiv:2311.10123}, 2023{\natexlab{a}}.

\bibitem[Feng et~al.(2023{\natexlab{b}})Feng, Zhang, Yu, Fang, Li, Chen, Lu,
  Liu, Yin, Feng, Sun, Chen, Tian, Wu, and Wang]{feng2023ernievilg}
Zhida Feng, Zhenyu Zhang, Xintong Yu, Yewei Fang, Lanxin Li, Xuyi Chen, Yuxiang
  Lu, Jiaxiang Liu, Weichong Yin, Shikun Feng, Yu Sun, Li Chen, Hao Tian, Hua
  Wu, and Haifeng Wang.
\newblock {ERNIE-ViLG 2.0: Improving Text-to-Image Diffusion Model With
  Knowledge-Enhanced Mixture-of-Denoising-Experts}.
\newblock In \emph{Proceedings of the IEEE/CVF Conference on Computer Vision
  and Pattern Recognition (CVPR)}, 2023{\natexlab{b}}.

\bibitem[Ge et~al.(2023)Ge, Nah, Liu, Poon, Tao, Catanzaro, Jacobs, Huang, Liu,
  and Balaji]{ge2022pyoco}
Songwei Ge, Seungjun Nah, Guilin Liu, Tyler Poon, Andrew Tao, Bryan Catanzaro,
  David Jacobs, Jia-Bin Huang, Ming-Yu Liu, and Yogesh Balaji.
\newblock {Preserve Your Own Correlation: A Noise Prior for Video Diffusion
  Models}.
\newblock In \emph{Proceedings of the IEEE/CVF International Conference on
  Computer Vision (ICCV)}, 2023.

\bibitem[Girdhar et~al.(2023)Girdhar, Singh, Brown, Duval, Azadi, Rambhatla,
  Shah, Yin, Parikh, and Misra]{girdhar2023emu}
Rohit Girdhar, Mannat Singh, Andrew Brown, Quentin Duval, Samaneh Azadi,
  Sai~Saketh Rambhatla, Akbar Shah, Xi Yin, Devi Parikh, and Ishan Misra.
\newblock {Emu Video: Factorizing Text-to-Video Generation by Explicit Image
  Conditioning}.
\newblock \emph{arXiv preprint arXiv:2311.10709}, 2023.

\bibitem[Gu et~al.(2023)Gu, Zhai, Zhang, Susskind, and
  Jaitly]{gu2023matryoshka}
Jiatao Gu, Shuangfei Zhai, Yizhe Zhang, Josh Susskind, and Navdeep Jaitly.
\newblock {Matryoshka Diffusion Models}.
\newblock \emph{arXiv preprint arXiv:2310.15111}, 2023.

\bibitem[Guo et~al.(2023)Guo, Yang, Rao, Wang, Qiao, Lin, and
  Dai]{guo2023animatediff}
Yuwei Guo, Ceyuan Yang, Anyi Rao, Yaohui Wang, Yu Qiao, Dahua Lin, and Bo Dai.
\newblock {AnimateDiff: Animate Your Personalized Text-to-Image Diffusion
  Models without Specific Tuning}.
\newblock \emph{arXiv preprint arXiv:2307.04725}, 2023.

\bibitem[Ho and Salimans(2021)]{ho2021classifierfree}
Jonathan Ho and Tim Salimans.
\newblock {Classifier-Free Diffusion Guidance}.
\newblock In \emph{NeurIPS 2021 Workshop on Deep Generative Models and
  Downstream Applications}, 2021.

\bibitem[Ho et~al.(2020)Ho, Jain, and Abbeel]{ho2020ddpm}
Jonathan Ho, Ajay Jain, and Pieter Abbeel.
\newblock {Denoising Diffusion Probabilistic Models}.
\newblock In \emph{Advances in Neural Information Processing Systems
  (NeurIPS)}, 2020.

\bibitem[Ho et~al.(2022)Ho, Chan, Saharia, Whang, Gao, Gritsenko, Kingma,
  Poole, Norouzi, Fleet, and Salimans]{ho2022imagen}
Jonathan Ho, William Chan, Chitwan Saharia, Jay Whang, Ruiqi Gao, Alexey
  Gritsenko, Diederik~P. Kingma, Ben Poole, Mohammad Norouzi, David~J. Fleet,
  and Tim Salimans.
\newblock {Imagen Video: High Definition Video Generation with Diffusion
  Models}.
\newblock \emph{arXiv preprint arXiv:2210.02303}, 2022.

\bibitem[Hong et~al.(2023)Hong, Zhang, Gu, Bi, Zhou, Liu, Liu, Sunkavalli, Bui,
  and Tan]{hong2023lrm}
Yicong Hong, Kai Zhang, Jiuxiang Gu, Sai Bi, Yang Zhou, Difan Liu, Feng Liu,
  Kalyan Sunkavalli, Trung Bui, and Hao Tan.
\newblock {LRM: Large Reconstruction Model for Single Image to 3D}.
\newblock \emph{arXiv preprint arXiv:2311.04400}, 2023.

\bibitem[Hoogeboom et~al.(2023)Hoogeboom, Heek, and
  Salimans]{hoogeboom2023simplediffusion}
Emiel Hoogeboom, Jonathan Heek, and Tim Salimans.
\newblock {Simple Diffusion: End-to-End Diffusion for High Resolution Images}.
\newblock In \emph{Proceedings of the 40th International Conference on Machine
  Learning (ICML)}, 2023.

\bibitem[Huang et~al.(2023)Huang, Wang, Shi, Qi, Zha, and
  Zhang]{huang2023dreamtime}
Yukun Huang, Jianan Wang, Yukai Shi, Xianbiao Qi, Zheng-Jun Zha, and Lei Zhang.
\newblock {DreamTime: An Improved Optimization Strategy for Text-to-3D Content
  Creation}.
\newblock \emph{arXiv preprint arXiv:2306.12422}, 2023.

\bibitem[Jain et~al.(2022)Jain, Mildenhall, Barron, Abbeel, and
  Poole]{jain2022dreamfields}
A. Jain, B. Mildenhall, J.~T. Barron, P. Abbeel, and B. Poole.
\newblock {Zero-Shot Text-Guided Object Generation with Dream Fields}.
\newblock In \emph{Proceedings of the IEEE/CVF Conference on Computer Vision
  and Pattern Recognition (CVPR)}, 2022.

\bibitem[Jiang et~al.(2023)Jiang, Zhang, Gao, Hu, and
  Yao]{jiang2023consistent4d}
Yanqin Jiang, Li Zhang, Jin Gao, Weimin Hu, and Yao Yao.
\newblock {Consistent4D: Consistent 360$^{\circ}$ Dynamic Object Generation
  from Monocular Video}.
\newblock \emph{arXiv preprint arxiv:2311.02848}, 2023.

\bibitem[Kalischek et~al.(2022)Kalischek, Peters, Wegner, and
  Schindler]{kalischek2022tetrahedral}
Nikolai Kalischek, Torben Peters, Jan~D. Wegner, and Konrad Schindler.
\newblock {Tetrahedral Diffusion Models for 3D Shape Generation}.
\newblock \emph{arXiv preprint arXiv:2211.13220}, 2022.

\bibitem[Katzir et~al.(2023)Katzir, Patashnik, Cohen-Or, and
  Lischinski]{katzir2023noisefree}
Oren Katzir, Or Patashnik, Daniel Cohen-Or, and Dani Lischinski.
\newblock {Noise-Free Score Distillation}.
\newblock \emph{arXiv preprint arXiv:2310.17590}, 2023.

\bibitem[Kerbl et~al.(2023)Kerbl, Kopanas, Leimk{\"u}hler, and
  Drettakis]{kerbl20233Dgaussians}
Bernhard Kerbl, Georgios Kopanas, Thomas Leimk{\"u}hler, and George Drettakis.
\newblock {3D Gaussian Splatting for Real-Time Radiance Field Rendering}.
\newblock \emph{ACM Transactions on Graphics}, 42\penalty0 (4), 2023.

\bibitem[Kerlow(2009)]{kerlow2009book}
Isaac Kerlow.
\newblock \emph{{The Art of 3D Computer Animation and Effects}}.
\newblock Wiley Publishing, 4th edition, 2009.

\bibitem[Khachatryan et~al.(2023)Khachatryan, Movsisyan, Tadevosyan, Henschel,
  Wang, Navasardyan, and Shi]{khachatryan2023text2videozero}
Levon Khachatryan, Andranik Movsisyan, Vahram Tadevosyan, Roberto Henschel,
  Zhangyang Wang, Shant Navasardyan, and Humphrey Shi.
\newblock {Text2Video-Zero: Text-to-Image Diffusion Models are Zero-Shot Video
  Generators}.
\newblock \emph{arXiv preprint arXiv:2303.13439}, 2023.

\bibitem[Kim et~al.(2023)Kim, Brown, Yin, Kreis, Schwarz, Li, Rombach,
  Torralba, and Fidler]{kim2023nfldm}
Seung~Wook Kim, Bradley Brown, Kangxue Yin, Karsten Kreis, Katja Schwarz,
  Daiqing Li, Robin Rombach, Antonio Torralba, and Sanja Fidler.
\newblock {NeuralField-LDM: Scene Generation with Hierarchical Latent Diffusion
  Models}.
\newblock In \emph{Proceedings of the IEEE/CVF Conference on Computer Vision
  and Pattern Recognition (CVPR)}, 2023.

\bibitem[Liang et~al.(2023)Liang, Yang, Lin, Li, Xu, and
  Chen]{liang2023luciddreamer}
Yixun Liang, Xin Yang, Jiantao Lin, Haodong Li, Xiaogang Xu, and Yingcong Chen.
\newblock {LucidDreamer: Towards High-Fidelity Text-to-3D Generation via
  Interval Score Matching}.
\newblock \emph{arXiv preprint arXiv:2311.11284}, 2023.

\bibitem[Lin et~al.(2023)Lin, Gao, Tang, Takikawa, Zeng, Huang, Kreis, Fidler,
  Liu, and Lin]{lin2023magic3d}
Chen-Hsuan Lin, Jun Gao, Luming Tang, Towaki Takikawa, Xiaohui Zeng, Xun Huang,
  Karsten Kreis, Sanja Fidler, Ming-Yu Liu, and Tsung-Yi Lin.
\newblock {Magic3D: High-Resolution Text-to-3D Content Creation}.
\newblock In \emph{Proceedings of the IEEE/CVF Conference on Computer Vision
  and Pattern Recognition (CVPR)}, 2023.

\bibitem[Liu et~al.(2023{\natexlab{a}})Liu, Shi, Chen, Zhang, Xu, Wei, Chen,
  Zeng, Gu, and Su]{liu2023one2345}
Minghua Liu, Ruoxi Shi, Linghao Chen, Zhuoyang Zhang, Chao Xu, Xinyue Wei,
  Hansheng Chen, Chong Zeng, Jiayuan Gu, and Hao Su.
\newblock {One-2-3-45++: Fast Single Image to 3D Objects with Consistent
  Multi-View Generation and 3D Diffusion}.
\newblock \emph{arXiv preprint arXiv:2311.07885}, 2023{\natexlab{a}}.

\bibitem[Liu et~al.(2022)Liu, Li, Du, Torralba, and
  Tenenbaum]{liu2022compositional}
Nan Liu, Shuang Li, Yilun Du, Antonio Torralba, and Joshua~B. Tenenbaum.
\newblock {Compositional Visual Generation with Composable Diffusion Models}.
\newblock In \emph{Computer Vision -- ECCV 2022}, 2022.

\bibitem[Liu et~al.(2023{\natexlab{b}})Liu, Wu, Van~Hoorick, Tokmakov,
  Zakharov, and Vondrick]{liu2023zero123}
Ruoshi Liu, Rundi Wu, Basile Van~Hoorick, Pavel Tokmakov, Sergey Zakharov, and
  Carl Vondrick.
\newblock {Zero-1-to-3: Zero-shot One Image to 3D Object}.
\newblock In \emph{Proceedings of the IEEE/CVF International Conference on
  Computer Vision (ICCV)}, 2023{\natexlab{b}}.

\bibitem[Liu et~al.(2023{\natexlab{c}})Liu, Lin, Zeng, Long, Liu, Komura, and
  Wang]{liu2023syncdreamer}
Yuan Liu, Cheng Lin, Zijiao Zeng, Xiaoxiao Long, Lingjie Liu, Taku Komura, and
  Wenping Wang.
\newblock {SyncDreamer: Generating Multiview-consistent Images from a
  Single-view Image}.
\newblock \emph{arXiv preprint arXiv:2309.03453}, 2023{\natexlab{c}}.

\bibitem[Liu et~al.(2023{\natexlab{d}})Liu, Feng, Black, Nowrouzezahrai, Paull,
  and Liu]{liu2023MeshDiffusion}
Zhen Liu, Yao Feng, Michael~J. Black, Derek Nowrouzezahrai, Liam Paull, and
  Weiyang Liu.
\newblock {MeshDiffusion: Score-based Generative 3D Mesh Modeling}.
\newblock In \emph{International Conference on Learning Representations
  (ICLR)}, 2023{\natexlab{d}}.

\bibitem[Long et~al.(2023)Long, Guo, Lin, Liu, Dou, Liu, Ma, Zhang, Habermann,
  Theobalt, and Wang]{long2023wonder3d}
Xiaoxiao Long, Yuan-Chen Guo, Cheng Lin, Yuan Liu, Zhiyang Dou, Lingjie Liu,
  Yuexin Ma, Song-Hai Zhang, Marc Habermann, Christian Theobalt, and Wenping
  Wang.
\newblock {Wonder3D: Single Image to 3D using Cross-Domain Diffusion}.
\newblock \emph{arXiv preprint arXiv:2310.15008}, 2023.

\bibitem[Lorraine et~al.(2023)Lorraine, Xie, Zeng, Lin, Takikawa, Sharp, Lin,
  Liu, Fidler, and Lucas]{lorraine2023att3d}
Jonathan Lorraine, Kevin Xie, Xiaohui Zeng, Chen-Hsuan Lin, Towaki Takikawa,
  Nicholas Sharp, Tsung-Yi Lin, Ming-Yu Liu, Sanja Fidler, and James Lucas.
\newblock {ATT3D: Amortized Text-to-3D Object Synthesis}.
\newblock In \emph{Proceedings of the IEEE/CVF International Conference on
  Computer Vision (ICCV)}, 2023.

\bibitem[Lucas and Kanade(1981)]{lucas1981iterative}
Bruce~D Lucas and Takeo Kanade.
\newblock An iterative image registration technique with an application to
  stereo vision.
\newblock In \emph{IJCAI'81: 7th international joint conference on Artificial
  intelligence}, 1981.

\bibitem[Luiten et~al.(2023)Luiten, Kopanas, Leibe, and
  Ramanan]{luiten2023dynamic}
Jonathon Luiten, Georgios Kopanas, Bastian Leibe, and Deva Ramanan.
\newblock {Dynamic 3D Gaussians: Tracking by Persistent Dynamic View
  Synthesis}.
\newblock \emph{arXiv preprint arXiv:2308.09713}, 2023.

\bibitem[Luo and Hu(2021)]{luo2021diffusion}
Shitong Luo and Wei Hu.
\newblock {Diffusion Probabilistic Models for 3D Point Cloud Generation}.
\newblock In \emph{Proceedings of the IEEE/CVF Conference on Computer Vision
  and Pattern Recognition (CVPR)}, 2021.

\bibitem[Metzer et~al.(2023)Metzer, Richardson, Patashnik, Giryes, and
  Cohen-Or]{metzer2023latent}
G. Metzer, E. Richardson, O. Patashnik, R. Giryes, and D. Cohen-Or.
\newblock {Latent-NeRF for Shape-Guided Generation of 3D Shapes and Textures}.
\newblock In \emph{Proceedings of the IEEE/CVF Conference on Computer Vision
  and Pattern Recognition (CVPR)}, 2023.

\bibitem[Mihajlovic et~al.(2023)Mihajlovic, Prokudin, Pollefeys, and
  Tang]{mihajlovic2023ResFields}
Marko Mihajlovic, Sergey Prokudin, Marc Pollefeys, and Siyu Tang.
\newblock {{ResFields}: Residual Neural Fields for Spatiotemporal Signals}.
\newblock \emph{arXiv preprint arXiv:2309.03160}, 2023.

\bibitem[Mildenhall et~al.(2020)Mildenhall, Srinivasan, Tancik, Barron,
  Ramamoorthi, and Ng]{mildenhall2020nerf}
Ben Mildenhall, Pratul~P. Srinivasan, Matthew Tancik, Jonathan~T. Barron, Ravi
  Ramamoorthi, and Ren Ng.
\newblock {NeRF: Representing Scenes as Neural Radiance Fields for View
  Synthesis}.
\newblock In \emph{ECCV}, 2020.

\bibitem[Nair and Hinton(2010)]{nair2010rectified}
Vinod Nair and Geoffrey~E Hinton.
\newblock {Rectified Linear Units Improve Restricted Boltzmann Machines}.
\newblock In \emph{Proceedings of the 27th international conference on machine
  learning (ICML-10)}, pages 807--814, 2010.

\bibitem[Nichol et~al.(2022)Nichol, Jun, Dhariwal, Mishkin, and
  Chen]{nichol2022pointe}
Alex Nichol, Heewoo Jun, Prafulla Dhariwal, Pamela Mishkin, and Mark Chen.
\newblock {Point-E: A System for Generating 3D Point Clouds from Complex
  Prompts}, 2022.

\bibitem[Nichol and Dhariwal(2021)]{nichol2021improved}
Alexander~Quinn Nichol and Prafulla Dhariwal.
\newblock {Improved Denoising Diffusion Probabilistic Models}.
\newblock In \emph{Proceedings of the 38th International Conference on Machine
  Learning (ICML)}, 2021.

\bibitem[Park et~al.(2021{\natexlab{a}})Park, Azadi, Liu, Darrell, and
  Rohrbach]{park2021benchmark}
Dong~Huk Park, Samaneh Azadi, Xihui Liu, Trevor Darrell, and Anna Rohrbach.
\newblock {Benchmark for Compositional Text-to-Image Synthesis}.
\newblock In \emph{NeurIPS Datasets and Benchmarks}, 2021{\natexlab{a}}.

\bibitem[Park et~al.(2021{\natexlab{b}})Park, Sinha, Barron, Bouaziz, Goldman,
  Seitz, and Martin-Brualla]{park2021nerfies}
Keunhong Park, Utkarsh Sinha, Jonathan~T. Barron, Sofien Bouaziz, Dan~B
  Goldman, Steven~M. Seitz, and Ricardo Martin-Brualla.
\newblock {Nerfies: Deformable Neural Radiance Fields}.
\newblock In \emph{Proceedings of the IEEE/CVF International Conference on
  Computer Vision (ICCV)}, 2021{\natexlab{b}}.

\bibitem[Park et~al.(2021{\natexlab{c}})Park, Sinha, Hedman, Barron, Bouaziz,
  Goldman, Martin-Brualla, and Seitz]{park2021hypernerf}
Keunhong Park, Utkarsh Sinha, Peter Hedman, Jonathan~T. Barron, Sofien Bouaziz,
  Dan~B Goldman, Ricardo Martin-Brualla, and Steven~M. Seitz.
\newblock {HyperNeRF: A Higher-Dimensional Representation for Topologically
  Varying Neural Radiance Fields}.
\newblock \emph{ACM Trans. Graph.}, 40\penalty0 (6), 2021{\natexlab{c}}.

\bibitem[Podell et~al.(2023)Podell, English, Lacey, Blattmann, Dockhorn,
  Müller, Penna, and Rombach]{podell2023sdxl}
Dustin Podell, Zion English, Kyle Lacey, Andreas Blattmann, Tim Dockhorn, Jonas
  Müller, Joe Penna, and Robin Rombach.
\newblock {SDXL: Improving Latent Diffusion Models for High-Resolution Image
  Synthesis}.
\newblock \emph{arXiv preprint arXiv:2307.01952}, 2023.

\bibitem[Poole et~al.(2023)Poole, Jain, Barron, and
  Mildenhall]{poole2023dreamfusion}
Ben Poole, Ajay Jain, Jonathan~T. Barron, and Ben Mildenhall.
\newblock {DreamFusion: Text-to-3D using 2D Diffusion}.
\newblock In \emph{The Eleventh International Conference on Learning
  Representations (ICLR)}, 2023.

\bibitem[Pumarola et~al.(2020)Pumarola, Corona, Pons-Moll, and
  Moreno-Noguer]{pumarola2020d}
Albert Pumarola, Enric Corona, Gerard Pons-Moll, and Francesc Moreno-Noguer.
\newblock {D-NeRF: Neural Radiance Fields for Dynamic Scenes}.
\newblock In \emph{Proceedings of the IEEE/CVF Conference on Computer Vision
  and Pattern Recognition (CVPR)}, 2020.

\bibitem[Qian et~al.(2023)Qian, Mai, Hamdi, Ren, Siarohin, Li, Lee,
  Skorokhodov, Wonka, Tulyakov, and Ghanem]{qian2023magic123}
Guocheng Qian, Jinjie Mai, Abdullah Hamdi, Jian Ren, Aliaksandr Siarohin, Bing
  Li, Hsin-Ying Lee, Ivan Skorokhodov, Peter Wonka, Sergey Tulyakov, and
  Bernard Ghanem.
\newblock {Magic123: One Image to High-Quality 3D Object Generation Using Both
  2D and 3D Diffusion Priors}.
\newblock \emph{arXiv preprint arXiv:2306.17843}, 2023.

\bibitem[Radford et~al.(2021)Radford, Kim, Hallacy, Ramesh, Goh, Agarwal,
  Sastry, Askell, Mishkin, Clark, Krueger, and Sutskever]{radford2021clip}
Alec Radford, Jong~Wook Kim, Chris Hallacy, Aditya Ramesh, Gabriel Goh,
  Sandhini Agarwal, Girish Sastry, Amanda Askell, Pamela Mishkin, Jack Clark,
  Gretchen Krueger, and Ilya Sutskever.
\newblock {Learning Transferable Visual Models From Natural Language
  Supervision}.
\newblock In \emph{Proceedings of the 38th International Conference on Machine
  Learning (ICML)}, 2021.

\bibitem[Raj et~al.(2023)Raj, Kaza, Poole, Niemeyer, Mildenhall, Ruiz, Zada,
  Aberman, Rubenstein, Barron, Li, and Jampani]{raj2023dreambooth3d}
Amit Raj, Srinivas Kaza, Ben Poole, Michael Niemeyer, Ben Mildenhall, Nataniel
  Ruiz, Shiran Zada, Kfir Aberman, Michael Rubenstein, Jonathan Barron,
  Yuanzhen Li, and Varun Jampani.
\newblock {DreamBooth3D: Subject-Driven Text-to-3D Generation}.
\newblock In \emph{Proceedings of the IEEE/CVF International Conference on
  Computer Vision (ICCV)}, 2023.

\bibitem[Ramesh et~al.(2022)Ramesh, Dhariwal, Nichol, Chu, and
  Chen]{ramesh2022dalle2}
Aditya Ramesh, Prafulla Dhariwal, Alex Nichol, Casey Chu, and Mark Chen.
\newblock {Hierarchical Text-Conditional Image Generation with CLIP Latents}.
\newblock \emph{arXiv preprint arXiv:2204.06125}, 2022.

\bibitem[Rempe et~al.(2023)Rempe, Luo, Peng, Yuan, Kitani, Kreis, Fidler, and
  Litany]{rempeluo2023tracepace}
Davis Rempe, Zhengyi Luo, Xue~Bin Peng, Ye Yuan, Kris Kitani, Karsten Kreis,
  Sanja Fidler, and Or Litany.
\newblock {Trace and Pace: Controllable Pedestrian Animation via Guided
  Trajectory Diffusion}.
\newblock In \emph{Conference on Computer Vision and Pattern Recognition
  (CVPR)}, 2023.

\bibitem[Roeder et~al.(2017)Roeder, Wu, and Duvenaud]{roeder2017sticking}
Geoffrey Roeder, Yuhuai Wu, and David Duvenaud.
\newblock {Sticking the Landing: Simple, Lower-Variance Gradient Estimators for
  Variational Inference}.
\newblock In \emph{Advances in Neural Information Processing Systems
  (NeurIPS)}, 2017.

\bibitem[Rombach et~al.(2022)Rombach, Blattmann, Lorenz, Esser, and
  Ommer]{rombach2021highresolution}
Robin Rombach, Andreas Blattmann, Dominik Lorenz, Patrick Esser, and Bj{\"o}rn
  Ommer.
\newblock {High-Resolution Image Synthesis with Latent Diffusion Models}.
\newblock In \emph{Proceedings of the IEEE/CVF Conference on Computer Vision
  and Pattern Recognition (CVPR)}, 2022.

\bibitem[Ruiz et~al.(2023)Ruiz, Li, Jampani, Pritch, Rubinstein, and
  Aberman]{ruiz2023dreambooth}
Nataniel Ruiz, Yuanzhen Li, Varun Jampani, Yael Pritch, Michael Rubinstein, and
  Kfir Aberman.
\newblock {Dreambooth: Fine tuning text-to-image diffusion models for
  subject-driven generation}.
\newblock In \emph{Proceedings of the IEEE/CVF Conference on Computer Vision
  and Pattern Recognition (CVPR)}, 2023.

\bibitem[Saharia et~al.(2022)Saharia, Chan, Saxena, Li, Whang, Denton,
  Ghasemipour, Gontijo-Lopes, Ayan, Salimans, Ho, Fleet, and
  Norouzi]{saharia2022photorealistic}
Chitwan Saharia, William Chan, Saurabh Saxena, Lala Li, Jay Whang, Emily
  Denton, Seyed Kamyar~Seyed Ghasemipour, Raphael Gontijo-Lopes, Burcu~Karagol
  Ayan, Tim Salimans, Jonathan Ho, David~J. Fleet, and Mohammad Norouzi.
\newblock {Photorealistic Text-to-Image Diffusion Models with Deep Language
  Understanding}.
\newblock In \emph{Advances in Neural Information Processing Systems
  (NeurIPS)}, 2022.

\bibitem[Schuhmann et~al.(2022)Schuhmann, Beaumont, Vencu, Gordon, Wightman,
  Cherti, Coombes, Katta, Mullis, Wortsman, et~al.]{schuhmann2022laion}
Christoph Schuhmann, Romain Beaumont, Richard Vencu, Cade Gordon, Ross
  Wightman, Mehdi Cherti, Theo Coombes, Aarush Katta, Clayton Mullis, Mitchell
  Wortsman, et~al.
\newblock {LAION-5B: An open large-scale dataset for training next generation
  image-text models}.
\newblock In \emph{Advances in Neural Information Processing Systems
  (NeurIPS)}, 2022.

\bibitem[Schwarz et~al.(2023)Schwarz, Kim, Gao, Fidler, Geiger, and
  Kreis]{schwarz2023wildfusion}
Katja Schwarz, Seung~Wook Kim, Jun Gao, Sanja Fidler, Andreas Geiger, and
  Karsten Kreis.
\newblock {WildFusion: Learning 3D-Aware Latent Diffusion Models in View
  Space}.
\newblock \emph{arXiv preprint arXiv:2311.13570}, 2023.

\bibitem[Shi et~al.(2023{\natexlab{a}})Shi, Chen, Zhang, Liu, Xu, Wei, Chen,
  Zeng, and Su]{shi2023zero123plus}
Ruoxi Shi, Hansheng Chen, Zhuoyang Zhang, Minghua Liu, Chao Xu, Xinyue Wei,
  Linghao Chen, Chong Zeng, and Hao Su.
\newblock {Zero123++: a Single Image to Consistent Multi-view Diffusion Base
  Model}.
\newblock \emph{arXiv preprint arXiv:2310.15110}, 2023{\natexlab{a}}.

\bibitem[Shi et~al.(2023{\natexlab{b}})Shi, Wang, Ye, Long, Li, and
  Yang]{shi2023mvdream}
Yichun Shi, Peng Wang, Jianglong Ye, Mai Long, Kejie Li, and Xiao Yang.
\newblock {MVDream: Multi-view Diffusion for 3D Generation}.
\newblock \emph{arXiv preprint arXiv:2308.16512}, 2023{\natexlab{b}}.

\bibitem[Shue et~al.(2023)Shue, Chan, Po, Ankner, Wu, and
  Wetzstein]{shue2023triplanediffusion}
J. Shue, E. Chan, R. Po, Z. Ankner, J. Wu, and G. Wetzstein.
\newblock {3D Neural Field Generation Using Triplane Diffusion}.
\newblock In \emph{Proceedings of the IEEE/CVF Conference on Computer Vision
  and Pattern Recognition (CVPR)}, 2023.

\bibitem[Singer et~al.(2023{\natexlab{a}})Singer, Polyak, Hayes, Yin, An,
  Zhang, Hu, Yang, Ashual, Gafni, Parikh, Gupta, and
  Taigman]{singer2023makeavideo}
Uriel Singer, Adam Polyak, Thomas Hayes, Xi Yin, Jie An, Songyang Zhang, Qiyuan
  Hu, Harry Yang, Oron Ashual, Oran Gafni, Devi Parikh, Sonal Gupta, and Yaniv
  Taigman.
\newblock {Make-A-Video: Text-to-Video Generation without Text-Video Data}.
\newblock In \emph{The Eleventh International Conference on Learning
  Representations (ICLR)}, 2023{\natexlab{a}}.

\bibitem[Singer et~al.(2023{\natexlab{b}})Singer, Sheynin, Polyak, Ashual,
  Makarov, Kokkinos, Goyal, Vedaldi, Parikh, Johnson, and
  Taigman]{singer2023mav3d}
Uriel Singer, Shelly Sheynin, Adam Polyak, Oron Ashual, Iurii Makarov, Filippos
  Kokkinos, Naman Goyal, Andrea Vedaldi, Devi Parikh, Justin Johnson, and Yaniv
  Taigman.
\newblock {Text-to-4D Dynamic Scene Generation}.
\newblock In \emph{Proceedings of the 40th International Conference on Machine
  Learning}, 2023{\natexlab{b}}.

\bibitem[Sohl-Dickstein et~al.(2015)Sohl-Dickstein, Weiss, Maheswaranathan, and
  Ganguli]{sohl2015deep}
Jascha Sohl-Dickstein, Eric Weiss, Niru Maheswaranathan, and Surya Ganguli.
\newblock {Deep Unsupervised Learning using Nonequilibrium Thermodynamics}.
\newblock In \emph{International Conference on Machine Learning (ICML)}, 2015.

\bibitem[Song et~al.(2021)Song, Sohl-Dickstein, Kingma, Kumar, Ermon, and
  Poole]{song2020score}
Yang Song, Jascha Sohl-Dickstein, Diederik~P Kingma, Abhishek Kumar, Stefano
  Ermon, and Ben Poole.
\newblock {Score-Based Generative Modeling through Stochastic Differential
  Equations}.
\newblock In \emph{International Conference on Learning Representations
  (ICLR)}, 2021.

\bibitem[Sun et~al.(2023)Sun, Zhang, Shao, Wang, Liu, Xie, and
  Liu]{sun2023dreamcraft3d}
Jingxiang Sun, Bo Zhang, Ruizhi Shao, Lizhen Wang, Wen Liu, Zhenda Xie, and
  Yebin Liu.
\newblock {DreamCraft3D: Hierarchical 3D Generation with Bootstrapped Diffusion
  Prior}.
\newblock \emph{arXiv preprint arXiv:2310.16818}, 2023.

\bibitem[Tang et~al.(2023)Tang, Ren, Zhou, Liu, and
  Zeng]{tang2023dreamgaussian}
Jiaxiang Tang, Jiawei Ren, Hang Zhou, Ziwei Liu, and Gang Zeng.
\newblock {DreamGaussian: Generative Gaussian Splatting for Efficient 3D
  Content Creation}.
\newblock \emph{arXiv preprint arXiv:2309.16653}, 2023.

\bibitem[Tretschk et~al.(2021)Tretschk, Tewari, Golyanik, Zollh\"{o}fer,
  Lassner, and Theobalt]{tretschk2021nonrigid}
Edgar Tretschk, Ayush Tewari, Vladislav Golyanik, Michael Zollh\"{o}fer,
  Christoph Lassner, and Christian Theobalt.
\newblock {Non-Rigid Neural Radiance Fields: Reconstruction and Novel View
  Synthesis of a Dynamic Scene From Monocular Video}.
\newblock In \emph{Proceedings of the IEEE/CVF International Conference on
  Computer Vision (ICCV)}, 2021.

\bibitem[Tsalicoglou et~al.(2024)Tsalicoglou, Manhardt, Tonioni, Niemeyer, and
  Tombari]{tsalicoglou2024textmesh}
Christina Tsalicoglou, Fabian Manhardt, Alessio Tonioni, Michael Niemeyer, and
  Federico Tombari.
\newblock {TextMesh: Generation of Realistic 3D Meshes From Text Prompts}.
\newblock In \emph{International conference on 3D vision (3DV)}, 2024.

\bibitem[Vahdat et~al.(2021)Vahdat, Kreis, and Kautz]{vahdat2021score}
Arash Vahdat, Karsten Kreis, and Jan Kautz.
\newblock {Score-based Generative Modeling in Latent Space}.
\newblock In \emph{Neural Information Processing Systems (NeurIPS)}, 2021.

\bibitem[Vaswani et~al.(2017)Vaswani, Shazeer, Parmar, Uszkoreit, Jones, Gomez,
  Kaiser, and Polosukhin]{vaswani2017attention}
Ashish Vaswani, Noam Shazeer, Niki Parmar, Jakob Uszkoreit, Llion Jones,
  Aidan~N Gomez, {\L}ukasz Kaiser, and Illia Polosukhin.
\newblock {Attention Is All You Need}.
\newblock \emph{Advances in neural information processing systems}, 30, 2017.

\bibitem[Wang et~al.(2023{\natexlab{a}})Wang, Du, Li, Yeh, and
  Shakhnarovich]{wang2023score}
Haochen Wang, Xiaodan Du, Jiahao Li, Raymond~A. Yeh, and Greg Shakhnarovich.
\newblock {Score Jacobian Chaining: Lifting Pretrained 2D Diffusion Models for
  3D Generation}.
\newblock In \emph{Proceedings of the IEEE/CVF Conference on Computer Vision
  and Pattern Recognition (CVPR)}, 2023{\natexlab{a}}.

\bibitem[Wang et~al.(2023{\natexlab{b}})Wang, Zhang, Zhang, Gu, Bao,
  Baltrusaitis, Shen, Chen, Wen, Chen, and Guo]{wang2023rodin}
Tengfei Wang, Bo Zhang, Ting Zhang, Shuyang Gu, Jianmin Bao, Tadas
  Baltrusaitis, Jingjing Shen, Dong Chen, Fang Wen, Qifeng Chen, and Baining
  Guo.
\newblock {RODIN: A Generative Model for Sculpting 3D Digital Avatars Using
  Diffusion}.
\newblock In \emph{Proceedings of the IEEE/CVF Conference on Computer Vision
  and Pattern Recognition (CVPR)}, 2023{\natexlab{b}}.

\bibitem[Wang et~al.(2023{\natexlab{c}})Wang, Yang, Tuo, He, Zhu, Fu, and
  Liu]{wang2023videofactory}
Wenjing Wang, Huan Yang, Zixi Tuo, Huiguo He, Junchen Zhu, Jianlong Fu, and
  Jiaying Liu.
\newblock {VideoFactory: Swap Attention in Spatiotemporal Diffusions for
  Text-to-Video Generation}.
\newblock \emph{arXiv preprint arXiv:2305.10874}, 2023{\natexlab{c}}.

\bibitem[Wang et~al.(2023{\natexlab{d}})Wang, Chen, Ma, Zhou, Huang, Wang,
  Yang, He, Yu, Yang, Guo, Wu, Si, Jiang, Chen, Loy, Dai, Lin, Qiao, and
  Liu]{wang2023lavie}
Yaohui Wang, Xinyuan Chen, Xin Ma, Shangchen Zhou, Ziqi Huang, Yi Wang, Ceyuan
  Yang, Yinan He, Jiashuo Yu, Peiqing Yang, Yuwei Guo, Tianxing Wu, Chenyang
  Si, Yuming Jiang, Cunjian Chen, Chen~Change Loy, Bo Dai, Dahua Lin, Yu Qiao,
  and Ziwei Liu.
\newblock {LAVIE: High-Quality Video Generation with Cascaded Latent Diffusion
  Models}.
\newblock \emph{arXiv preprint arXiv:2309.15103}, 2023{\natexlab{d}}.

\bibitem[Wang et~al.(2023{\natexlab{e}})Wang, Lu, Wang, Bao, Li, Su, and
  Zhu]{wang2023prolificdreamer}
Zhengyi Wang, Cheng Lu, Yikai Wang, Fan Bao, Chongxuan Li, Hang Su, and Jun
  Zhu.
\newblock {ProlificDreamer: High-Fidelity and Diverse Text-to-3D Generation
  with Variational Score Distillation}.
\newblock In \emph{Thirty-seventh Conference on Neural Information Processing
  Systems (NeurIPS)}, 2023{\natexlab{e}}.

\bibitem[Wu et~al.(2023{\natexlab{a}})Wu, Yi, Fang, Xie, Zhang, Wei, Liu, Tian,
  and Wang]{wu20234d}
Guanjun Wu, Taoran Yi, Jiemin Fang, Lingxi Xie, Xiaopeng Zhang, Wei Wei, Wenyu
  Liu, Qi Tian, and Xinggang Wang.
\newblock {4D Gaussian Splatting for Real-Time Dynamic Scene Rendering}.
\newblock \emph{arXiv preprint arXiv:2310.08528}, 2023{\natexlab{a}}.

\bibitem[Wu et~al.(2023{\natexlab{b}})Wu, Ge, Wang, Lei, Gu, Shi, Hsu, Shan,
  Qie, and Shou]{wu2023tune}
Jay~Zhangjie Wu, Yixiao Ge, Xintao Wang, Stan~Weixian Lei, Yuchao Gu, Yufei
  Shi, Wynne Hsu, Ying Shan, Xiaohu Qie, and Mike~Zheng Shou.
\newblock {Tune-a-video: One-shot tuning of image diffusion models for
  text-to-video generation}.
\newblock In \emph{Proceedings of the IEEE/CVF International Conference on
  Computer Vision (ICCV)}, 2023{\natexlab{b}}.

\bibitem[Xie et~al.(2023)Xie, Zong, Qiu, Li, Feng, Yang, and
  Jiang]{xie2023physgaussian}
Tianyi Xie, Zeshun Zong, Yuxing Qiu, Xuan Li, Yutao Feng, Yin Yang, and
  Chenfanfu Jiang.
\newblock {PhysGaussian: Physics-Integrated 3D Gaussians for Generative
  Dynamics}.
\newblock \emph{arXiv preprint arXiv:2311.12198}, 2023.

\bibitem[Xu et~al.(2023{\natexlab{a}})Xu, Jiang, Wang, Fan, Wang, and
  Wang]{xu2023neuralLift}
Dejia Xu, Yifan Jiang, Peihao Wang, Zhiwen Fan, Yi Wang, and Zhangyang Wang.
\newblock {NeuralLift-360: Lifting An In-the-wild 2D Photo to A 3D Object with
  360° Views}.
\newblock In \emph{Proceedings of the IEEE/CVF Conference on Computer Vision
  and Pattern Recognition (CVPR)}, 2023{\natexlab{a}}.

\bibitem[Xu et~al.(2023{\natexlab{b}})Xu, Tan, Luan, Bi, Wang, Li, Shi,
  Sunkavalli, Wetzstein, Xu, and Zhang]{xu2023dmv3d}
Yinghao Xu, Hao Tan, Fujun Luan, Sai Bi, Peng Wang, Jiahao Li, Zifan Shi,
  Kalyan Sunkavalli, Gordon Wetzstein, Zexiang Xu, and Kai Zhang.
\newblock {DMV3D: Denoising Multi-View Diffusion using 3D Large Reconstruction
  Model}.
\newblock \emph{arXiv preprint arXiv:2311.09217}, 2023{\natexlab{b}}.

\bibitem[Xue et~al.(2023)Xue, Song, Guo, Liu, Zong, Liu, and
  Luo]{xue2023raphael}
Zeyue Xue, Guanglu Song, Qiushan Guo, Boxiao Liu, Zhuofan Zong, Yu Liu, and
  Ping Luo.
\newblock {RAPHAEL: Text-to-Image Generation via Large Mixture of Diffusion
  Paths}.
\newblock \emph{arXiv preprint arXiv:2305.18295}, 2023.

\bibitem[Yang et~al.(2022)Yang, Vo, Neverova, Ramanan, Vedaldi, and
  Joo]{yang2022banmo}
Gengshan Yang, Minh Vo, Natalia Neverova, Deva Ramanan, Andrea Vedaldi, and
  Hanbyul Joo.
\newblock {BANMo: Building Animatable 3D Neural Models from Many Casual
  Videos}.
\newblock In \emph{Proceedings of the IEEE/CVF Conference on Computer Vision
  and Pattern Recognition (CVPR)}, 2022.

\bibitem[Yang et~al.(2023)Yang, Gao, Zhou, Jiao, Zhang, and
  Jin]{yang2023deformable}
Ziyi Yang, Xinyu Gao, Wen Zhou, Shaohui Jiao, Yuqing Zhang, and Xiaogang Jin.
\newblock {Deformable 3D Gaussians for High-Fidelity Monocular Dynamic Scene
  Reconstruction}.
\newblock \emph{arXiv preprint arXiv:2309.13101}, 2023.

\bibitem[Yi et~al.(2023)Yi, Fang, Wu, Xie, Zhang, Liu, Tian, and
  Wang]{yi2023gaussiandreamer}
Taoran Yi, Jiemin Fang, Guanjun Wu, Lingxi Xie, Xiaopeng Zhang, Wenyu Liu, Qi
  Tian, and Xinggang Wang.
\newblock {GaussianDreamer: Fast Generation from Text to 3D Gaussian Splatting
  with Point Cloud Priors}.
\newblock \emph{arXiv preprint arxiv:2310.08529}, 2023.

\bibitem[Yu et~al.(2023)Yu, Guo, Li, Liang, Zhang, and Qi]{yu2023csd}
Xin Yu, Yuan-Chen Guo, Yangguang Li, Ding Liang, Song-Hai Zhang, and Xiaojuan
  Qi.
\newblock {Text-to-3D with Classifier Score Distillation}.
\newblock \emph{arXiv preprint arXiv:2310.19415}, 2023.

\bibitem[Yuan et~al.(2023)Yuan, Song, Iqbal, Vahdat, and
  Kautz]{yuan2023physdiff}
Ye Yuan, Jiaming Song, Umar Iqbal, Arash Vahdat, and Jan Kautz.
\newblock {PhysDiff: Physics-Guided Human Motion Diffusion Model}.
\newblock In \emph{Proceedings of the IEEE/CVF International Conference on
  Computer Vision (ICCV)}, 2023.

\bibitem[Zeng et~al.(2022)Zeng, Vahdat, Williams, Gojcic, Litany, Fidler, and
  Kreis]{zeng2022lion}
Xiaohui Zeng, Arash Vahdat, Francis Williams, Zan Gojcic, Or Litany, Sanja
  Fidler, and Karsten Kreis.
\newblock {LION: Latent Point Diffusion Models for 3D Shape Generation}.
\newblock In \emph{Advances in Neural Information Processing Systems
  (NeurIPS)}, 2022.

\bibitem[Zhao et~al.(2023)Zhao, Yan, Xie, Hong, Li, and
  Lee]{zhao2023animate124}
Yuyang Zhao, Zhiwen Yan, Enze Xie, Lanqing Hong, Zhenguo Li, and Gim~Hee Lee.
\newblock {Animate124: Animating One Image to 4D Dynamic Scene}.
\newblock \emph{arXiv preprint arXiv:2311.14603}, 2023.

\bibitem[Zheng et~al.(2023)Zheng, Li, Nagano, Liu, Kreis, Hilliges, and
  Mello]{zheng2023unified}
Yufeng Zheng, Xueting Li, Koki Nagano, Sifei Liu, Karsten Kreis, Otmar
  Hilliges, and Shalini~De Mello.
\newblock {A Unified Approach for Text- and Image-guided 4D Scene Generation}.
\newblock \emph{arXiv preprint arXiv:2311.16854}, 2023.

\bibitem[Zhou et~al.(2023)Zhou, Wang, Yan, Lv, Zhu, and
  Feng]{zhou2023magicvideo}
Daquan Zhou, Weimin Wang, Hanshu Yan, Weiwei Lv, Yizhe Zhu, and Jiashi Feng.
\newblock {MagicVideo: Efficient Video Generation With Latent Diffusion
  Models}.
\newblock \emph{arXiv preprint arXiv:2211.11018}, 2023.

\bibitem[Zhou et~al.(2021)Zhou, Du, and Wu]{zhou2021pvd}
Linqi Zhou, Yilun Du, and Jiajun Wu.
\newblock {3D Shape Generation and Completion Through Point-Voxel Diffusion}.
\newblock In \emph{Proceedings of the IEEE/CVF International Conference on
  Computer Vision (ICCV)}, 2021.

\bibitem[Zhu and Zhuang(2023)]{zhu2023hifa}
Junzhe Zhu and Peiye Zhuang.
\newblock {HiFA: High-fidelity Text-to-3D with Advanced Diffusion Guidance}.
\newblock \emph{arXiv preprint arXiv:2305.18766}, 2023.

\bibitem[Zielonka et~al.(2023)Zielonka, Bagautdinov, Saito, Zollhöfer, Thies,
  and Romero]{zielonka2023drivable}
Wojciech Zielonka, Timur Bagautdinov, Shunsuke Saito, Michael Zollhöfer,
  Justus Thies, and Javier Romero.
\newblock {Drivable 3D Gaussian Avatars}.
\newblock \emph{arXiv preprint arxiv:2311.08581}, 2023.

\bibitem[Zwicker et~al.(2001)Zwicker, Pfister, van Baar, and
  Gross]{zwicker2001splatting}
M. Zwicker, H. Pfister, J. van Baar, and M. Gross.
\newblock {EWA volume splatting}.
\newblock In \emph{Proceedings Visualization, 2001. VIS '01.}, 2001.

\end{thebibliography}
